\documentclass[10pt,twocolumn,letterpaper]{article}

\usepackage{wacv}
\usepackage{times}
\usepackage{epsfig}
\usepackage{graphicx}
\usepackage{amsmath}
\usepackage{amssymb}

\usepackage{caption}
\usepackage{subcaption}
\usepackage{array}

\usepackage[pagebackref=true,breaklinks=true,letterpaper=true,colorlinks,bookmarks=false]{hyperref}

\newcolumntype{C}[1]{>{\centering\let\newline\\\arraybackslash\hspace{0pt}}m{#1}}
\newcolumntype{L}[1]{>{\raggedright\let\newline\\\arraybackslash\hspace{0pt}}m{#1}}

\newcommand\Tstrut{\rule{0pt}{2.2ex}}       
\newcommand\Bstrut{\rule[-1.0ex]{0pt}{0pt}} 
\newcommand{\TBstrut}{\Tstrut\Bstrut}       

\wacvfinalcopy 

\begin{document}

\title{Learning to Detect Human-Object Interactions}

\renewcommand\footnotemark{}
\author{Yu-Wei Chao$^1$, Yunfan Liu$^1$, Xieyang Liu$^1$, Huayi Zeng$^2$, and Jia Deng$^1$ \vspace{3mm} \\
  \begin{minipage}{0.5\textwidth}
    \centering
    $^1$University of Michigan, Ann Arbor
    {\tt\small \{ywchao,yunfan,lxieyang,jiadeng\}@umich.edu}\\
  \end{minipage}
  \begin{minipage}{0.5\textwidth}
    \centering
    $^2$Washington University in St. Louis$^*$\thanks{$^*$Work done at the University of Michigan as a visiting student.}
    {\tt\small \{zengh\}@wustl.edu}\\
  \end{minipage}
}

\maketitle

\begin{abstract}
We study the problem of detecting human-object interactions (HOI) in static
images, defined as predicting a human and an object bounding box with an
interaction class label that connects them. HOI detection is a fundamental
problem in computer vision as it provides semantic information about the
interactions among the detected objects. We introduce HICO-DET, a new large
benchmark for HOI detection, by augmenting the current HICO classification
benchmark with instance annotations. To solve the task, we propose Human-Object
Region-based Convolutional Neural Networks (HO-RCNN). At the core of our
HO-RCNN is the \textit{Interaction Pattern}, a novel DNN input that
characterizes the spatial relations between two bounding boxes. Experiments on
HICO-DET demonstrate that our HO-RCNN, by exploiting human-object spatial
relations through Interaction Patterns, significantly improves the performance
of HOI detection over baseline approaches.
\end{abstract}

\section{Introduction}

Visual recognition of human-object interactions (HOI) (e.g. ``riding a horse'',
``eating a sandwich'') is a fundamental problem in computer vision. Successful
HOI recognition could identify not only objects but also the relationships
between them, providing a deeper understanding of the semantics of visual
scenes than just object recognition
\cite{krizhevsky:nips2012,szegedy:cvpr2015,he:cvpr2016} or object detection
\cite{girshick:iccv2015,ren:nips2015,liu:eccv2016,dai:nips2016}. Without HOI
recognition, an image can only be interpreted as a collection of object
bounding boxes. An AI system can only pick up information such as ``A baseball
bat is in the right corner'' and ``A boy is close to the baseball bat'', but
not ``A boy wearing a cap is swinging a baseball bat''.

HOI recognition has recently attracted increasing attention in the field of
computer vision
\cite{gupta:pami2009,yao:cvpr2010a,yao:cvpr2010b,desai:cvprw2010,maji:cvpr2011,delaitre:nips2011,desai:eccv2012,prest:pami2012,hu:iccv2013}.
While significant progress has been made, the problem of HOI recognition is
still far from being solved. A key issue is that these approaches have been
evaluated using small datasets with limited HOI categories, e.g. 10 categories
in PASCAL VOC \cite{everingham:ijcv2015} and 40 categories in Stanford 40
Actions \cite{yao:iccv2011}. Furthermore, these datasets offer only a limited
variety of interaction classes for each object category. For example, in
Stanford 40 Actions, ``repairing a car'' is the only HOI category involving the
object ``car''. It is unclear whether a successful algorithm can really
recognize the interaction (e.g. ``repairing''), or whether it simply recognizes
the present object (e.g. ``car''). This issue has recently been addressed by
\cite{chao:iccv2015}, which introduced ``Humans interacting with Common
Objects'' (HICO), a large image dataset containing 600 HOI categories over 80
common object categories and featuring a diverse set of interactions for each
object category. HICO was used in \cite{chao:iccv2015} to provide the first
benchmark for image-level HOI classification, i.e. classifying whether an HOI
class is present in an image.

\begin{figure}[t]
  \centering
  \begin{subfigure}[c]{0.40\linewidth}
    \centering
    \includegraphics[height=0.095\textheight]{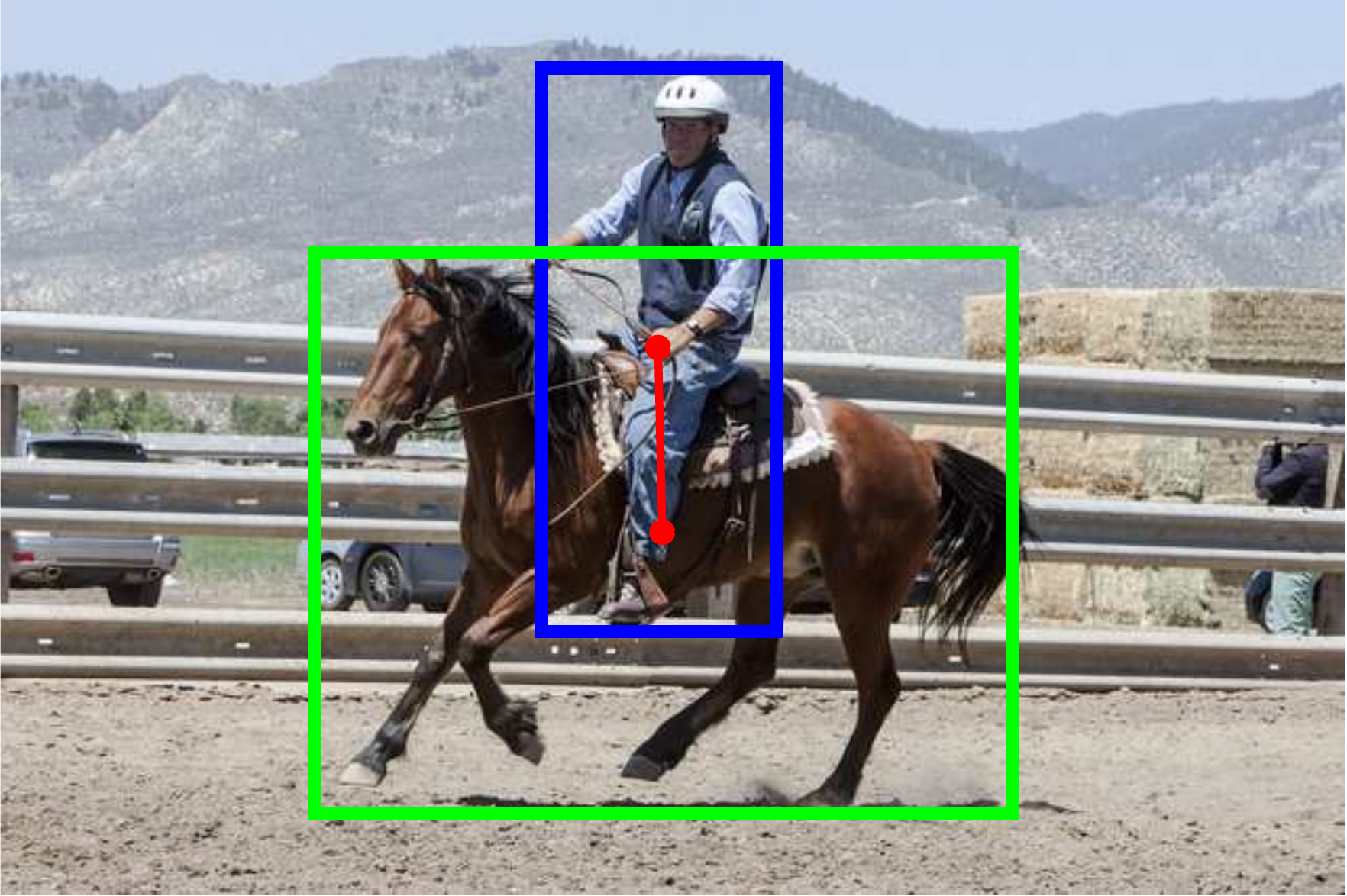}
    \caption{riding a horse}
  \end{subfigure}
  ~~~
  \begin{subfigure}[c]{0.40\linewidth}
    \centering
    \includegraphics[height=0.095\textheight]{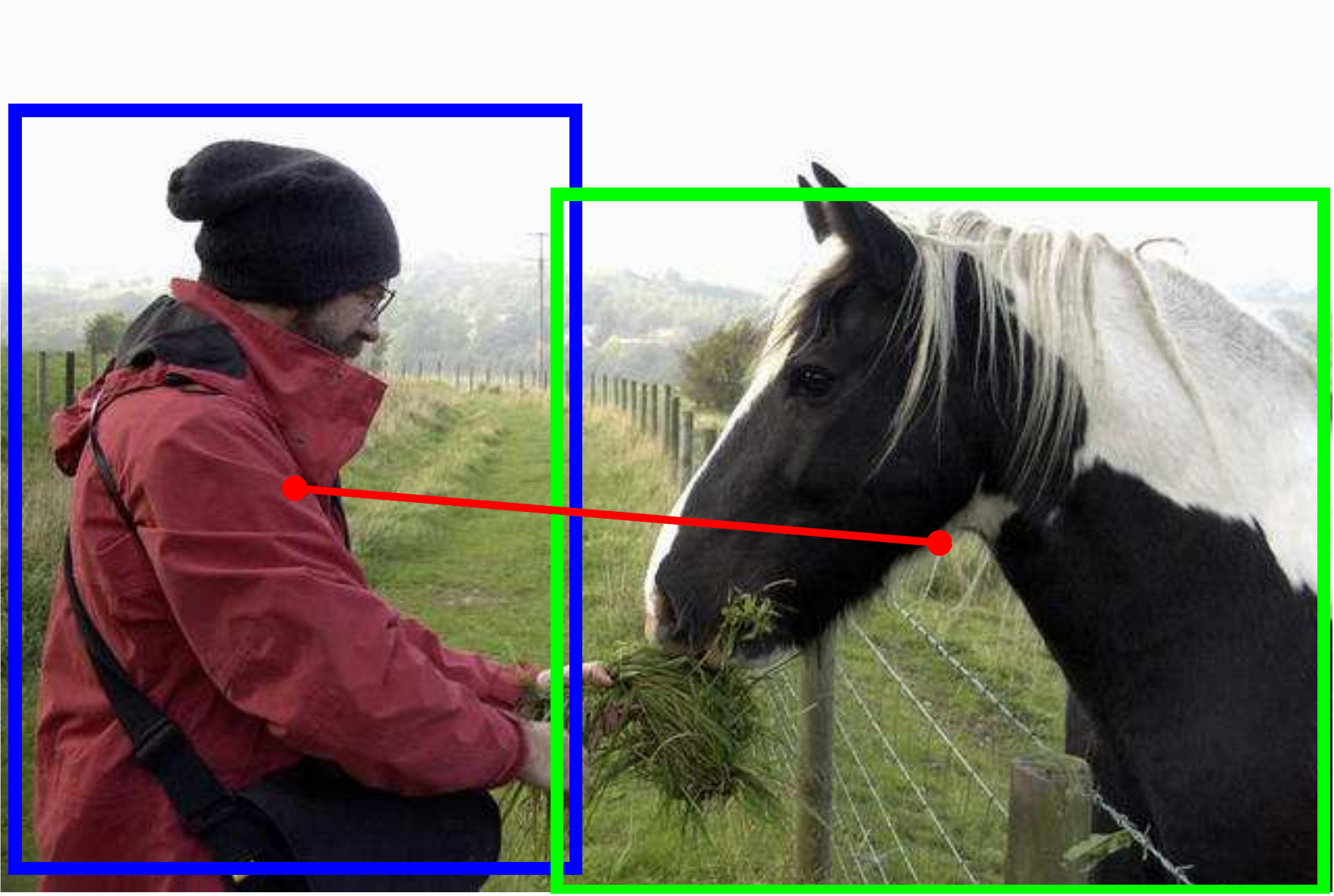}
    \caption{feeding horses}
  \end{subfigure}
  \\
  \vspace{1mm}
  \begin{subfigure}[c]{0.40\linewidth}
    \centering
    \includegraphics[height=0.095\textheight]{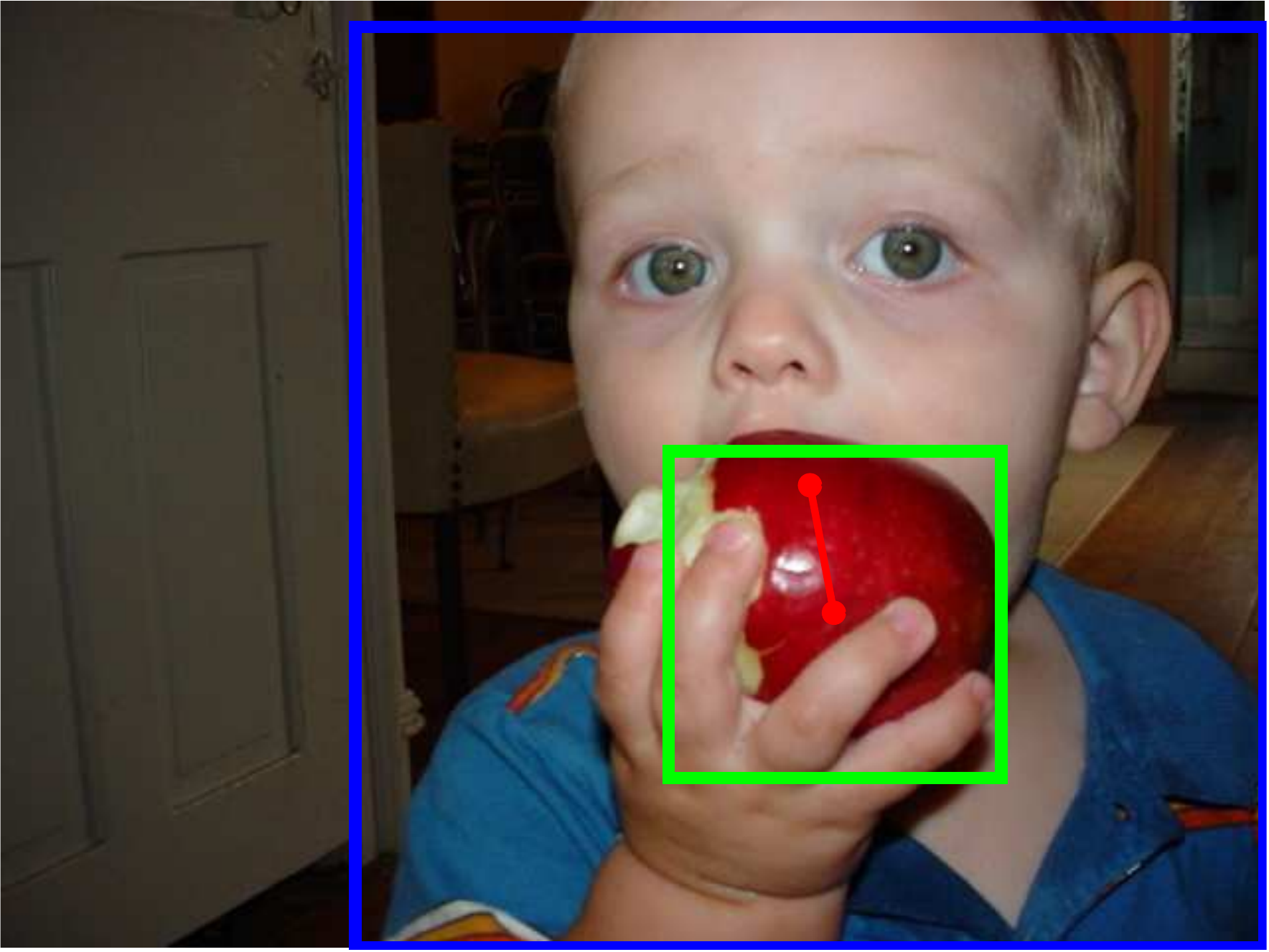}
    \caption{eating an apple}
  \end{subfigure}
  ~~~
  \begin{subfigure}[c]{0.40\linewidth}
    \centering
    \includegraphics[height=0.095\textheight]{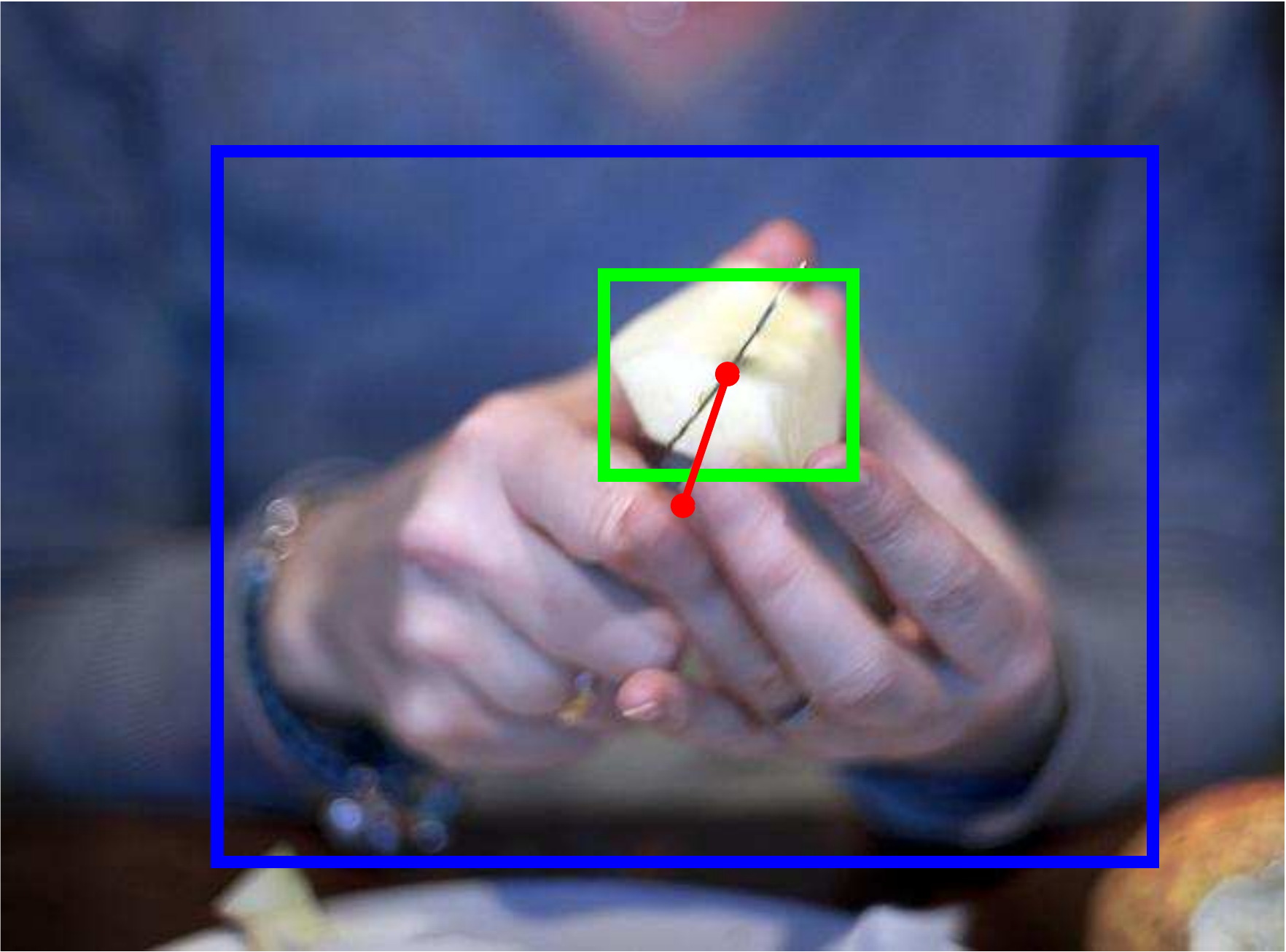}
    \caption{cutting an apple}
  \end{subfigure}
  \vspace{-2mm}
  \caption{\small Detecting human-object interactions. Blue boxes mark the
humans. Green boxes mark the objects. Each red line links a person and an
object involved in the labeled HOI class.}
  \label{fig:pull_fig}
\end{figure}

While the introduction of HICO may facilitate progress in the study of HOI
classification, HOI recognition still cannot be fully addressed, since with
only HOI classification computers are not able to accurately localize the
present interactions in images. To be able to ground HOIs to image regions, we
propose studying a new problem: detecting human-object interactions in static
images. The goal of HOI detection is not only to determine the presence of
HOIs, but also to estimate their locations. Formally, we define the problem of
HOI detection as predicting a pair of bounding boxes---first for a person and
second for an object---and identifying the interaction class, as illustrated in
Fig.~\ref{fig:pull_fig}. This is different from conventional object detection,
where the output is only a single bounding box with a class label. Addressing
HOI detection will bridge the gap between HOI classification and object
detection by identifying the interaction relations between detected objects.

The contributions of this paper are two-fold: (1) We introduce HICO-DET, the
first large benchmark for HOI detection, by augmenting the current HICO
classification benchmark with instance annotations. HICO-DET offers more than
150K annotated instances of human-object pairs, spanning the 600 HOI categories
in HICO, i.e. an average of 250 instances per HOI category. (2) We propose
Human-Object Region-based Convolutional Neural Networks (HO-RCNN), a DNN-based
framework that extends state-of-the-art region-based object detectors
\cite{girshick:cvpr2014,girshick:iccv2015,ren:nips2015} from detecting a single
bounding box to a pair of bounding boxes. At the core of our HO-RCNN is the
Interaction Pattern, a novel DNN input that characterizes the spatial relations
between two bounding boxes. Experiments on HICO-DET demonstrate that our
HO-RCNN, by exploiting human-object spatial relations through Interaction
Patterns, significantly improves the performance of HOI detection over baseline
approaches. The dataset and code are publicly available at
\href{http://www.umich.edu/~ywchao/hico/}{\texttt{http://www.umich.edu/$\sim$ywchao/hico/}}.

\section{Related Work}

\paragraph{HOI Recognition} A surge of work on HOI recognition has been
published since 2009. Results produced in these works were evaluated on either
action classification
\cite{yao:cvpr2010a,yao:cvpr2010b,desai:cvprw2010,maji:cvpr2011,delaitre:nips2011,desai:eccv2012,prest:pami2012,hu:iccv2013},
object detection \cite{yao:cvpr2010a}, or human pose estimation
\cite{yao:cvpr2010a,desai:eccv2012}; none of them were directly evaluated on
HOI detection. Chao \textit{et al.} \cite{chao:iccv2015} recently contributed a
large image dataset ``HICO'' for HOI classification
\cite{chao:iccv2015,mallya:eccv2016}. However, HICO does not provide
ground-truth annotations for evaluating HOI detection, which motivates us to
construct a new benchmark by augmenting HICO. We also highlight a few other
recent datasets. Gupta and Malik \cite{gupta:arxiv2015} augmented MS-COCO
\cite{lin:eccv2014} by connecting interacting people and objects and labeling
their semantic roles. Yatskar \emph{et al.} \cite{yatskar:cvpr2016} contributed
an image dataset for situation recognition, defined as identifying the activity
together with the participating objects and their roles. Both datasets, unlike
HICO, do not offer a diverse set of interaction classes for each object
category. Lu \emph{et al.} \cite{lu:eccv2016} and Krishna \emph{et al.}
\cite{krishna:ijcv2017} separately introduced two image datasets for detecting
object relationships. While they feature a diverse set of relationships, the
relationships are not exhaustively labeled in each image. As a result,
follow-up works
\cite{dai:cvpr2017,liang:cvpr2017,li:cvpr2017,zhang1:cvpr2017,zhang2:cvpr2017}
which benchmark on these datasets can only evaluate their detection result with
recall, but not precision. In contrast, we exhaustively labeled all the
instances for each positive HOI label in each image, enabling us to evaluate
our result with mean Average Precision (mAP).

\vspace{-3mm}

\paragraph{Object Detection} Standard object detectors
\cite{girshick:iccv2015,ren:nips2015,liu:eccv2016,dai:nips2016} only produce a
class-specific bounding box around each object instance; they do not label the
interaction among objects. Sadeghi and Farhadi \cite{sadeghi:cvpr2011} proposed
``visual phrases'' by treating each pair of interacting objects as a unit and
leveraged object detectors to localize them. HOI detection further extends the
detection of ``visual phrases'' to localize individual objects in each pair.
Our proposed HO-RCNN, built on recent advances in object detection, extends
region-based object detectors
\cite{girshick:cvpr2014,girshick:iccv2015,ren:nips2015} from taking single
bounding boxes to taking bounding box pairs.

\vspace{-3mm}

\paragraph{Grounding Text Descriptions to Images} HOI detection grounds the
semantics of subjects, objects, and interactions to image regions, which is
relevant to recent work on grounding text descriptions to images. Given an
image and its caption, Kong \textit{et al.} \cite{kong:cvpr2014} and Plummer
\textit{et al.} \cite{plummer:iccv2015} focus on localizing the mentioned
entities (e.g. nouns and pronouns) in the image. HOI detection, besides
grounding entities, i.e. people and objects, also grounds interactions to image
regions. Karpathy and Fei-Fei \cite{karpathy:cvpr2015} and Johnson \textit{et
al.} \cite{johnson:cvpr2016} address region-based captioning, which can be used
to generate HOI descriptions in image regions. However, they are unable to
localize individual persons and objects involved in the HOIs.

\section{HO-RCNN}
\label{sec:ho-rcnn}

\begin{figure}[t]
  \centering
  \includegraphics[width=1.00\linewidth]{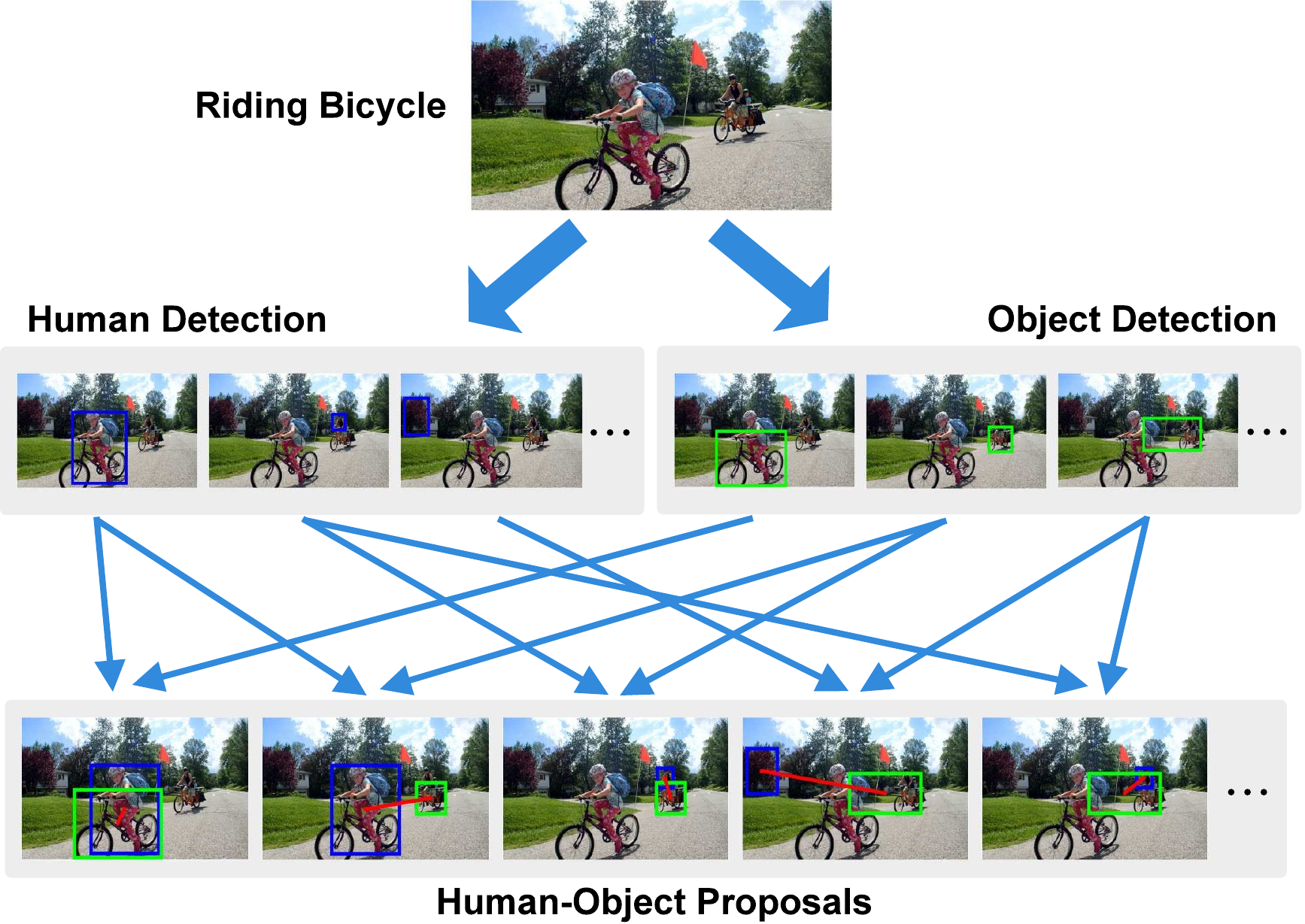}
  \vspace{-5mm}
  \caption{\small Generating human-object proposals from human and object detections.}
  \label{fig:proposal}
\end{figure}

\begin{figure*}[t]
  \centering
  \includegraphics[width=1.00\linewidth]{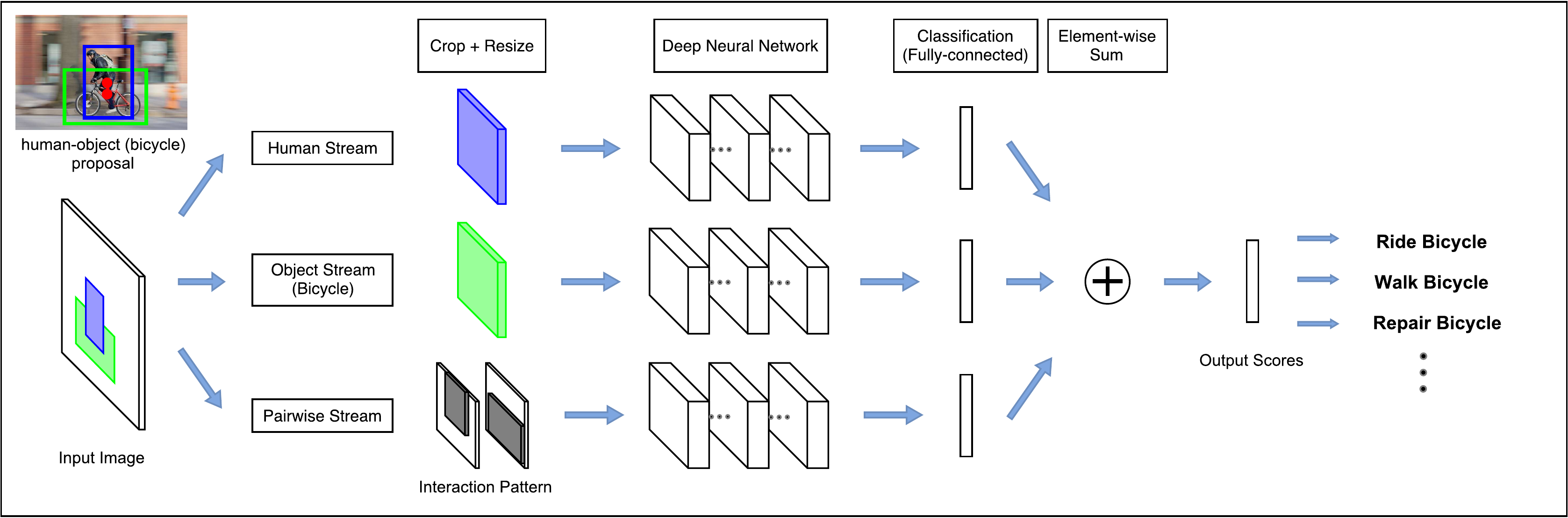}
  \caption{\small Multi-stream architecture of our HO-RCNN.}
  \label{fig:ho-rcnn}
\end{figure*}

Our HO-RCNN detects HOIs in two steps. First, we generate proposals of
human-object region pairs using state-of-the-art human and object detectors.
Second, each human-object proposal is passed into a ConvNet to generate HOI
classification scores. Our network adopts a multi-stream architecture to
extract features on the detected humans, objects, and human-object spatial
relations.

\vspace{-3mm}

\paragraph{Human-Object Proposals} We first generate proposals of human-object
region pairs. One naive way is to exploit a pool of class-agnostic bounding
boxes like other region-based object detection approaches
\cite{girshick:cvpr2014,girshick:iccv2015,ren:nips2015}. However, since each
proposal is a pairing between a human and object bounding box, the number of
proposals will be quadratic in the number of the candidate bounding boxes. To
ensure high recall, one usually needs to keep hundreds to thousands of
candidate bounding boxes, which results in more than tens of thousands of
human-object proposals. Classifying HOIs for all proposals will be intractable.
Instead, we assume having a list of HOI categories of interest (e.g. ``riding a
horse'', ``eating an apple'') beforehand, so we can first detect bounding boxes
for humans and the object categories of interest (e.g. ``horse'', ``apple'')
using state-of-the-art object detectors. We keep the bounding boxes with top
detection scores. For each HOI category (e.g. ``riding a horse''), the
proposals are then generated by pairing the detected humans and the detected
objects of interest (e.g. ``horse'') as illustrated in Fig.~\ref{fig:proposal}.

\vspace{-3mm}

\paragraph{Multi-stream Architecture} Given a human-object proposal, our
HO-RCNN classifies its HOIs using a multi-stream network
(Fig.~\ref{fig:ho-rcnn}), where different streams extract features from
different sources. To illustrate our idea, consider the classification of one
HOI class ``riding a bike''. Intuitively, local information around humans and
objects, such as human body poses and object local contexts, are critical in
distinguishing HOIs: A person riding a bike is more likely to be in a sitting
pose rather than standing; a bike being ridden by a person is more likely to be
occluded by the person's body in the upper region than those not being ridden.
In addition, human-object spatial relations are also important cues: The
position of a person is typically at the middle top of a bicycle when he is
riding it. Our multi-stream architecture is composed of three streams which
encode the above intuitions: (1) The \textit{human stream} extracts local
features from the detected humans. (2) The \textit{object stream} extracts
local features from the detected objects. (3) The \textit{pairwise stream}
extracts features which encode pairwise spatial relations between the detected
human and object. The last layer of each stream is a binary classifier that
outputs a confidence score for the HOI ``riding a bike''. The final confidence
score is obtained by summing the scores over all streams. To extend to mulitple
HOI classes, we train one binary classifier for each HOI class at the last
layer of each stream. The final score is summed over all streams separately for
each HOI class.

\vspace{-3mm}

\paragraph{Human and Object Stream} Given a human-object proposal, the human
stream extracts local features from the human bounding box, and generates
confidence scores for each HOI class. The full image is first cropped using the
bounding box and resized to a fixed size. This normalized image patch is then
passed into a ConvNet that extracts features through a seires of convolutional,
max pooling, and fully-connected layers. The last layer is a fully-connected
layer of size $K$, where $K$ is the number of HOI classes of interest, and each
output corresponds to the confidence score of one HOI class. The object stream
follows the same design except that the input is cropped and resized from the
object bounding box of the human-object proposal.

\begin{figure*}[t]
  \centering
  \includegraphics[width=1.00\linewidth]{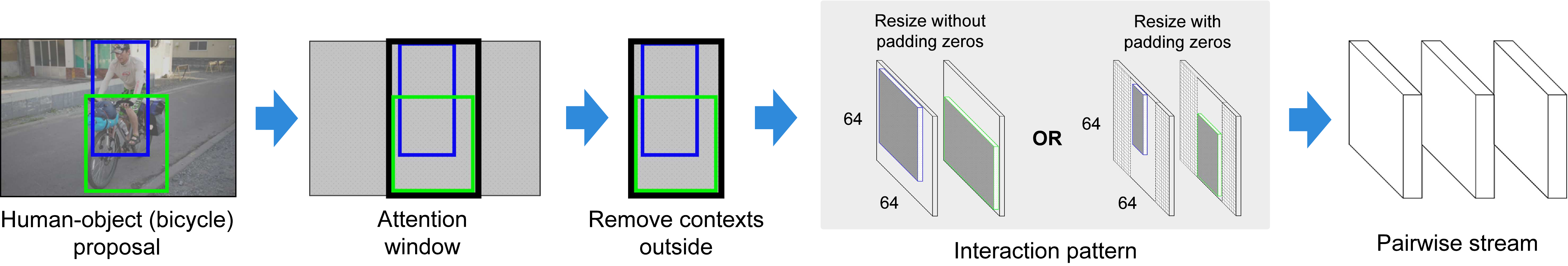}
  \caption{\small Construction of Interaction Patterns for the pairwise stream.}
  \label{fig:pairwise}
\end{figure*}

\vspace{-3mm}

\paragraph{Pairwise Stream} Given a human-object proposal, the pairwise stream
extracts features that encode the spatial relations between the human and
object, and generates a confidence score for each HOI class. Since the focus is
on spatial configurations of humans and objects, the input of this stream
should ignore pixel values and only exploit information of bounding box
locations. Instead of directly taking the bounding box coordinates as inputs,
we propose \textit{Interaction Patterns}, a special type of DNN input that
characterizes the relative location of two bounding boxes. Given a pair of
bounding boxes, its Interaction Pattern is a binary image with two channels:
The first channel has value $1$ at pixels enclosed by the first bounding box,
and value $0$ elsewhere; the second channel has value $1$ at pixels enclosed by
the second bounding box, and value $0$ elsewhere.~\footnote{In this work, we
apply the \textit{second-order} Interaction Pattern for learning pairwise
spatial relations. The Interaction Pattern can be extended to $n$-th order ($n
\in N$) by stacking additional images in the channel axis for learning
higher-order relations.} In our pairwise stream, the first channel corresponds
to the human bounding box and the second channel corresponds to the object
bounding box. Take the input image in Fig.~\ref{fig:ho-rcnn} as an example,
where the person is ``riding a bike''. The first (human) channel will have
value $1$ at the upper central region, while the second (object) channel will
have value $1$ at the lower central region. This representation enables DNN to
learn 2D filters that respond to similar 2D patterns of human-object spatial
configurations.

While the Interaction Patterns are able to characterize pairwise spatial
configurations, there are still two important details to work out. First, the
Interaction Patterns should be invariant to any joint translations of the
bounding box pair. In other words, the Interaction Patterns should be identical
for identitcal pair configurations whether the pair appears on the right or the
left side of the image. As a result, we remove all the pixels outside the
``attention window'', i.e. the tightest window enclosing the two bounding
boxes, from the Interaction Pattern. This makes the pairwise stream focus
solely on the local window containing the target bounding boxes and ignore
global translations. Second, the aspect ratio of Interaction Patterns may vary
depending on the attention window. This is problematic as DNNs take input of
fixed size (and aspect ratio). We propose two strategies to address this issue:
(1) We resize both sides of the Interaction Pattern to a fixed length
regardless of its aspect ratio. Note that this may change the aspect ratio of
the attention window. (2) We resize the longer side of the Interaction Pattern
to a fixed length while keeping the aspect ratio, followed by padding zeros on
both sides of the shorter side to achieve the fixed length. This normalizes the
size of the Interaction Pattern while keeping the aspect ratio of the attention
window. The construction of Interaction Patterns is illustrated in
Fig.~\ref{fig:pairwise}.

\vspace{-3mm}

\paragraph{Training with Multi-Label Classification Loss} Given a human-object
proposal, our HO-RCNN generates confidence scores for a list of HOI categories
of interest. As noted in \cite{chao:iccv2015}, a person can concurrently
perform different classes of actions to a target object, e.g. a person can be
``riding'' and ``holding'' a bicycle at the same time. Thus HOI recognition
should be treated as a mulit-label classification as opposed to the standard
$K$-way classification. As a result, we train the HO-RCNN by applying a sigmoid
cross entropy loss on the classification output of each HOI category, and
compute the total loss by summing over the individual losses.

\section{Constructing HICO-DET}

We contribute a new large-scale benchmark for HOI detection by augmenting HICO
\cite{chao:iccv2015} with instance annotations. HICO currently contains only
image-level annotations, i.e. 600 binary labels indicating the presence of the
600 HOI classes of interest (e.g. ``feeding a cat'', ``washing a knife''). We
further annotate the HOI instances present in each image, where each instance
is represented by a pairing between a human and object bounding box with a
class label (Fig.~\ref{fig:sample}).

\begin{figure*}[t]
  \centering
  \captionsetup[subfigure]{labelformat=empty}
  \begin{subfigure}[c]{0.30\textwidth}
    \centering
    \includegraphics[width=\textwidth]{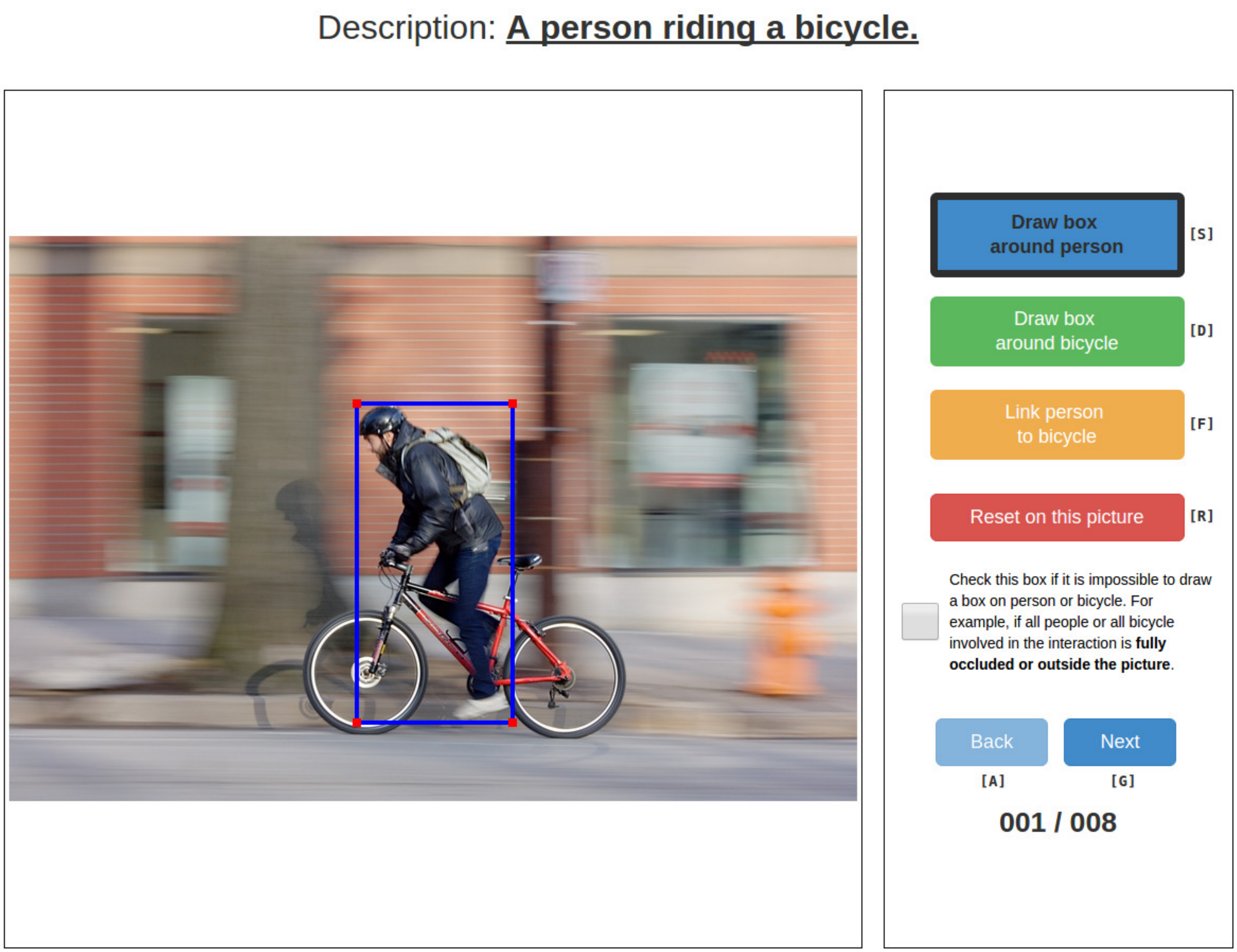}
    \caption{Step 1}
  \end{subfigure}
  ~
  \begin{subfigure}[c]{0.02\textwidth}
    \centering
    \includegraphics[width=\textwidth]{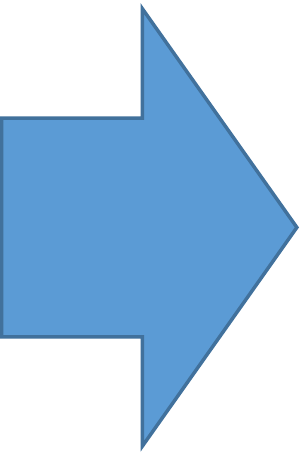}
  \end{subfigure}
  ~
  \begin{subfigure}[c]{0.30\textwidth}
    \centering
    \includegraphics[width=\textwidth]{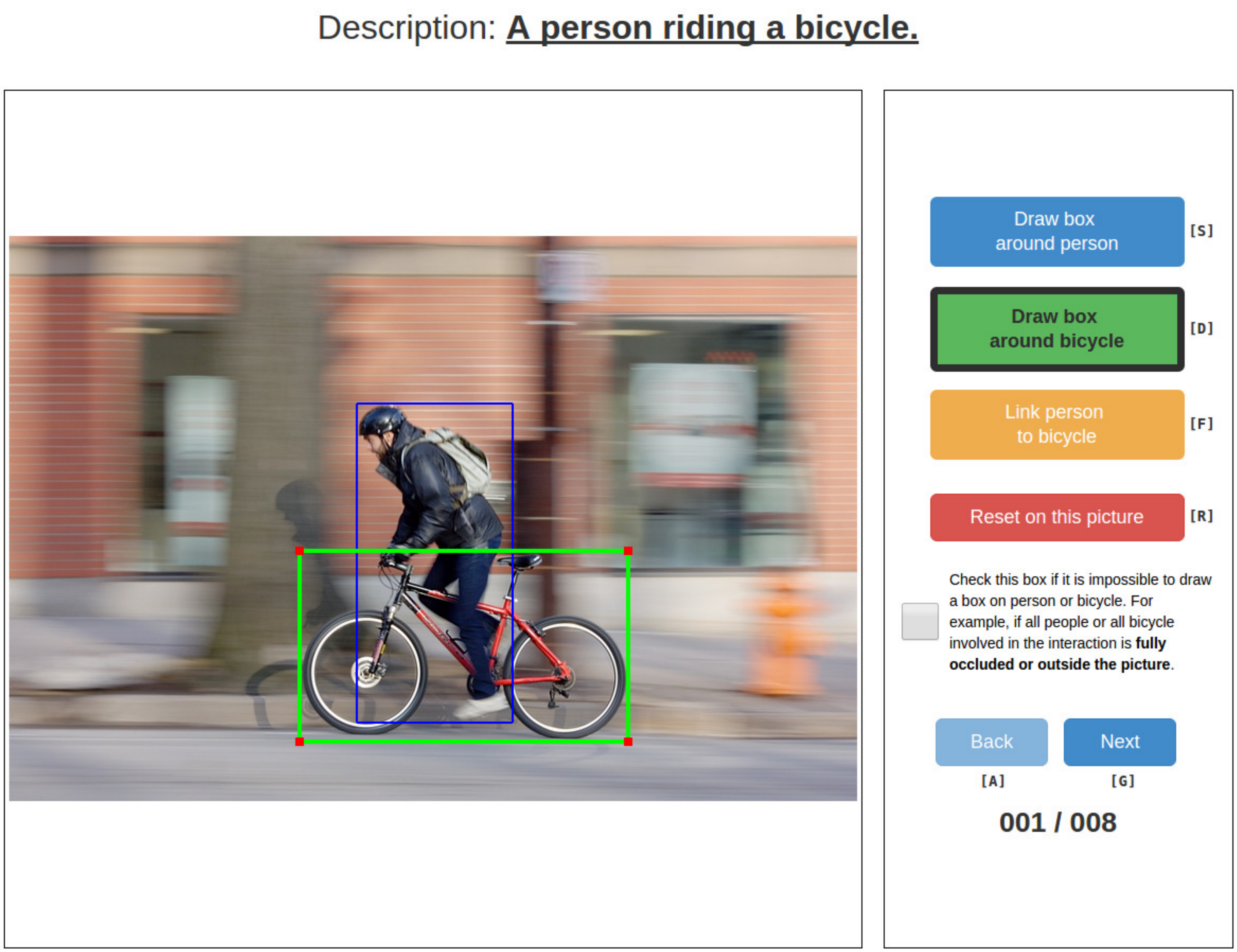}
    \caption{Step 2}
  \end{subfigure}
  ~
  \begin{subfigure}[c]{0.02\textwidth}
    \centering
    \includegraphics[width=\textwidth]{arrow.pdf}
  \end{subfigure}
  ~
  \begin{subfigure}[c]{0.30\textwidth}
    \centering
    \includegraphics[width=\textwidth]{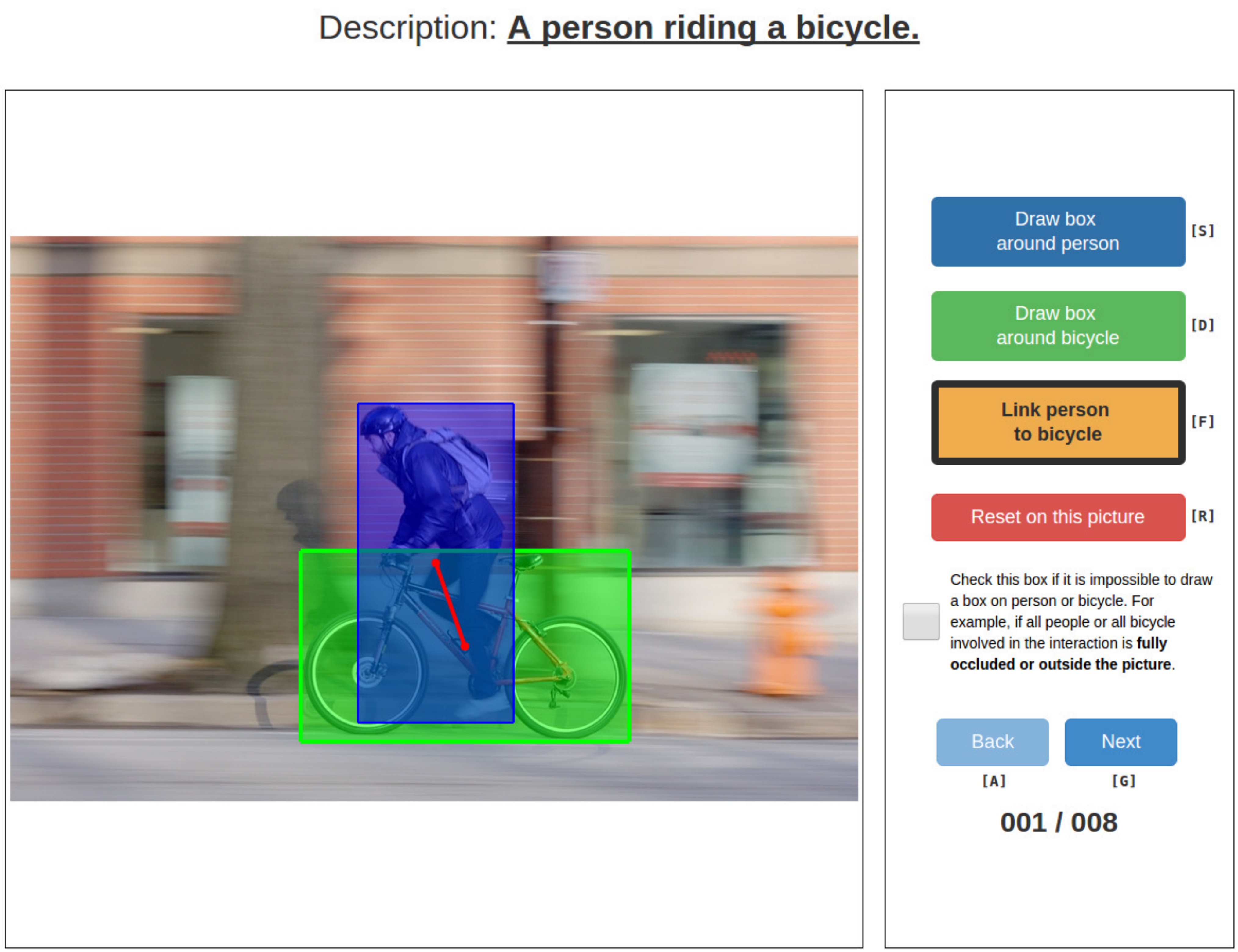}
    \caption{Step 3}
  \end{subfigure}
  \caption{\small Our data annotation task for each image involves three steps.}
  \vspace{-1mm}
  \label{fig:ui}
\end{figure*}

We collect human annotations by setting up annotation tasks on Amazon
Mechanical Turk (AMT). However, there are two key issues: First, given an image
and a presented HOI class (e.g. `` riding a bike''), the annotation task is not
as trivial as drawing bounding boxes around all the humans and objects
associated with the interaction (e.g. ``bike'') --- we also need to identify
the interacting relations, i.e. linking each person to the objects he is
interacting with. Second, although this linking step can be bypassed if the
annotator is allowed to draw only one human bounding box followed by one object
bounding box each time, such strategy is time intensive. Considering the cases
where there are multiple people interacting with one object (e.g. ``boarding an
airplane'' in Fig.~\ref{fig:sample}), or one person interacting with multiple
objects (e.g. ``herding cows'' in Fig.~\ref{fig:sample}), the annotator then
has to repeatedly draw bounding boxes around the shared persons and
objects.~\footnote{Although we formulate HOI detection as localizing
interactions between a single person and a single object, actual interactions
can be more complex such as the one-versus-many and many-versus-one cases.
However, these types of interactions can be decomposed into multiple instances
of person-object interaction pairs. Our goal is to detect all the decomposed
person-object pairs in such cases.} To efficiently collect such annotations, we
adopt a \textit{three-step} annotation procedure (Fig.~\ref{fig:ui}). For each
image, the annotator is presented with a sentence description, such as ``A
person riding a bicycle'', and asked to proceed with the following three steps:

\begin{figure}[t]
  \scriptsize
  \includegraphics[height=0.094\textwidth]{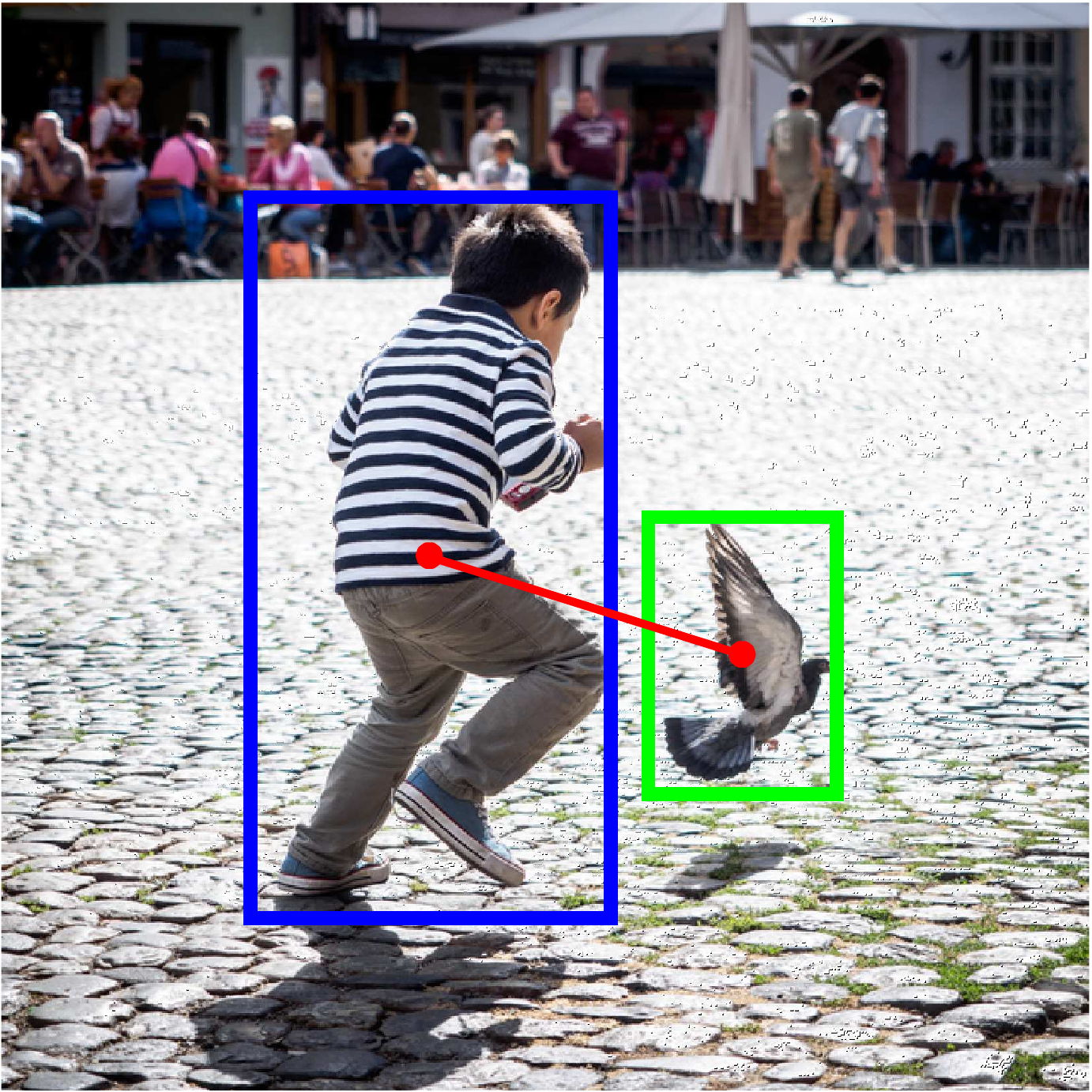}~
  \includegraphics[height=0.094\textwidth]{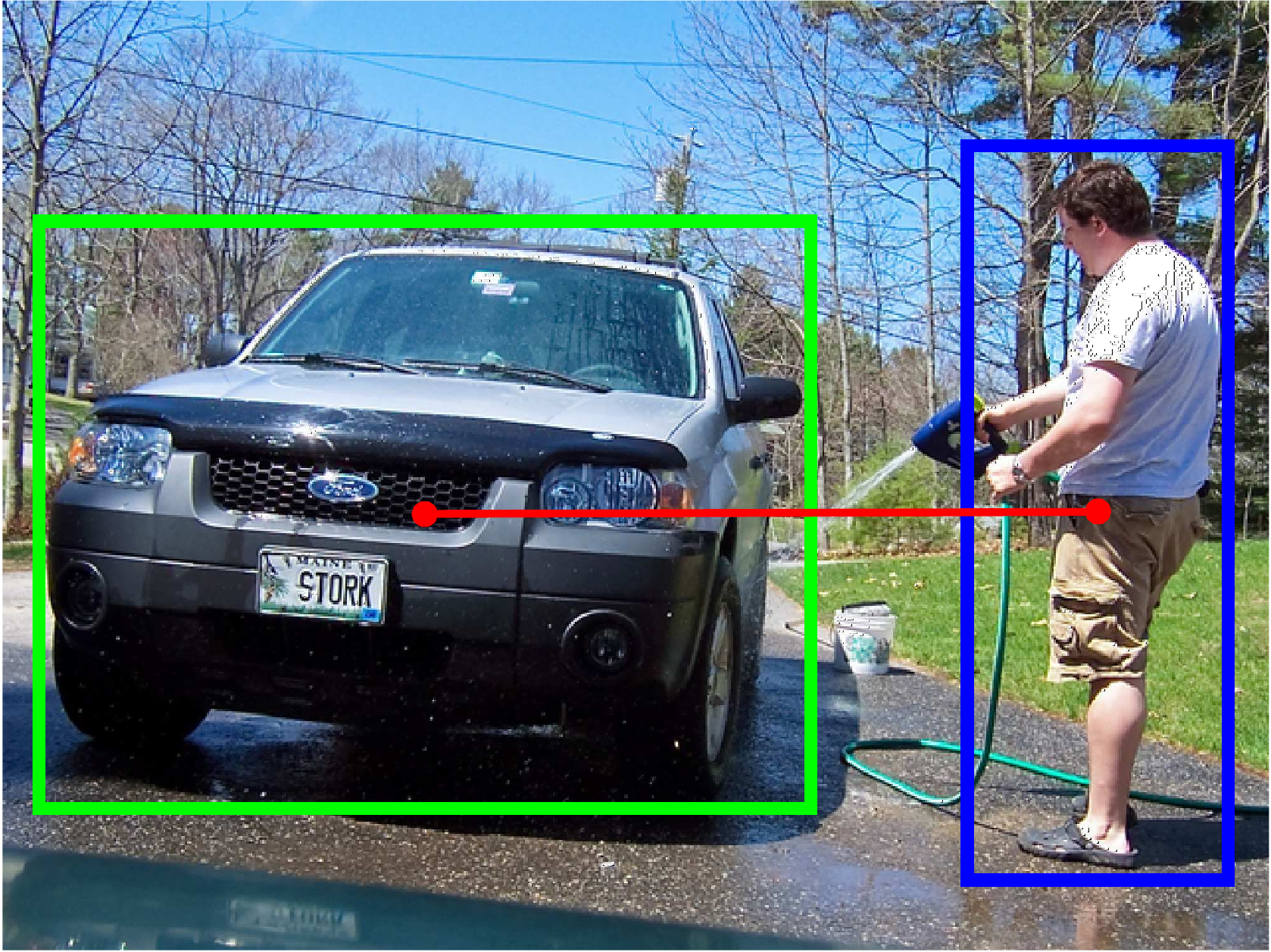}~
  \includegraphics[height=0.094\textwidth]{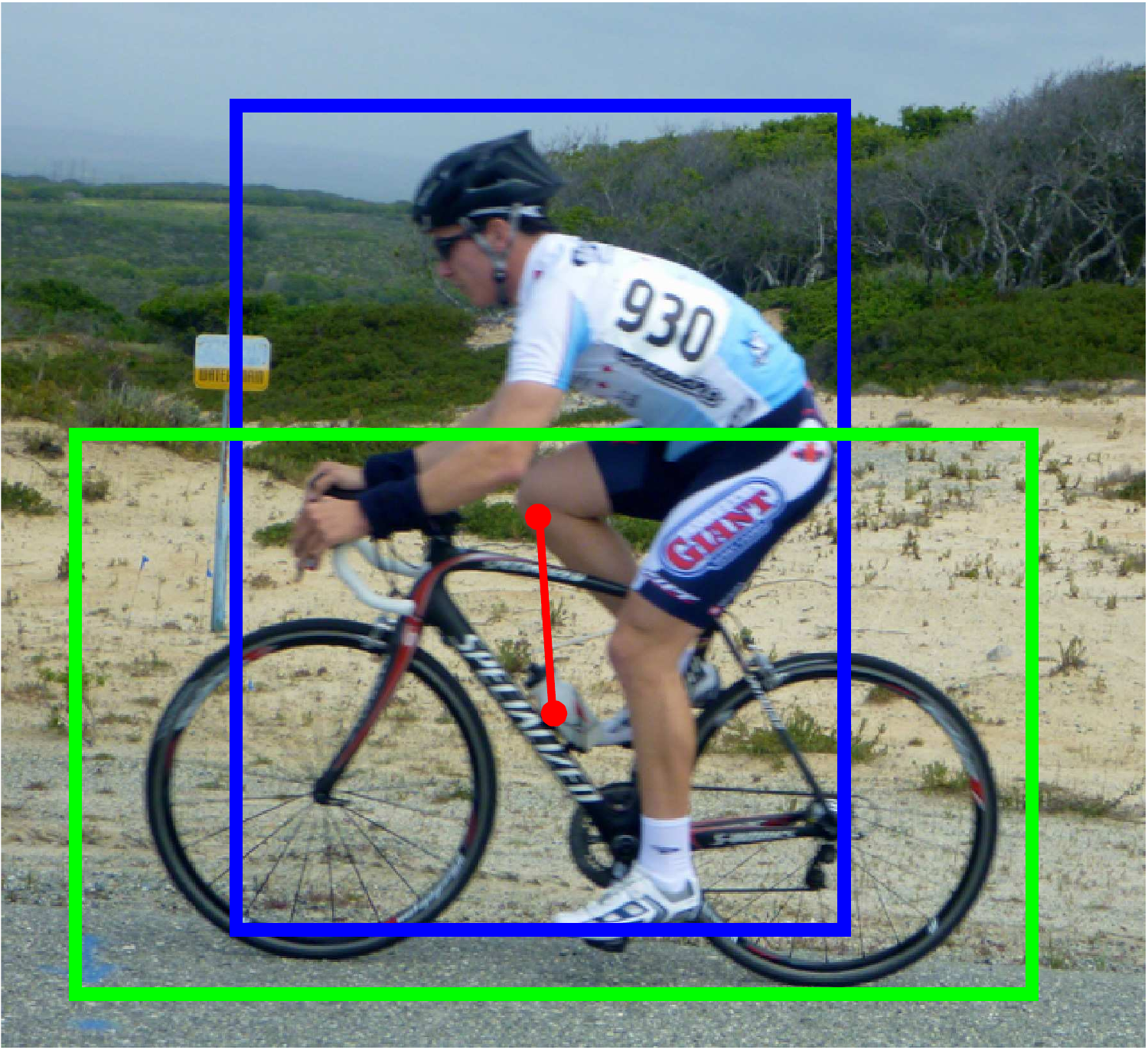}~
  \includegraphics[height=0.094\textwidth]{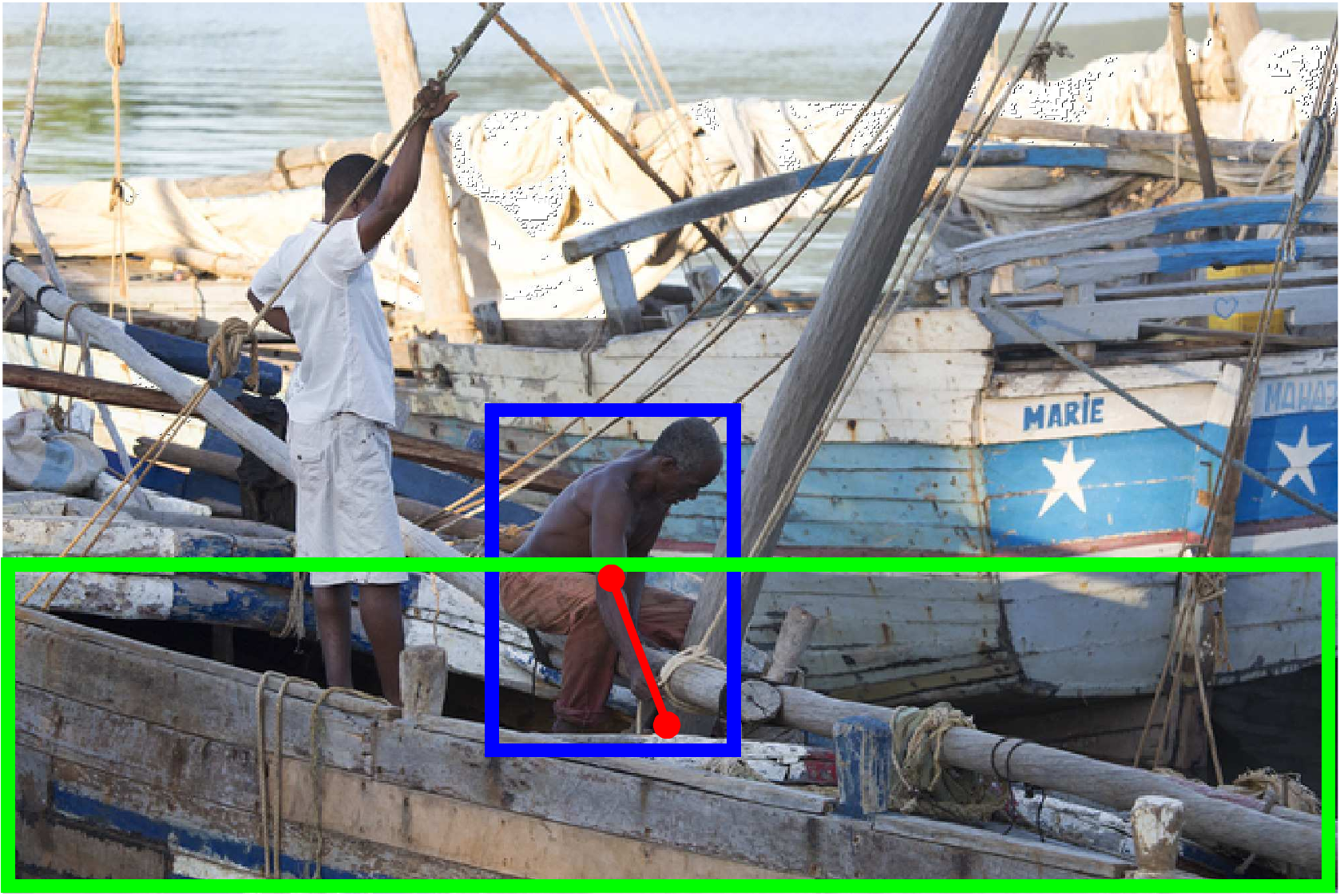}
  \\
  \vspace{1mm}~~chasing a bird~~~~~~~~~~~hosing a car~~~~~~~~~~~~~riding a bicycle~~~~~~~~~~~~~tying a boat
  \\
  \includegraphics[height=0.094\textwidth]{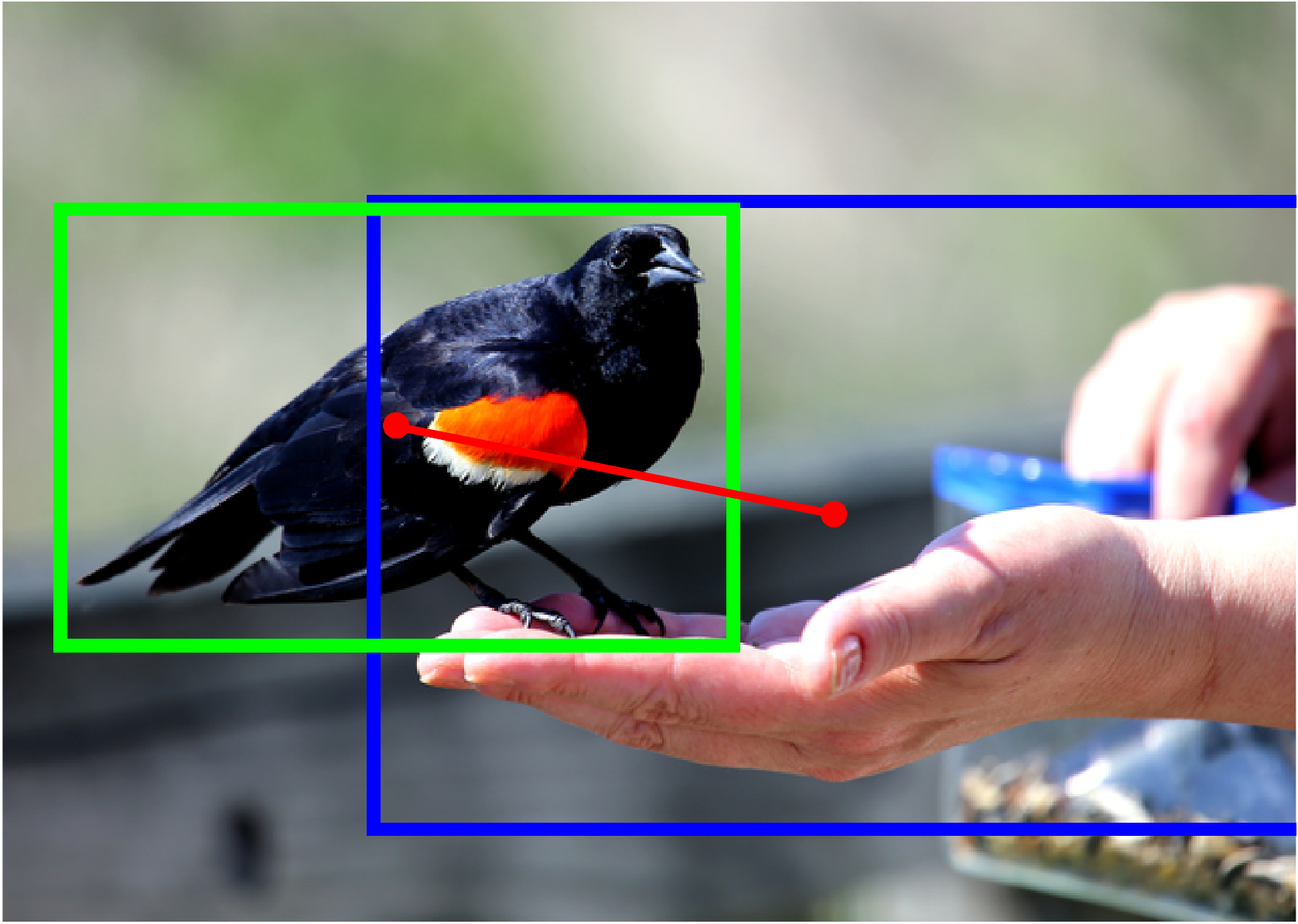}~
  \includegraphics[height=0.094\textwidth]{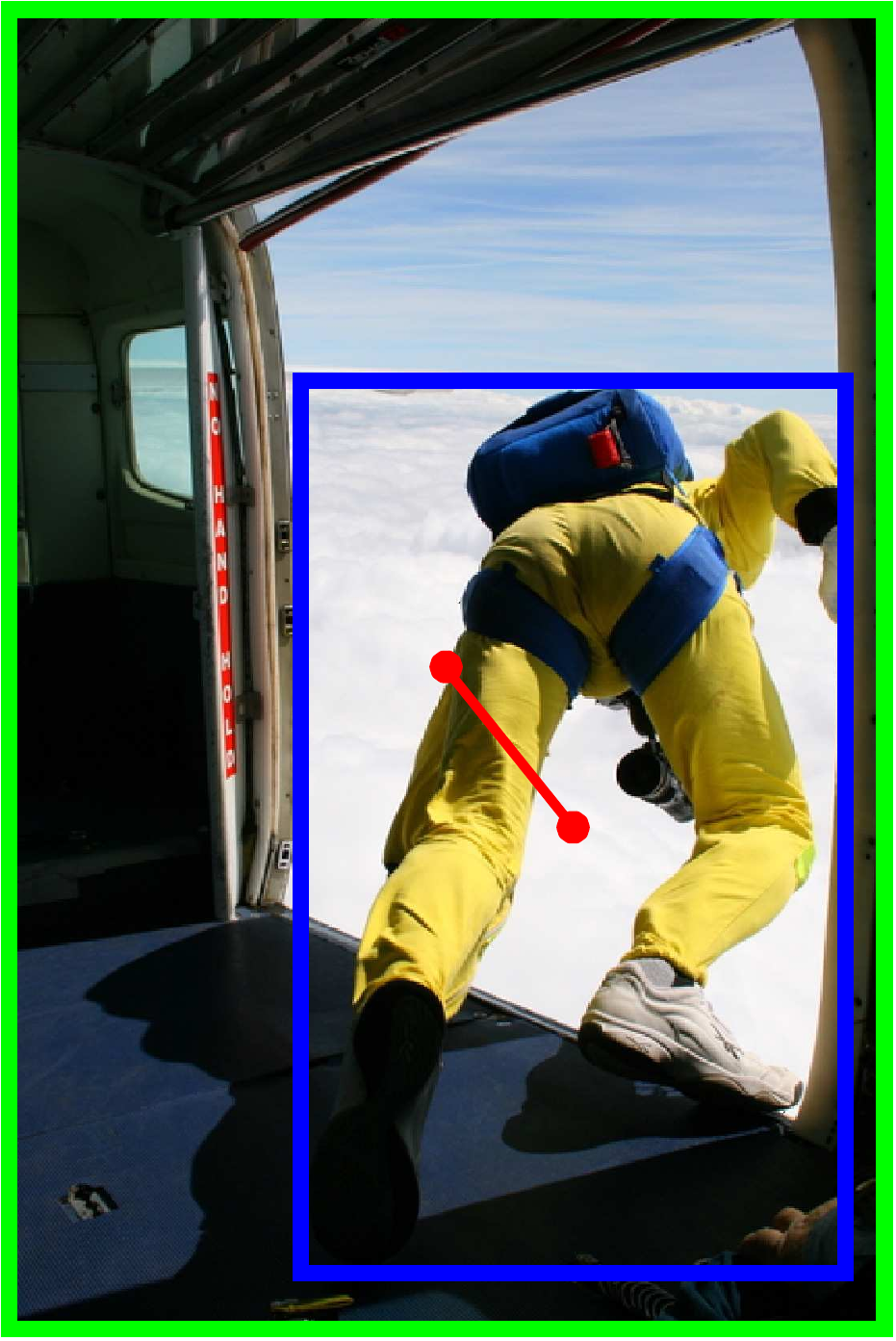}~
  \includegraphics[height=0.094\textwidth]{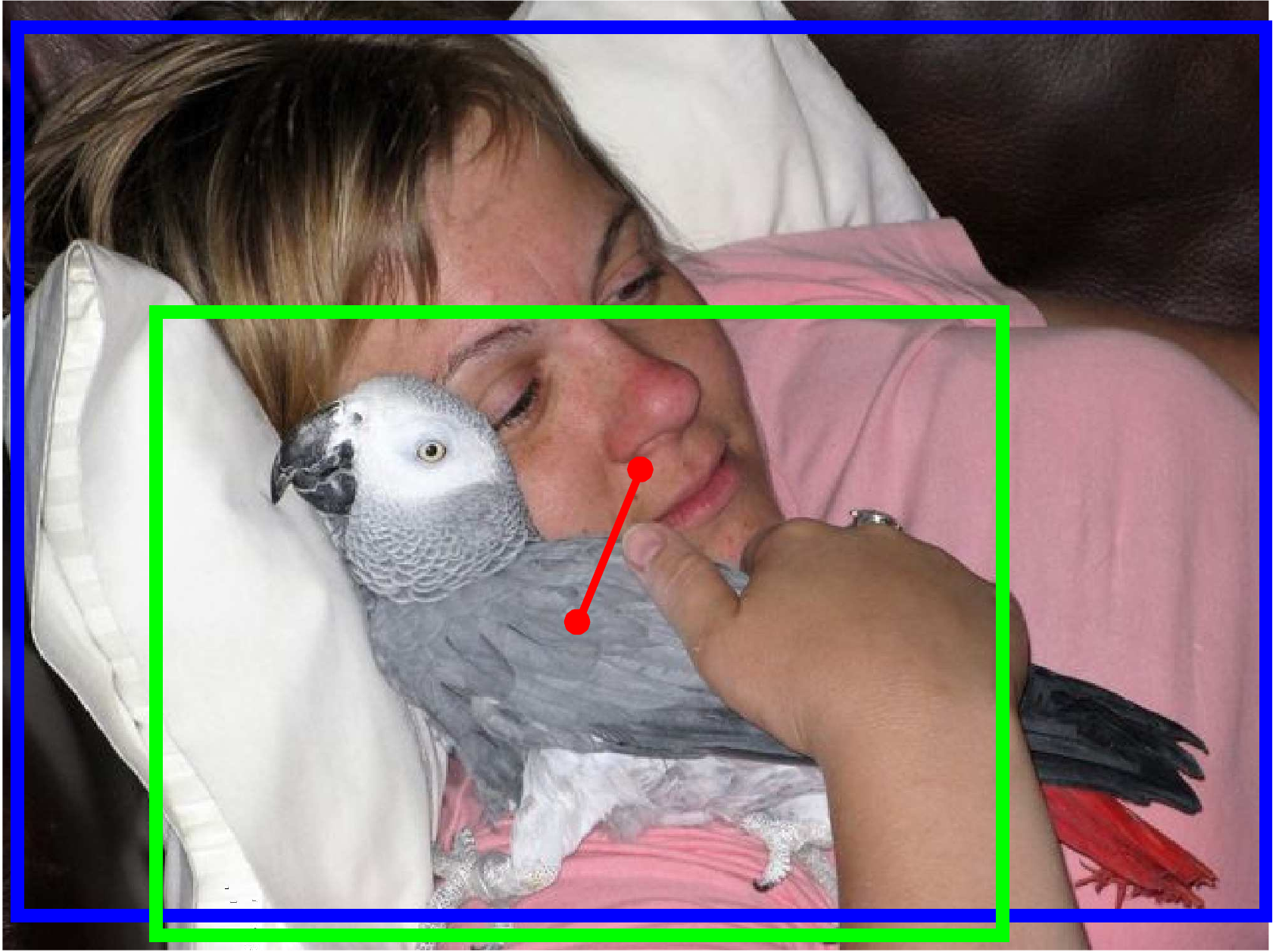}~
  \includegraphics[height=0.094\textwidth]{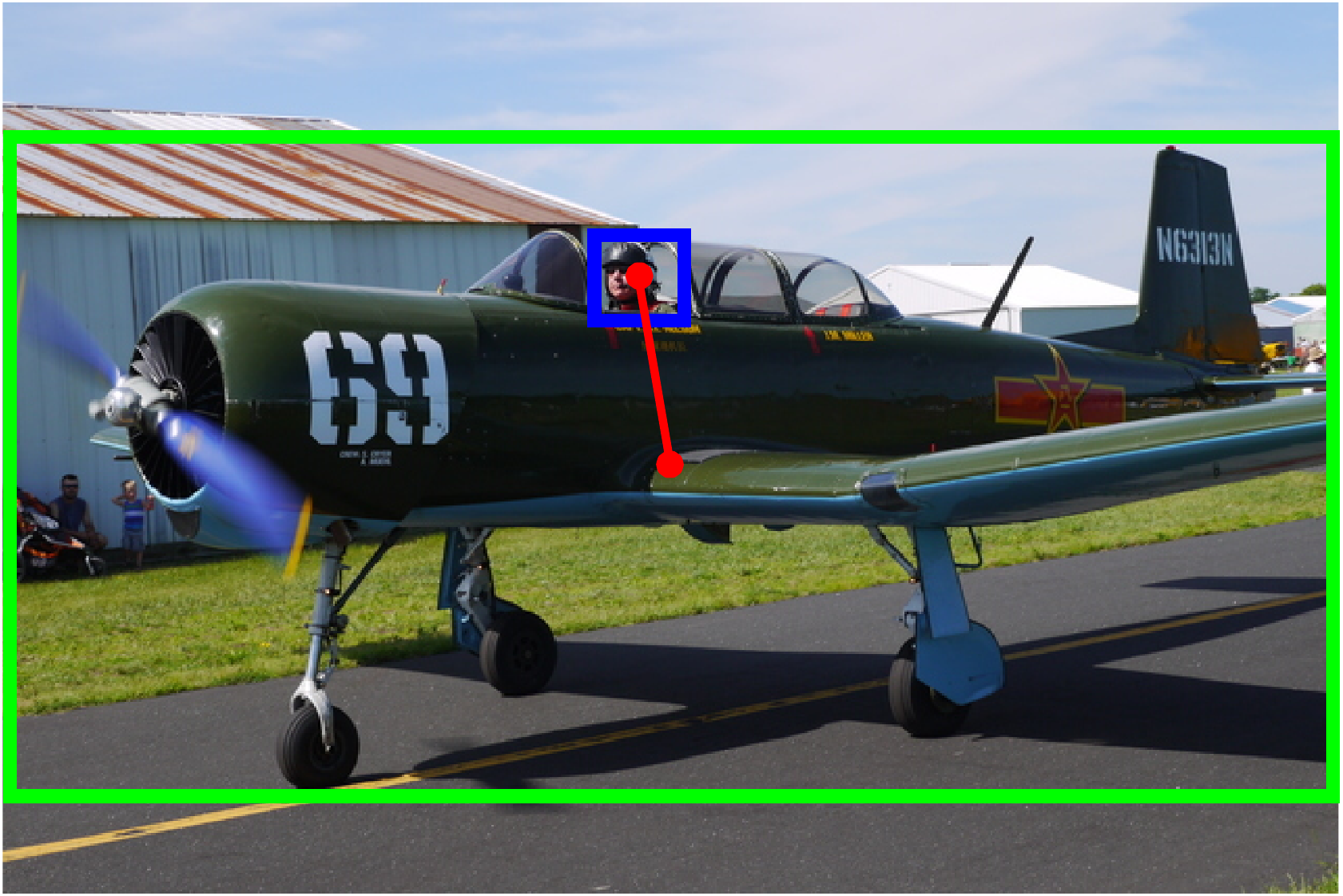}
  \\
  \vspace{1mm}~~~~~~~~feeding a bird~~~exiting an airplane~~~petting a bird~~~~~~~~~~~~~~riding an airplane
  \\
  \includegraphics[height=0.094\textwidth]{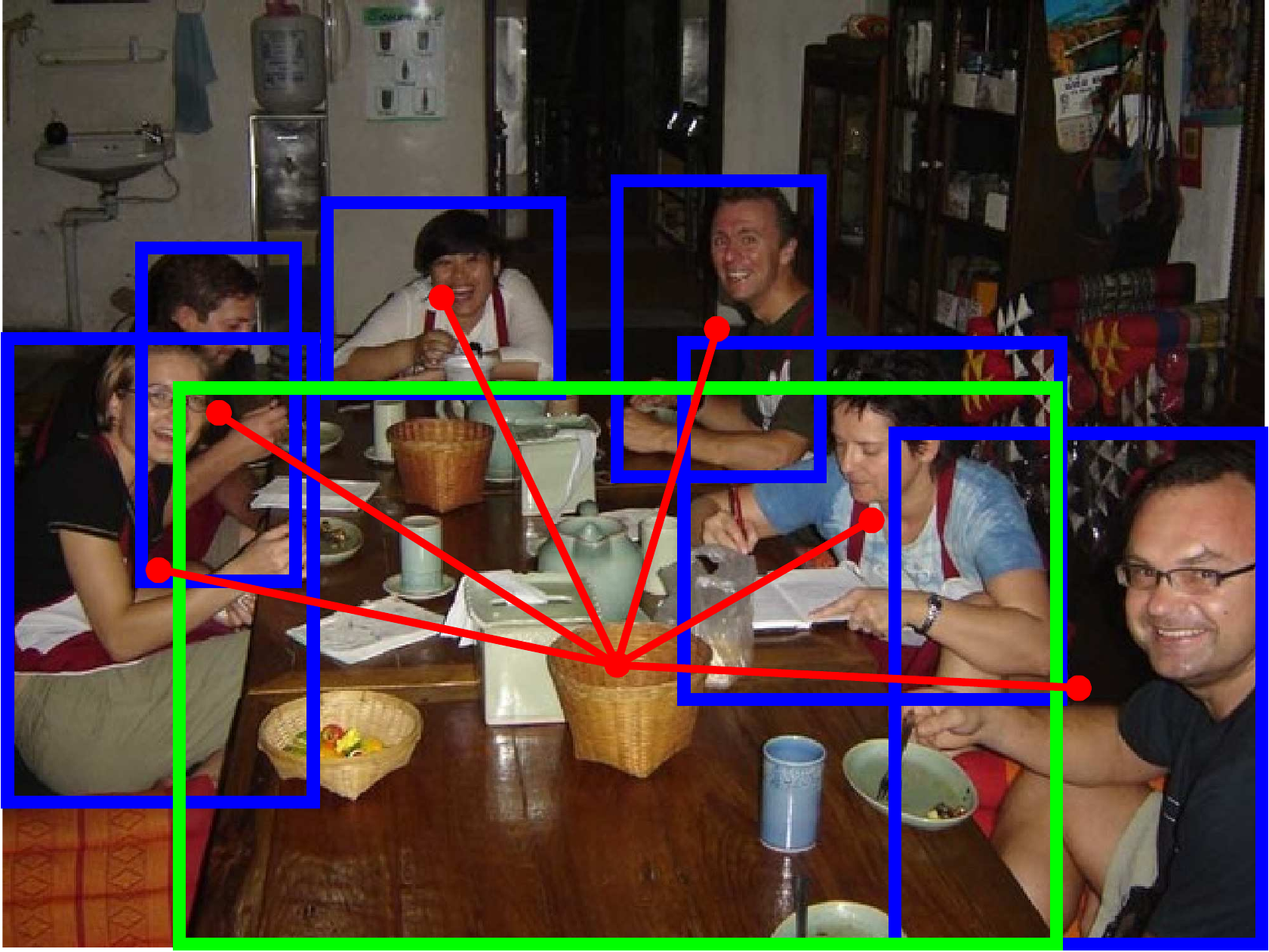}~
  \includegraphics[height=0.094\textwidth,trim={0.4cm 0.4cm 0.4cm 0.4cm},clip]{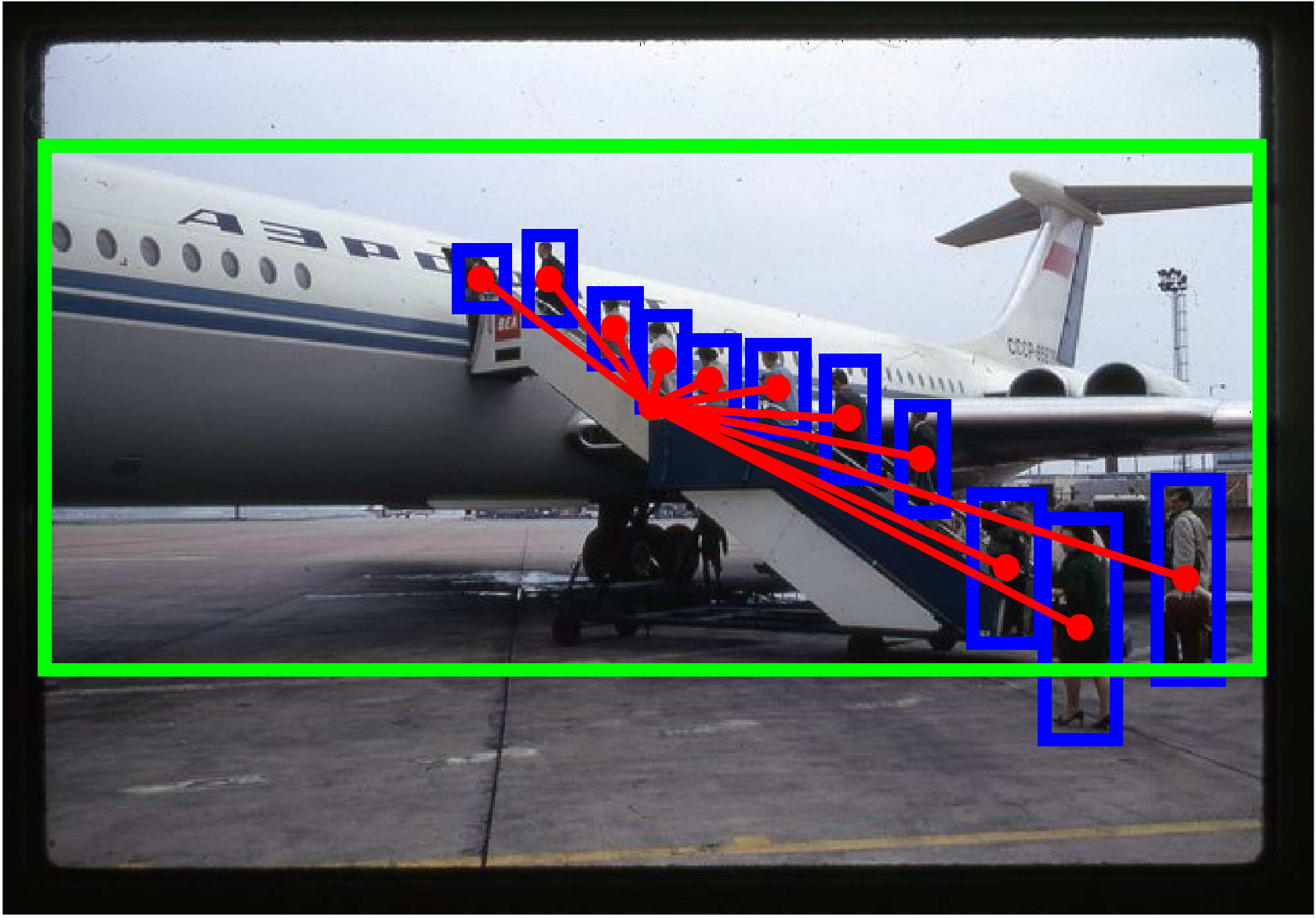}~
  \includegraphics[height=0.094\textwidth]{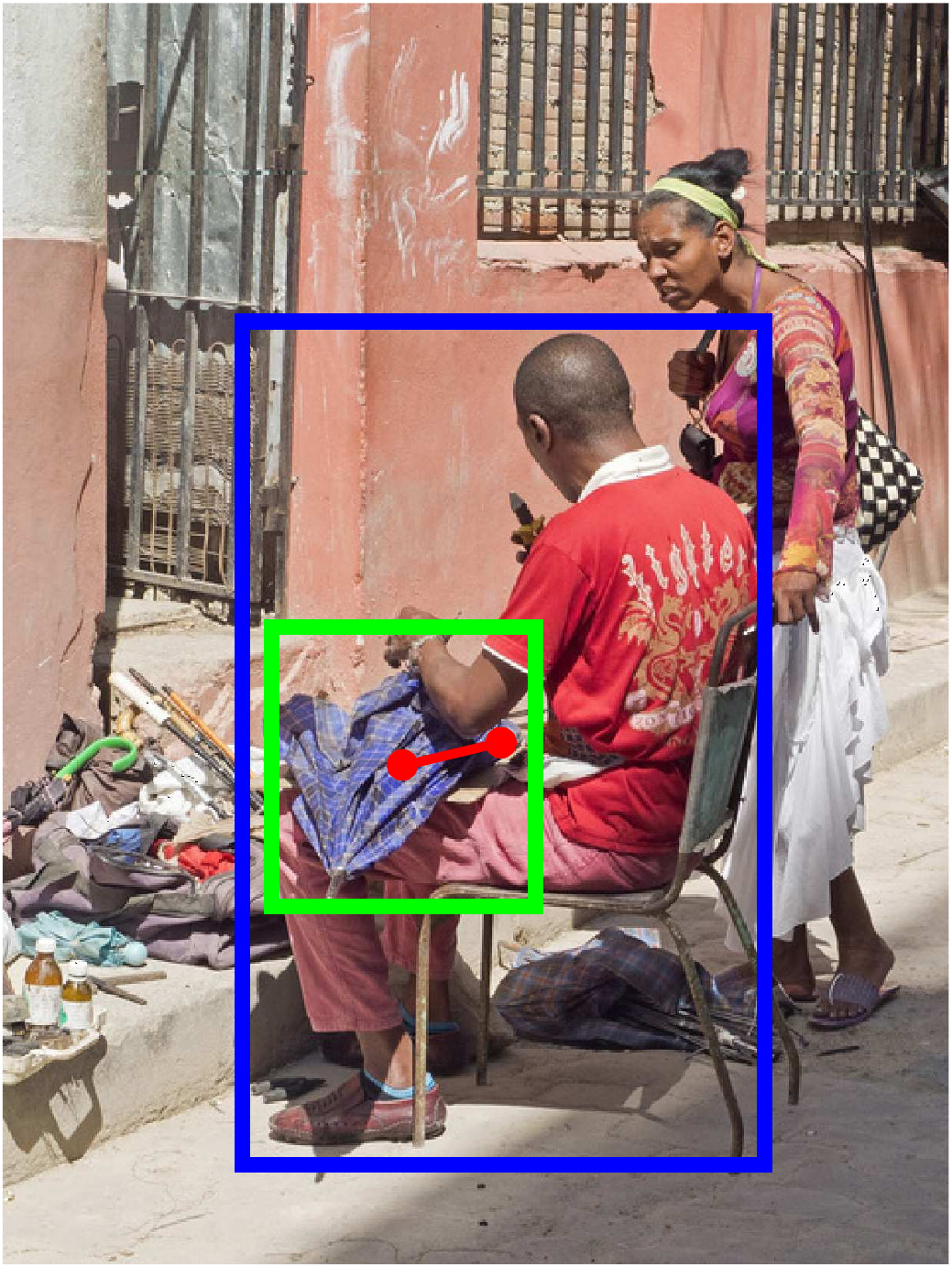}~
  \includegraphics[height=0.094\textwidth]{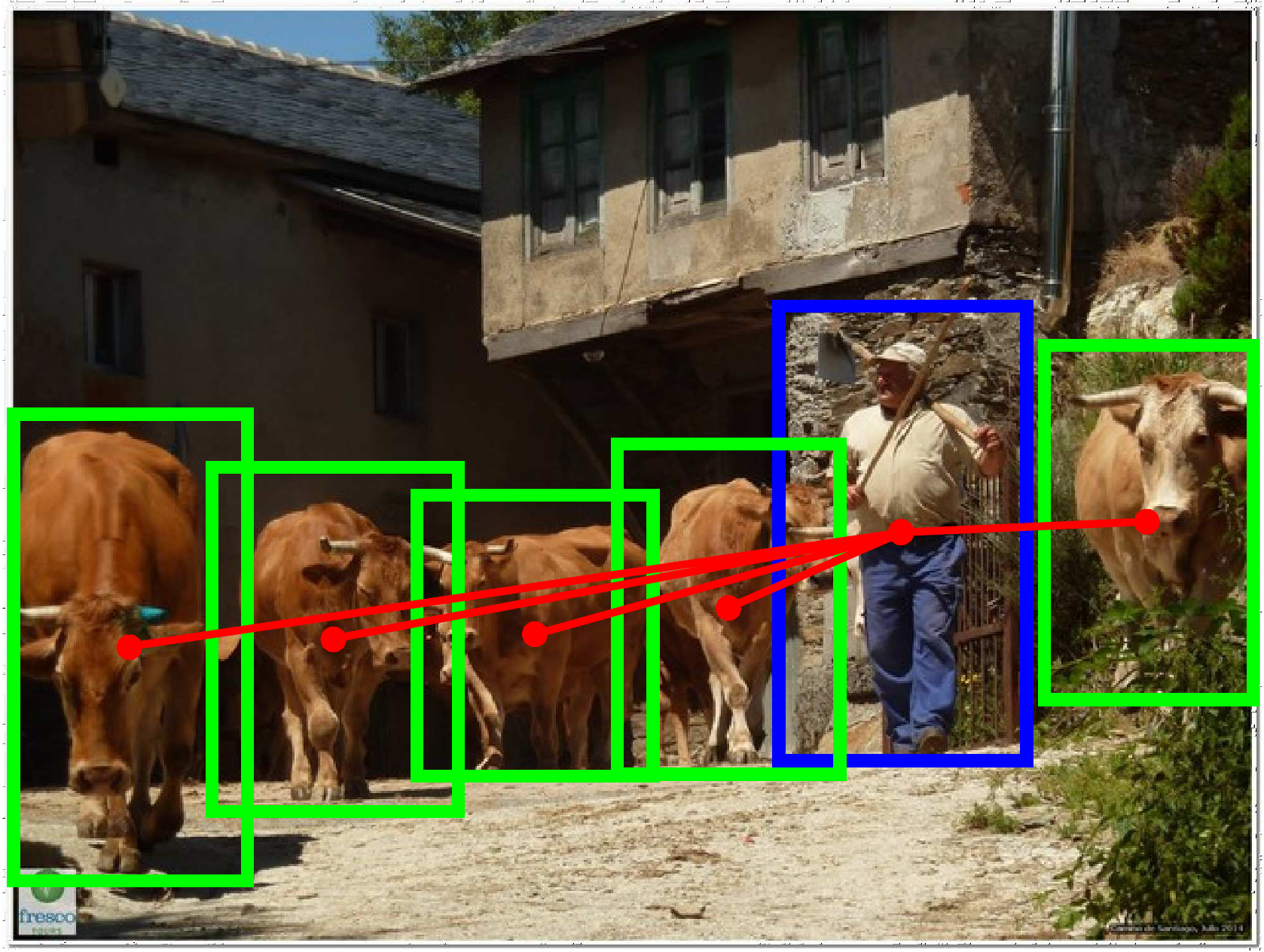}
  \\
  \vspace{1mm}eating at a dining table~~boarding an airplane~~repairing an umbrella~~herding cows
  \\
  \vspace{-4mm}
  \caption{\small Sample annotations of our HICO-DET.}
  \label{fig:sample}
\end{figure}

\vspace{-3mm}

\paragraph{Step 1: Draw a bounding box around each person.} The first step is
to draw bounding boxes around each person involved in the described interaction
(e.g. each person riding bicycles). Note that the annotators are explicitly
asked to ignore any person not involved in the described interaction, (e.g. any
person not riding a bicycle), since those people do not participate in any
instances of ``riding a bicycle''.

\vspace{-3mm}

\paragraph{Step 2: Draw a bounding box around each object.} The second step is
to draw bounding boxes around each object involved in the described
interaction, (e.g. each bicycle being ridden by someone). Similar to the first
step, the annotator should ignore any object that is not involved in the
described interaction (e.g. any bicycles not being ridden by someone).

\vspace{-3mm}

\paragraph{Step 3: Linking each person to objects.} The final step is to link a
person bounding box to an object bounding box if the described interaction is
taking place between them (e.g. link a person to a bicycle if the person is
riding the bicycle). Note that one person can be linked to multiple objects if
he is interacting with more than one objects (e.g. ``herding cows'' in
Fig.~\ref{fig:sample}), and one object can be linked with multiple people if it
is the case that more than one person are interacting with it (e.g. ``boarding
an airplane'' in Fig.~\ref{fig:sample}).

\begin{table}[t]
  \centering
  \footnotesize
  \setlength{\tabcolsep}{4.5pt}
  \begin{tabular}{|l|c|c|c|c|}
    \hline \TBstrut
              & \multicolumn{4}{c|}{HICO-DET} \\
    \cline{2-5} \TBstrut
              & \#image & \#positive & \#instance         & \#bounding box     \\
    \hline \Tstrut
    Train     & 38118   & 70373      & 117871 ~(1.67/pos) & 199733 ~(2.84/pos) \\ \Bstrut
    Test      &~~9658   & 20268      &~~33405 ~(1.65/pos) &~~56939 ~(2.81/pos) \\
    \hline \TBstrut
    Total     & 47776   & 90641      & 151276 ~(1.67/pos) & 256672 ~(2.83/pos) \\
    \hline
  \end{tabular}
  \vspace{-2mm}
  \caption{\small Statistics of annotations in our HICO-DET.}
  \label{tab:stats}
\end{table}

\vspace{3mm}

Note that in some rare cases, the involved person or object may be invisible,
even though the presence of the HOI can be inferred from the image. (e.g. We
can tell a person is ``sitting on a chair'' although the chair is
fully-occluded by the person's body.) If the involved person or object is
completely invisible in the image, the annotator is asked to mark those images
as ``invisible''. Among all 90641 annotation tasks (each corresponds to one
positive HOI label for one image in HICO), we found that there are 1209
(1.33\%) tasks labeled as ``invisible''. Since our instance annotations are
built upon HICO's HOI class annotations, our HICO-DET also has a long-tail
distribution in the number of instances per HOI class as in HICO. By keeping
the same training-test split, we found that there are 2 out of 600 classes
(``jumping a car'' and ``repairing a mouse'') which have no training instances
due to the invisibility of people or objects. As a result, we added 2 new
images to our HICO-DET so we have at least one training instance for each of
the 600 HOI classes. Tab.~\ref{tab:stats} shows the statistics of the newly
collected annotations. We see that each image in HICO-DET has on average more
than one (1.67) instance for each positive HOI label. Note that the total
number of bounding boxes (256672) is less than twice the total number of
instances (151274). This is because different instances can share people or
objects, as shown in Fig.~\ref{fig:sample}.

\section{Experiments}

\paragraph{Evaluation Setup} Following the standard evaluation metric for
object detection, we evaluate HOI detection using mean average precision (mAP).
In object detection, a detected bounding box is assigned a true positive if it
overlaps with a ground truth bounding box of the same class with intersection
over union (IoU) greater than $0.5$. Since we predict one human and one object
bounding box in HOI detection, we declare a true positive if the minimum of
human overlap $\text{IoU}_{h}$ and object overlap $\text{IoU}_o$ exceeds 0.5,
i.e. $\min(\text{IoU}_h,\text{IoU}_o) > 0.5$. We report the mean AP over three
different HOI category sets: (a) all 600 HOI categories in HICO (Full), (b) 138
HOI categories with less than 10 training instances (Rare), and (c) 462 HOI
categories with 10 or more training instances (Non-Rare). All reported results
are evaluated on the test set.

Following the HICO classification benchmark \cite{chao:iccv2015}, we also
consider two different evaluation settings: (1) \textit{Known Object} setting:
For each HOI category (e.g. ``riding a bike''), we evaluate the detection only
on the images containing the target object category (e.g. ``bike''). The
challenge is to localize HOI (e.g. human-bike pairs) as well as distinguishing
the interaction (e.g. ``riding''). (2) \textit{Default} setting: For each HOI
category, we evaluate the detection on the full test set, including images both
containing and not containing the target object category. This is a more
challenging setting as we also need to distinguish background images (e.g.
images without ``bike'').

\begin{table}[t]
  \centering
  \begin{subtable}{\linewidth}
    \centering
    \footnotesize
    \setlength{\tabcolsep}{5.45pt}
    \begin{tabular}{|l||C{0.58cm}|C{0.58cm}|C{0.58cm}||C{0.64cm}|C{0.58cm}|C{0.64cm}|}
      \hline \TBstrut
                     & \multicolumn{3}{c||}{Default}                 & \multicolumn{3}{c|}{Known Object}               \\
      \cline{2-7} \TBstrut
                     & Full          & Rare          & Non-Rare      & Full           & Rare          & Non-Rare       \\
      \hline \TBstrut
      HO             & 5.73          & 3.21          & 6.48          & 8.46           & 7.53          & 8.74           \\
      \hline \Tstrut
      HO+vec0 (fc)   & 6.47          & 3.57          & 7.34          & 9.32           & 8.19          & 9.65           \\
      HO+vec1 (fc)   & 6.24          & 3.59          & 7.03          & 9.13           & 8.09          & 9.45           \\
      HO+IP0 (fc)    & 7.07          & 4.06          & 7.97          & 10.10          & 8.38          & 10.61          \\
      HO+IP1 (fc)    & 6.93          & 3.91          & 7.84          & 10.07          & 8.43          & 10.56          \\
      HO+IP0 (conv)  & 7.15          & 4.47          & 7.95          & 10.23          & 8.85          & 10.64          \\ \Bstrut
      HO+IP1 (conv)  & \textbf{7.30} & \textbf{4.68} & \textbf{8.08} & \textbf{10.37} & \textbf{9.06} & \textbf{10.76} \\
      \hline
    \end{tabular}
  \end{subtable}
  \\
  \vspace{2mm}
  \begin{subtable}{\linewidth}
    \centering
    \setlength{\tabcolsep}{5.45pt}
    \fontsize{7.5}{10}\selectfont
    \begin{tabular}{|l||C{1.022cm}|C{1.022cm}|C{1.022cm}|}
      \hline \TBstrut
                          & \multicolumn{3}{c|}{Default}             \\
      \cline{2-4} \TBstrut
                                      & Full     & Rare   & Non-Rare \\
      \hline \Tstrut
      HO+vec1 (fc) vs. HO             & $<0.001$ & 0.132  & $<0.001$ \\
      HO+IP1 (conv) vs. HO            & $<0.001$ & 0.001  & $<0.001$ \\ \Bstrut
      HO+IP1 (conv) vs. HO+vec1 (fc)  & $<0.001$ & 0.001  & $<0.001$ \\
      \hline
    \end{tabular}
    \begin{tabular}{|l||C{1.022cm}|C{1.022cm}|C{1.022cm}|}
      \hline \TBstrut
                          & \multicolumn{3}{c|}{Known Object}        \\
      \cline{2-4} \TBstrut
                                      & Full    & Rare    & Non-Rare \\
      \hline \Tstrut
      HO+vec1 (fc) vs. HO             & $<0.001$ & 0.077  & $<0.001$ \\
      HO+IP1 (conv) vs. HO            & $<0.001$ & 0.005  & $<0.001$ \\ \Bstrut
      HO+IP1 (conv) vs. HO+vec1 (fc)  & $<0.001$ & 0.049  & $<0.001$ \\
      \hline
    \end{tabular}
  \end{subtable}
  \vspace{-2mm}
  \caption{\small Performace comparison of difference pairwise stream variants.
Top: mAP (\%). Bottom: p-value for the paired t-test.}
  \label{tab:pairwise}
\end{table}

\vspace{-3mm}

\paragraph{Training HO-RCNN} We first generate human-object proposals using
state-of-the-art object detectors. Since HICO and MS-COCO \cite{lin:eccv2014}
share the same 80 object categories, we train 80 object detectors using
Fast-RCNN \cite{girshick:iccv2015} on the MS-COCO training set. As detailed in
Sec.~\ref{sec:ho-rcnn}, we generate proposals for each HOI category (e.g.
``riding a bike'') by pairing the top detected humans and objects (e.g.
``bike'') in each image. In our experiments, we adopt the top 10 detections for
human and each object category, resulting in 100 proposals per object category
per image.

We implement our HO-RCNN using Caffe \cite{jia2014caffe}. For both the human
and object streams, we adopt the CaffeNet architecture with weights pre-trained
on the ImageNet classification task \cite{russakovsky:ijcv2015}. To train on
HICO-DET, we run SGD with a global learning rate 0.001 for 100k iterations, and
then lower the learning rate to 0.0001 and run for another 50k iterations. We
use an \textit{image-centric} sampling strategy similar to
\cite{girshick:iccv2015} for mini-batch sampling: Each mini-batch of size 64 is
constructed from 8 randomly sampled images, with 8 randomly sampled proposals
for each image. These 8 proposals are from three different sources. Suppose a
sampled image contains interactions with ``bike'', we sample: (a) 1
\textit{positive example}: human-bike proposals that have
$\min(\text{IoU}_h,\text{IoU}_o) \geq 0.5$ with at least one ground-truth
instance from a category involving ``bike''. (b) 3 \textit{type-I negatives}:
non-positve human-bike proposals that have $\min(\text{IoU}_h,\text{IoU}_o) \in
[0.1,0.5)$ with at least one ground-truth instance from a category involving
``bike''. (c) 4 \textit{type-II negatives}: proposals that do not involve
``bike''.

\vspace{-3mm}

\paragraph{Ablation Study} We first perform an ablation study on the pairwise
stream. We consider the two different variants of the Interaction Patterns
described in Sec.~\ref{sec:ho-rcnn}, i.e. without padding (IP0) and with
padding (IP1), each paired with two different DNN architectures: a
fully-connected network (fc) and a convolutional network (conv).~\footnote{The
architecture is detailed in the supplementary material.} We also report
baselines that use the same fc architecture but take the 2D vector from human's
center to object's center (vec0: without padding, vec1: with padding).
Tab.~\ref{tab:pairwise} (top) reports the mAP of using the human and object
stream alone (HO) as well as combined with different pairwise streams (vec0
(fc), vec1 (fc), IP0 (fc), IP1 (fc), IP0 (conv), IP1 (conv)). Note that for all
the methods, the Default setting has lower mAPs than the Known Object setting
due to the increasing challenge in the test set, and the rare categories have
lower mAP than the none-rare categories due to sparse training examples.
Although the mAPs are low overall (i.e. below 11\%), we still observe in both
settings that adding a pairwise stream improves the mAP. Among all pairwise
streams, using Interaction Patterns with the conv architecture achieves the
highest mAP (e.g. for IP1 (conv) on the full dataset, 7.30\% in the Default
setting and 10.37\% in the Known Object setting). To demonstrate the
signficance of the improvements, we perform \textit{paired t-test}: We compare
two methods by their AP difference on each HOI category. The null hypothesis is
that the mean of the AP differences over the categories is zero. We show the
p-values in Tab.~\ref{tab:pairwise} (bottom). While the 2D vector baselines
outperform the HO baseline in mAP, the p-value is above 0.05 on rare categories
(e.g. 0.13 for ``HO+vec1 (fc) vs. HO'' in the Default setting). On the other
hand, ``HO+IP1 (conv) vs. HO'' and ``HO+IP1 (conv) vs. HO+vec1 (fc)'' both have
all p-values below 0.05, suggesting that using Interaction Patterns with the
conv architecture has a significant improvement not only over the HO baseline,
but also over the 2D vector baseline.

\begin{figure}[t]
  \centering
  \captionsetup[subfigure]{labelformat=empty}
  \begin{subfigure}[c]{0.31\linewidth}
    \centering
    \includegraphics[width=0.48\textwidth]{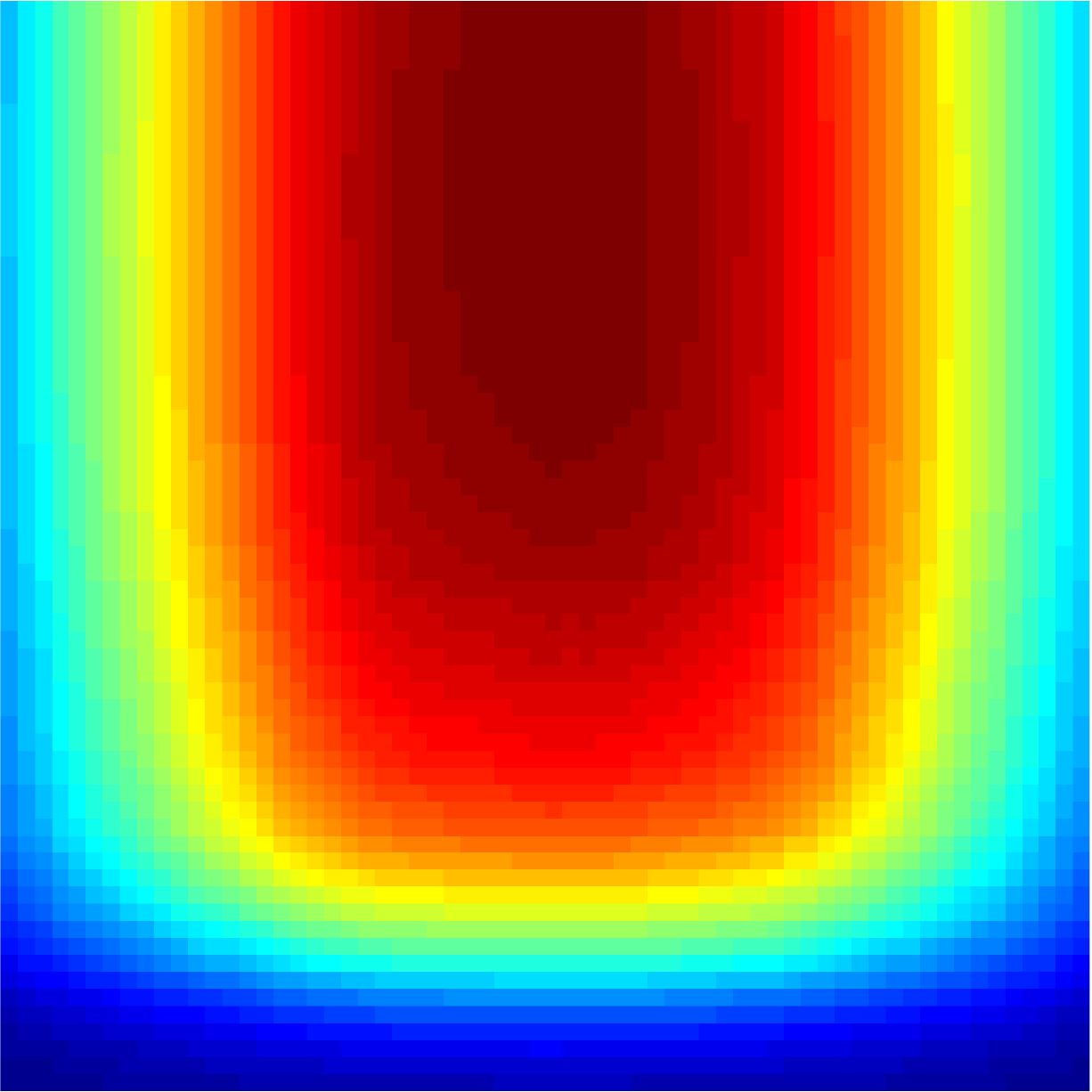}
    \includegraphics[width=0.48\textwidth]{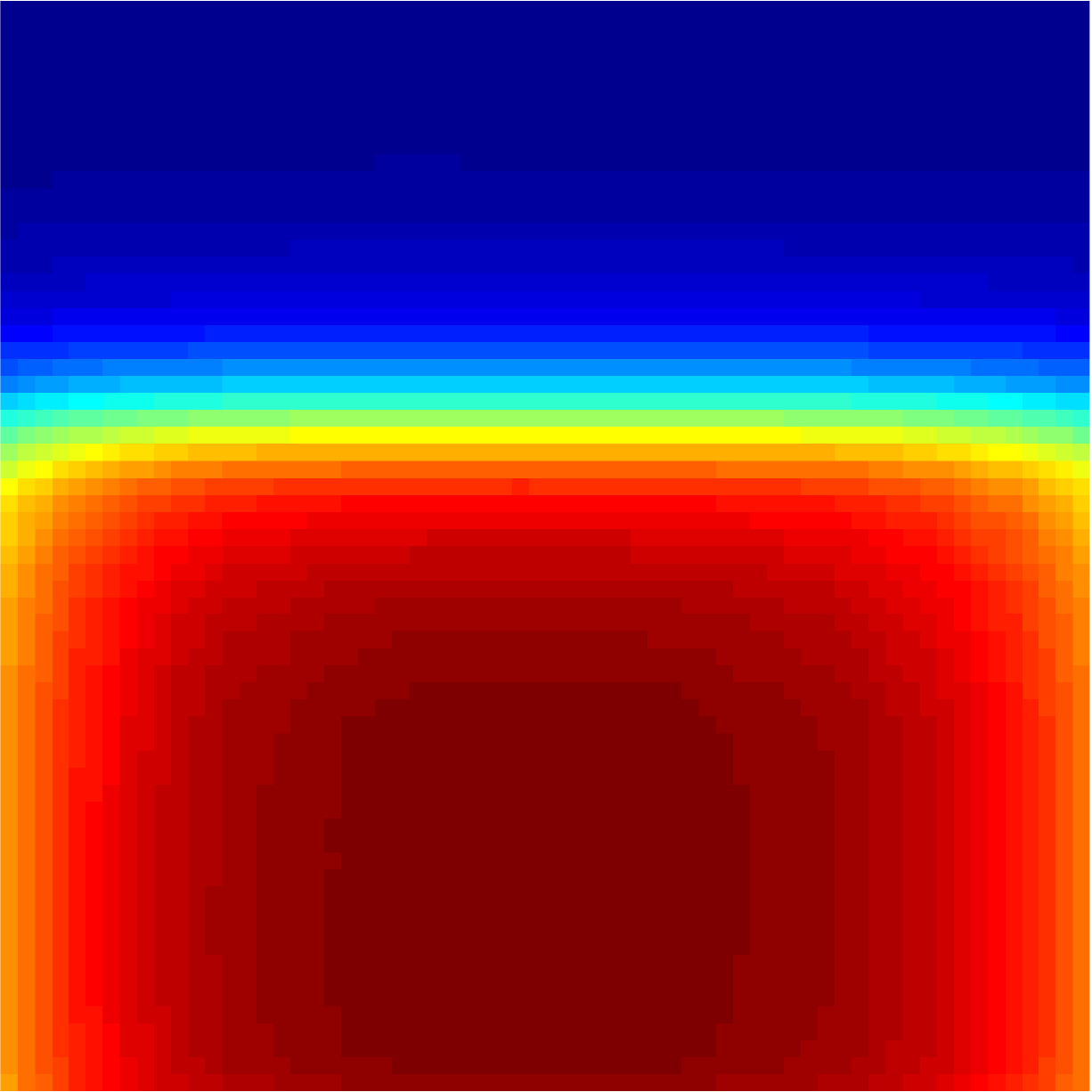}
    \caption{\footnotesize riding a bicycle}
  \end{subfigure}
  ~
  \begin{subfigure}[c]{0.31\linewidth}
    \centering
    \includegraphics[width=0.48\textwidth]{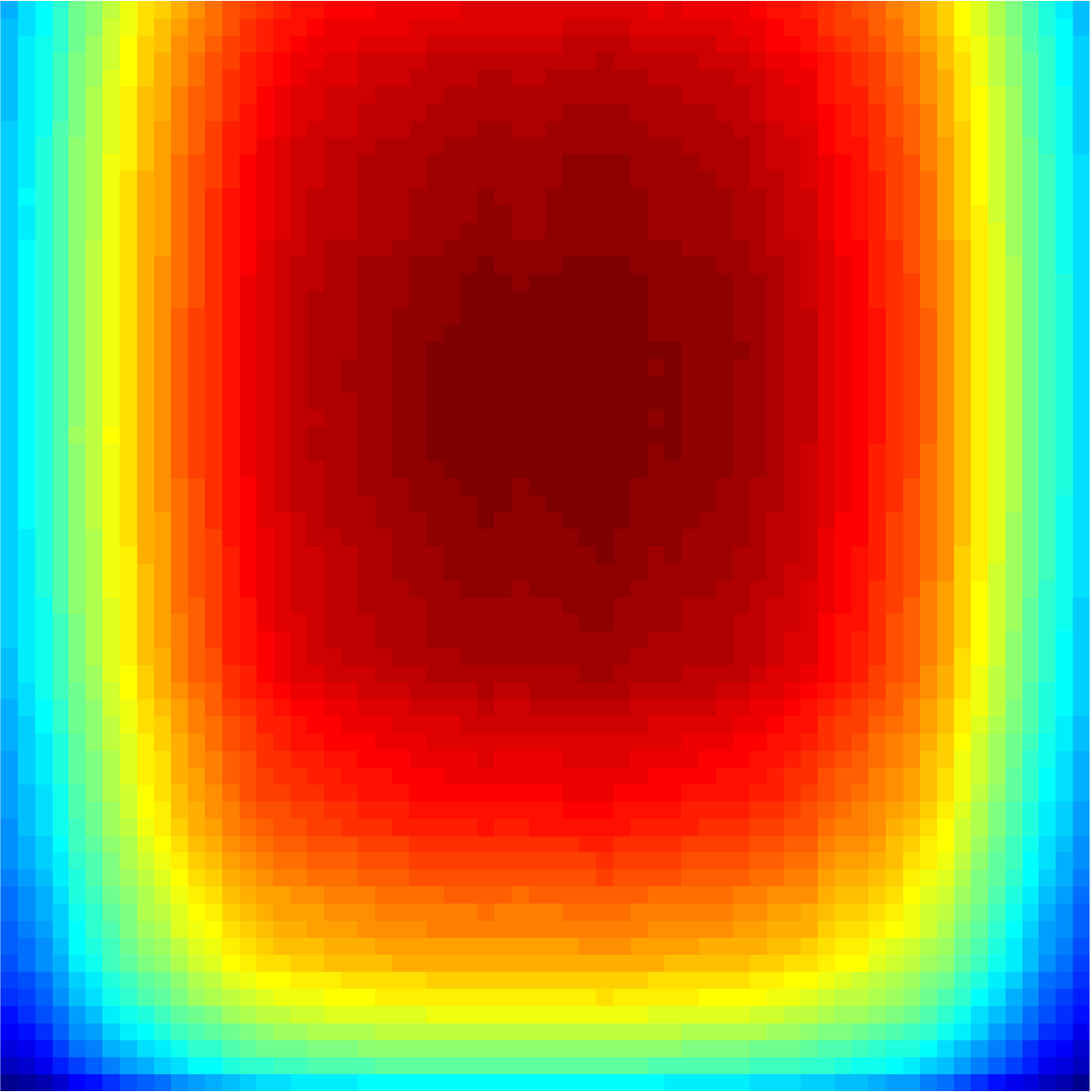}
    \includegraphics[width=0.48\textwidth]{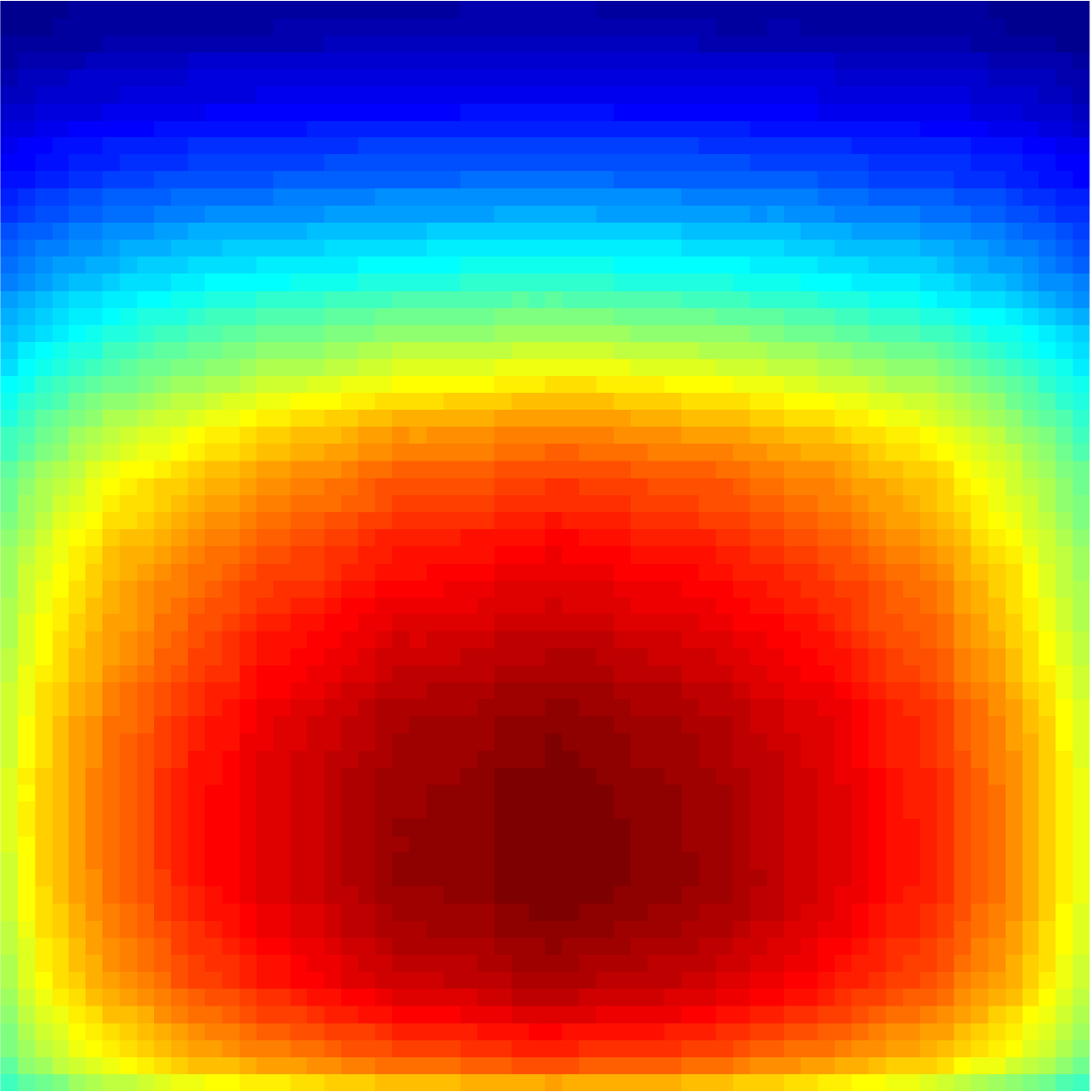}
    \caption{\footnotesize sitting on a chair}
  \end{subfigure}
  ~
  \begin{subfigure}[c]{0.31\linewidth}
    \centering
    \includegraphics[width=0.48\textwidth]{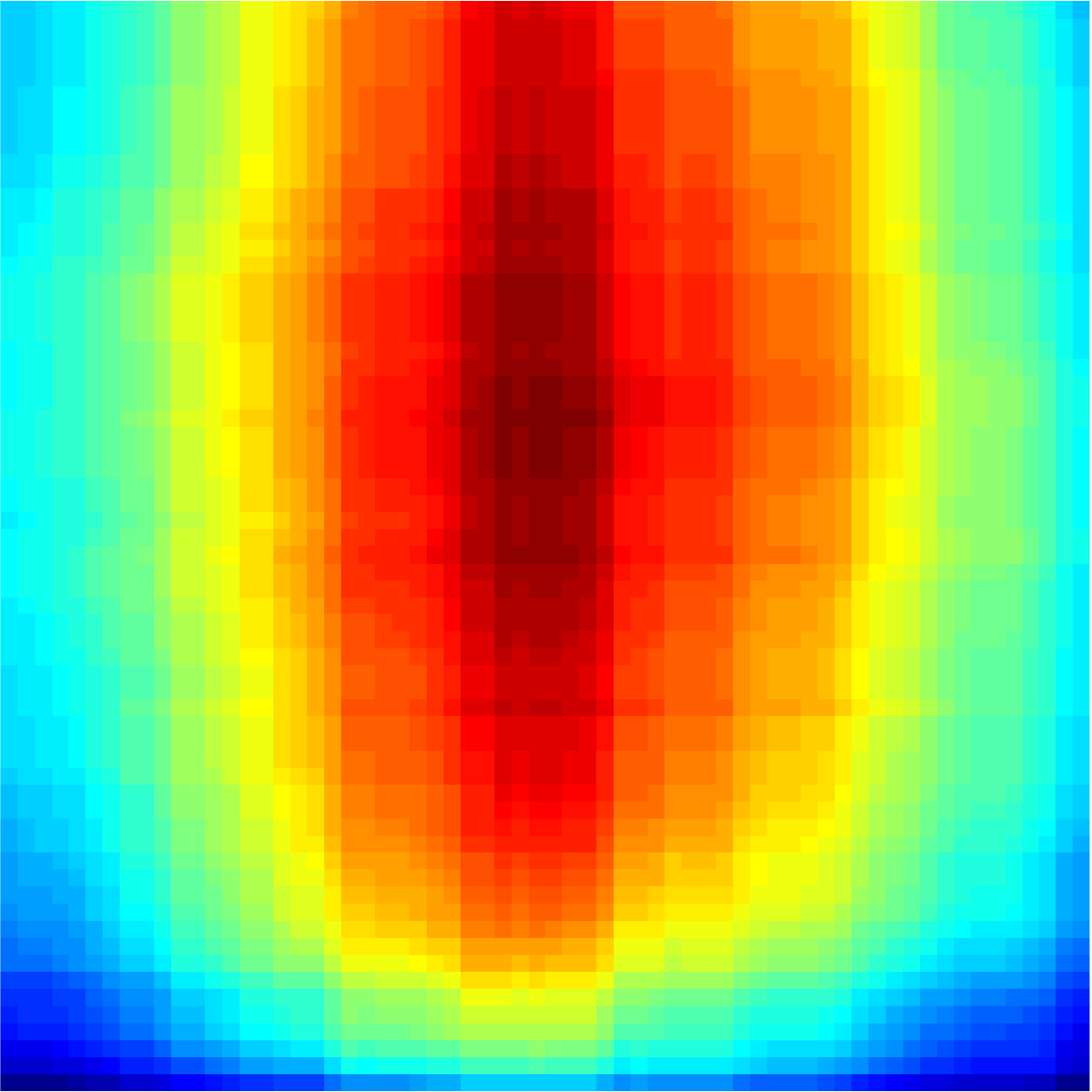}
    \includegraphics[width=0.48\textwidth]{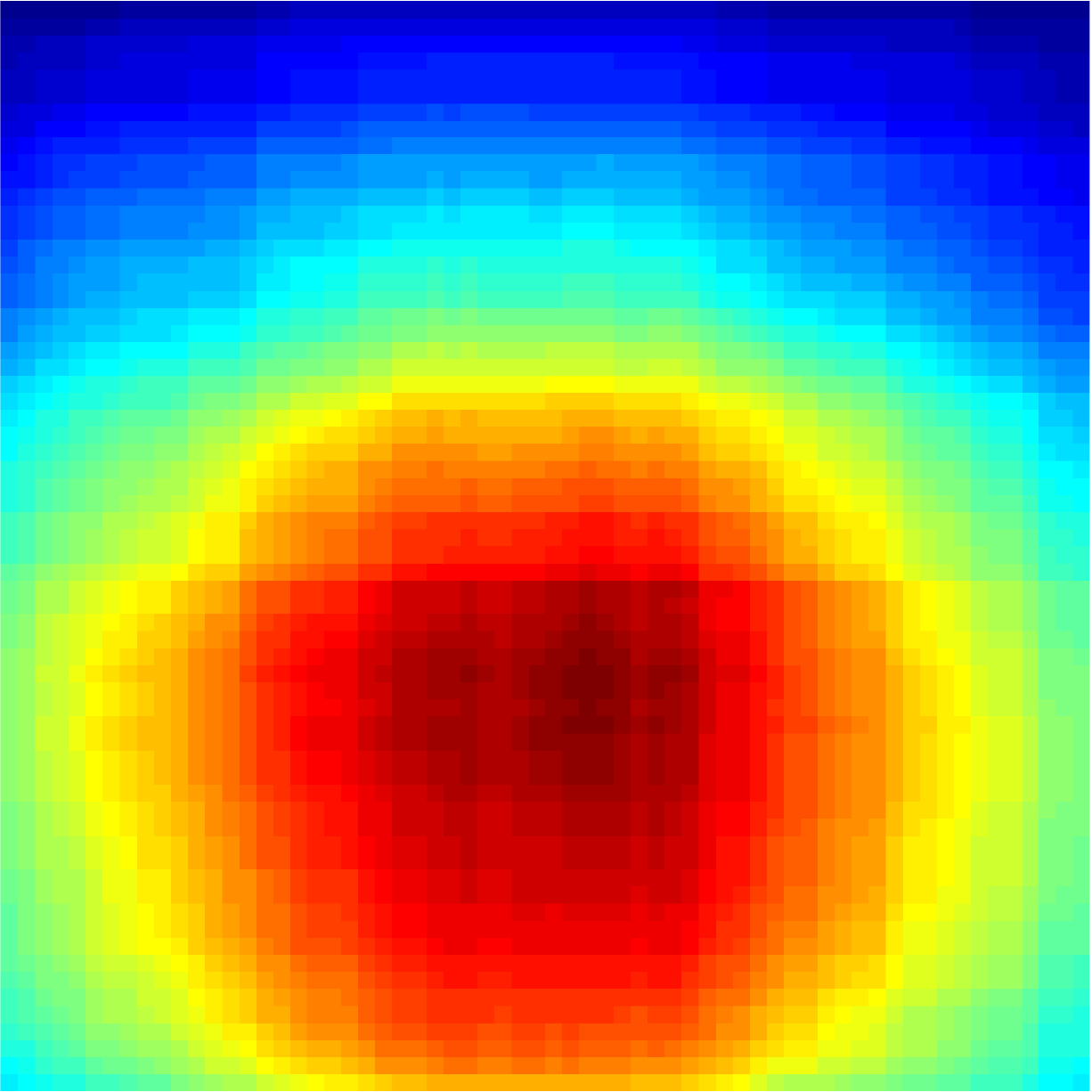}
    \caption{\footnotesize petting a dog}
  \end{subfigure}
  \\ \vspace{2mm}
  \begin{subfigure}[c]{0.31\linewidth}
    \centering
    \includegraphics[width=0.48\textwidth]{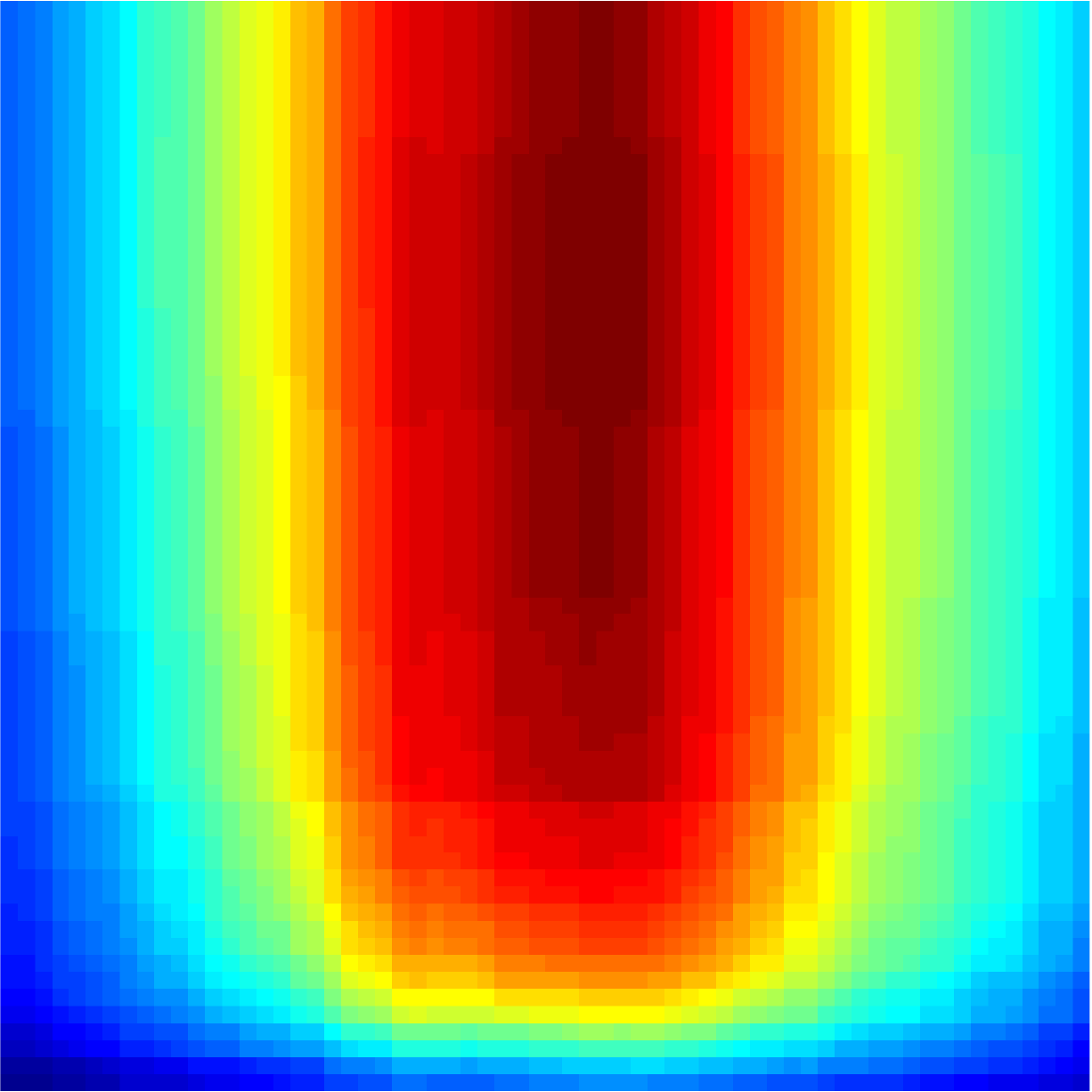}
    \includegraphics[width=0.48\textwidth]{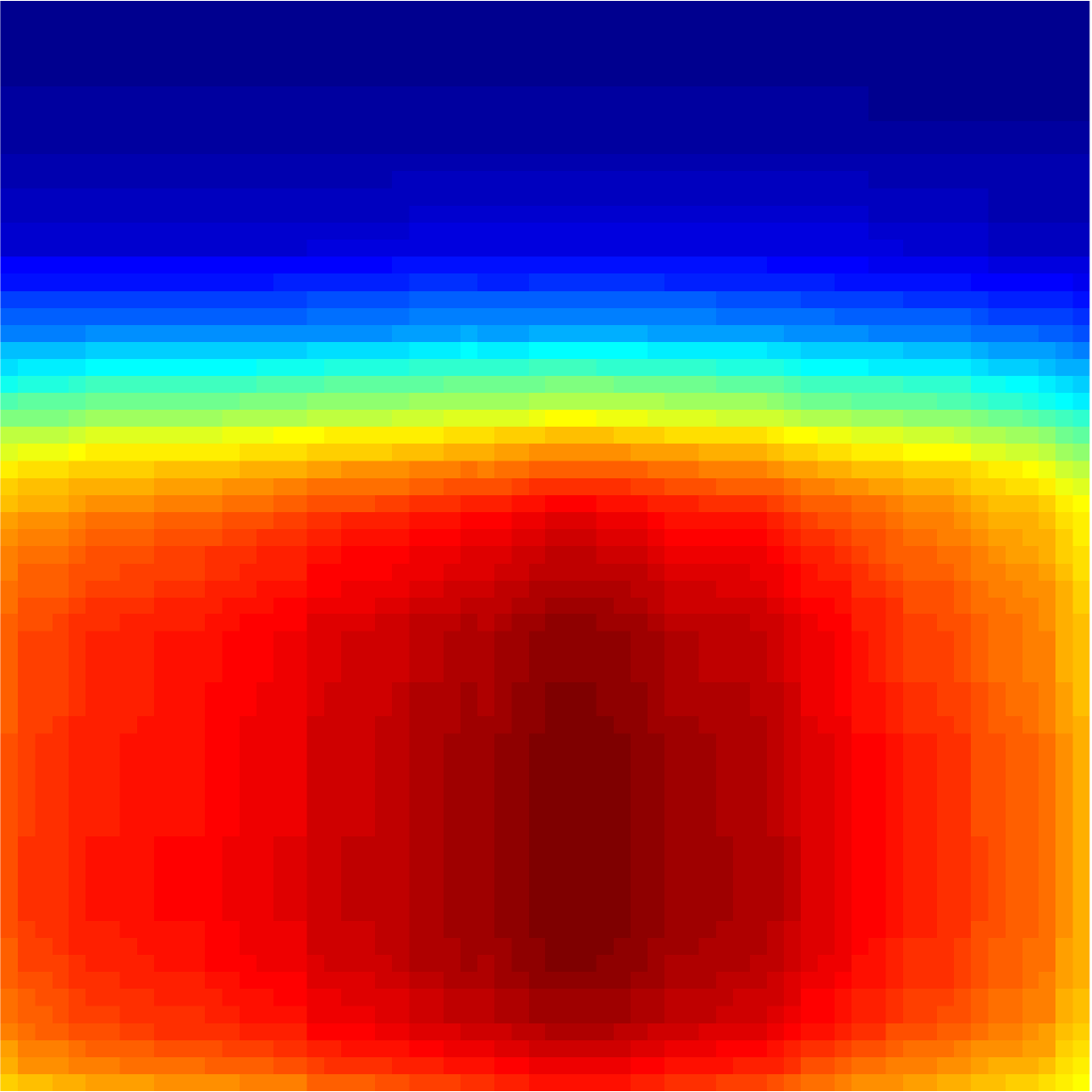}
    \caption{\footnotesize walking a bicycle}
  \end{subfigure}
  ~
  \begin{subfigure}[c]{0.31\linewidth}
    \centering
    \includegraphics[width=0.48\textwidth]{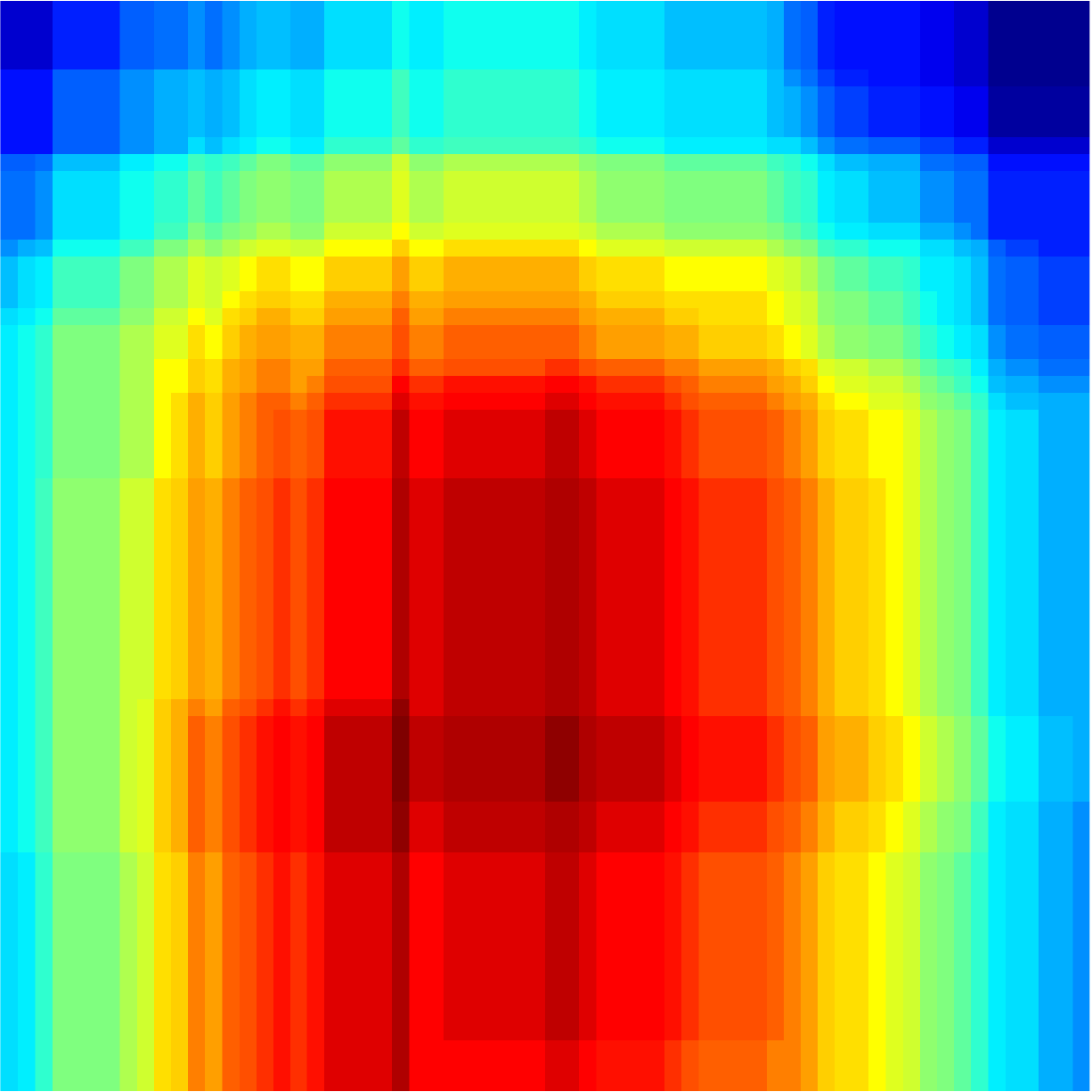}
    \includegraphics[width=0.48\textwidth]{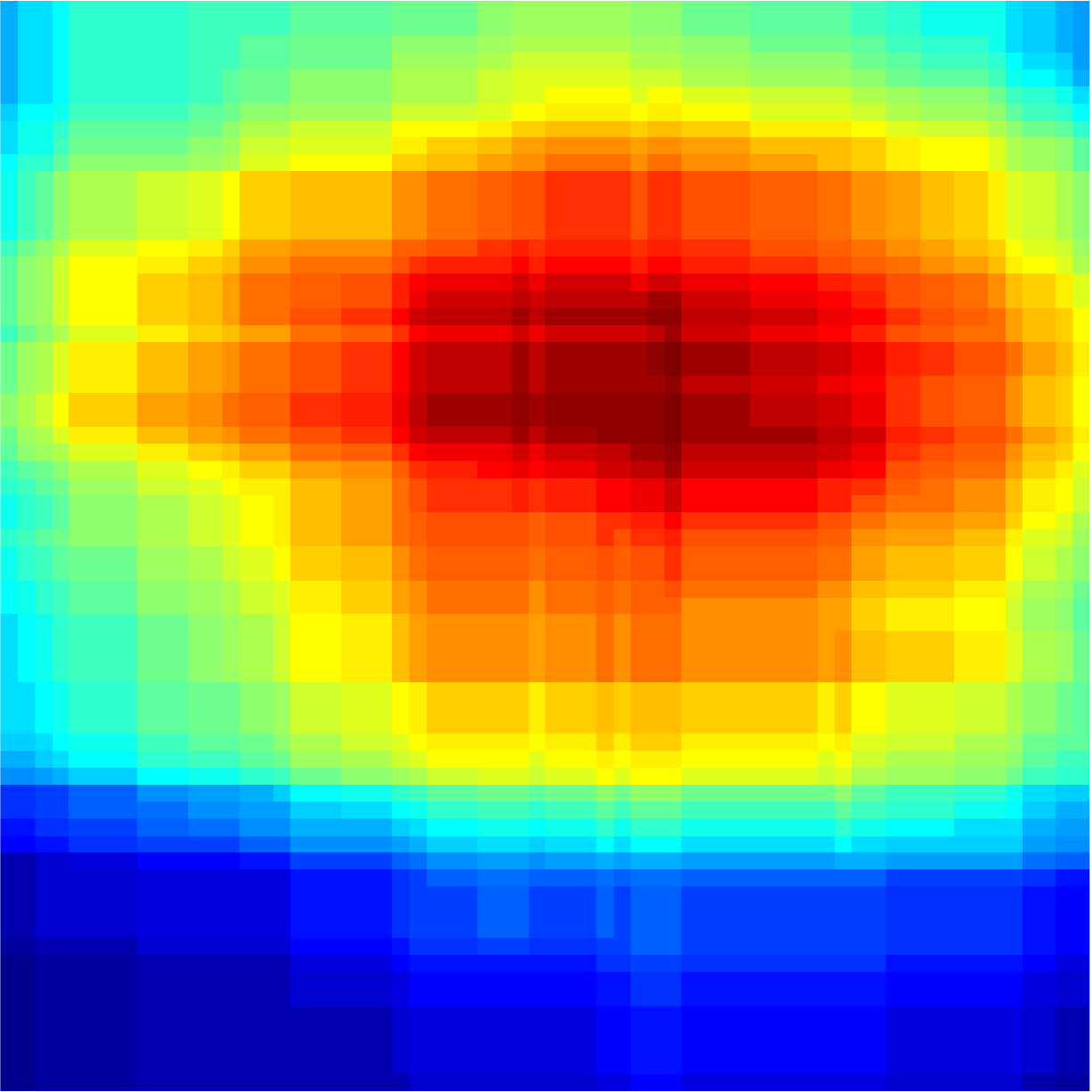}
    \caption{\footnotesize carrying a chair}
  \end{subfigure}
  ~
  \begin{subfigure}[c]{0.31\linewidth}
    \centering
    \includegraphics[width=0.48\textwidth]{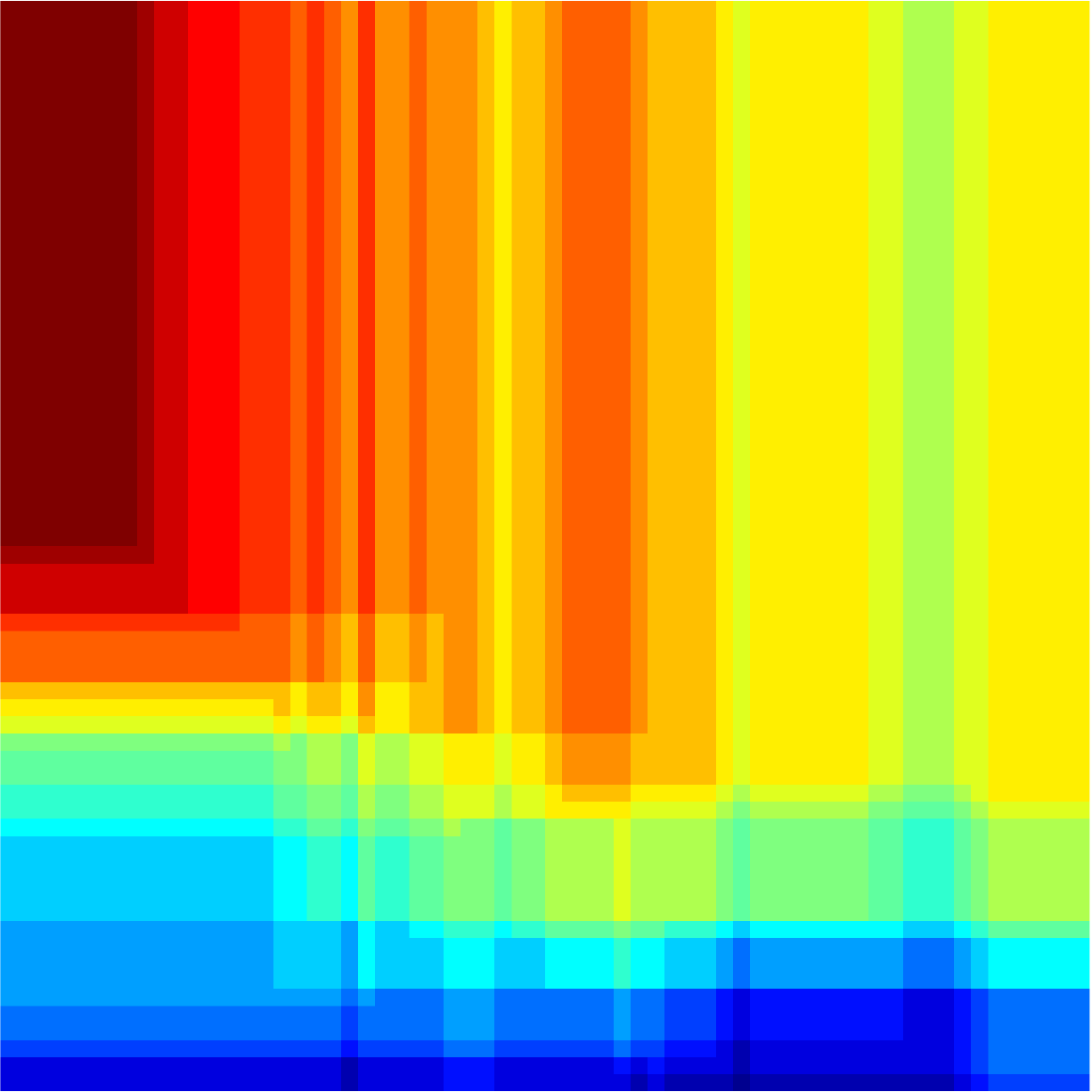}
    \includegraphics[width=0.48\textwidth]{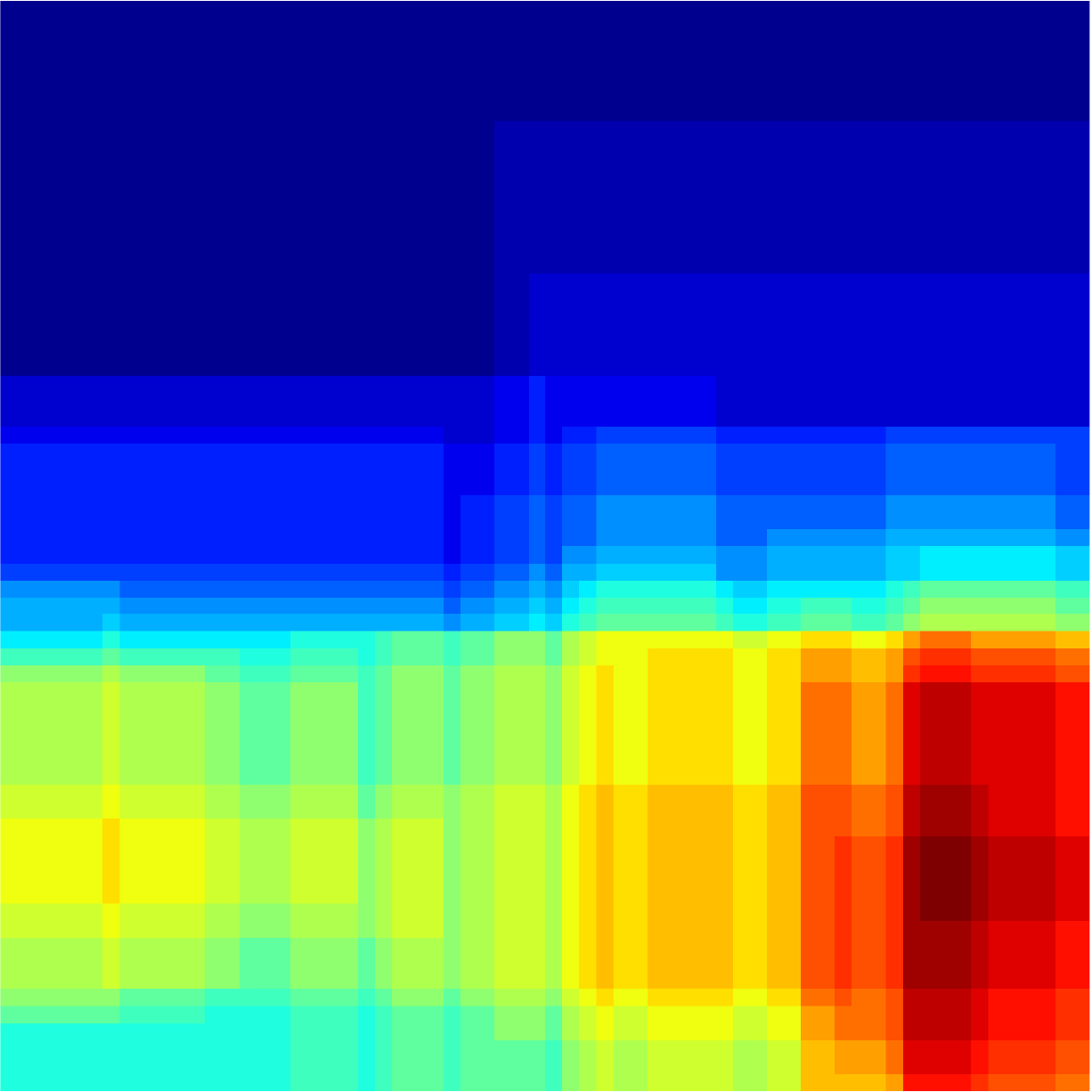}
    \caption{\footnotesize running a dog}
  \end{subfigure}
  \\ \vspace{2mm}
  \begin{subfigure}[c]{0.31\linewidth}
    \centering
   \includegraphics[width=0.48\textwidth]{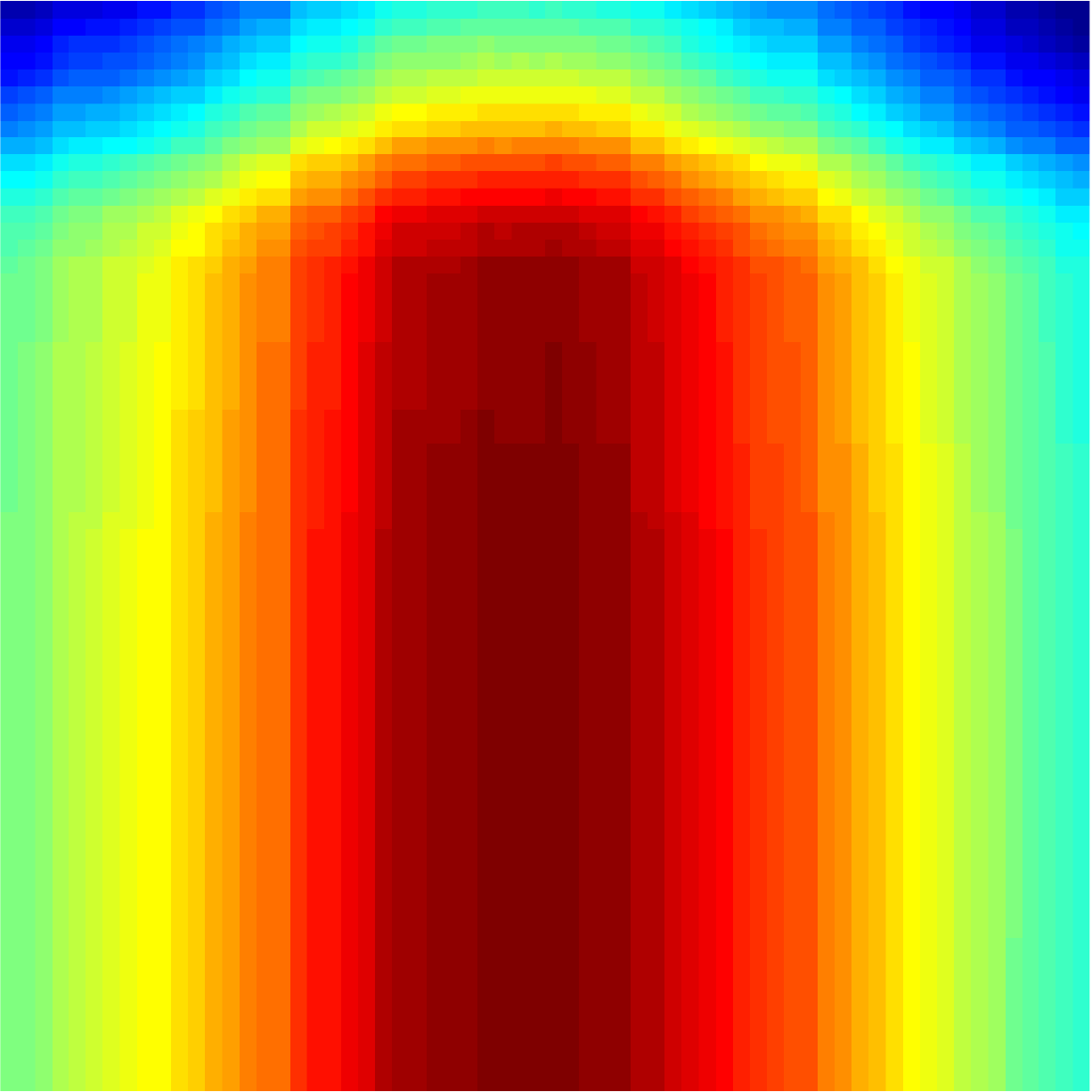}
    \includegraphics[width=0.48\textwidth]{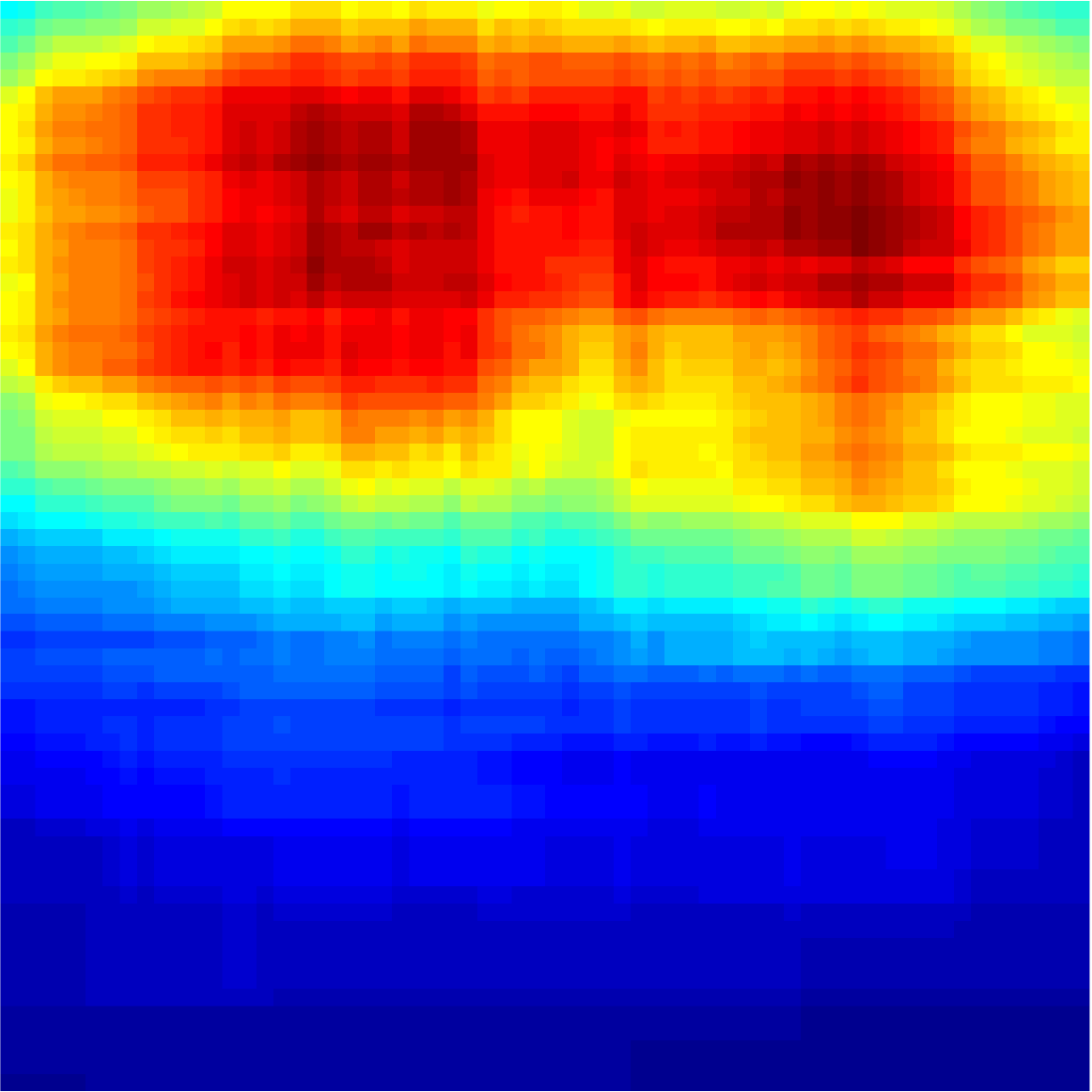}
    \caption{\footnotesize swinging a baseball bat}
  \end{subfigure}
  ~
  \begin{subfigure}[c]{0.31\linewidth}
    \centering
    \includegraphics[width=0.48\textwidth]{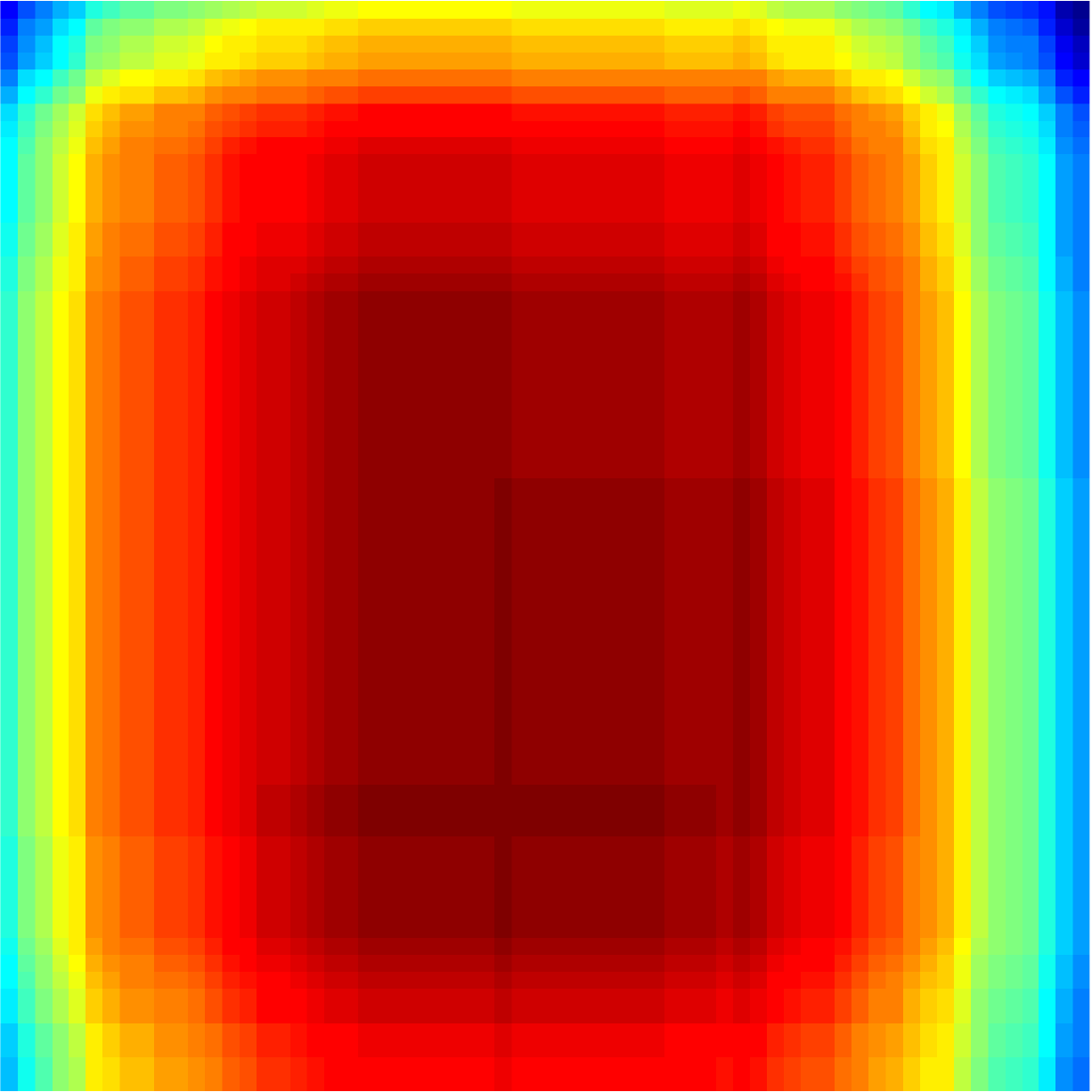}
    \includegraphics[width=0.48\textwidth]{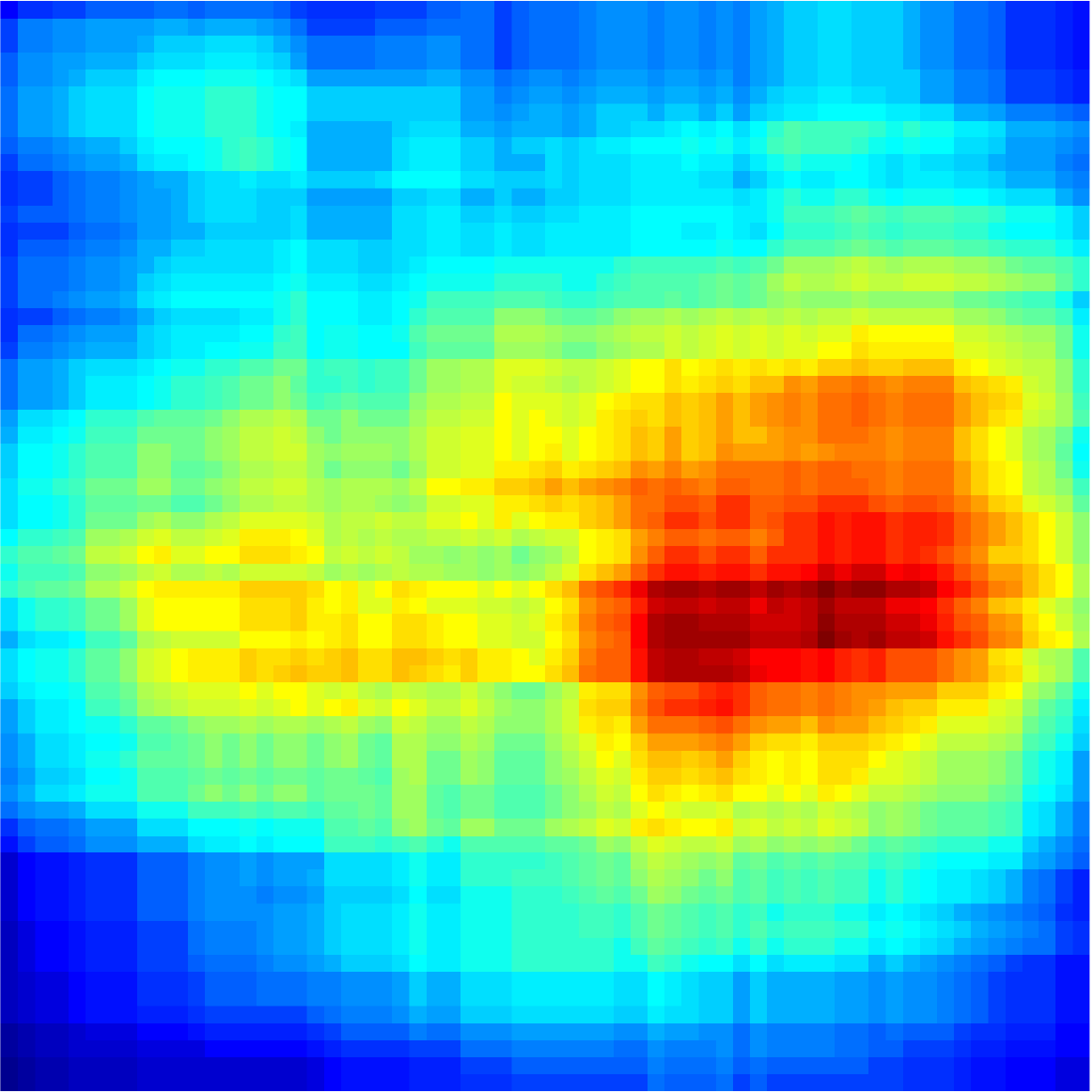}
    \caption{\fontsize{7.7}{9.2}\selectfont holding a baseball glove}
  \end{subfigure}
  ~
  \begin{subfigure}[c]{0.31\linewidth}
    \centering
    \includegraphics[width=0.48\textwidth]{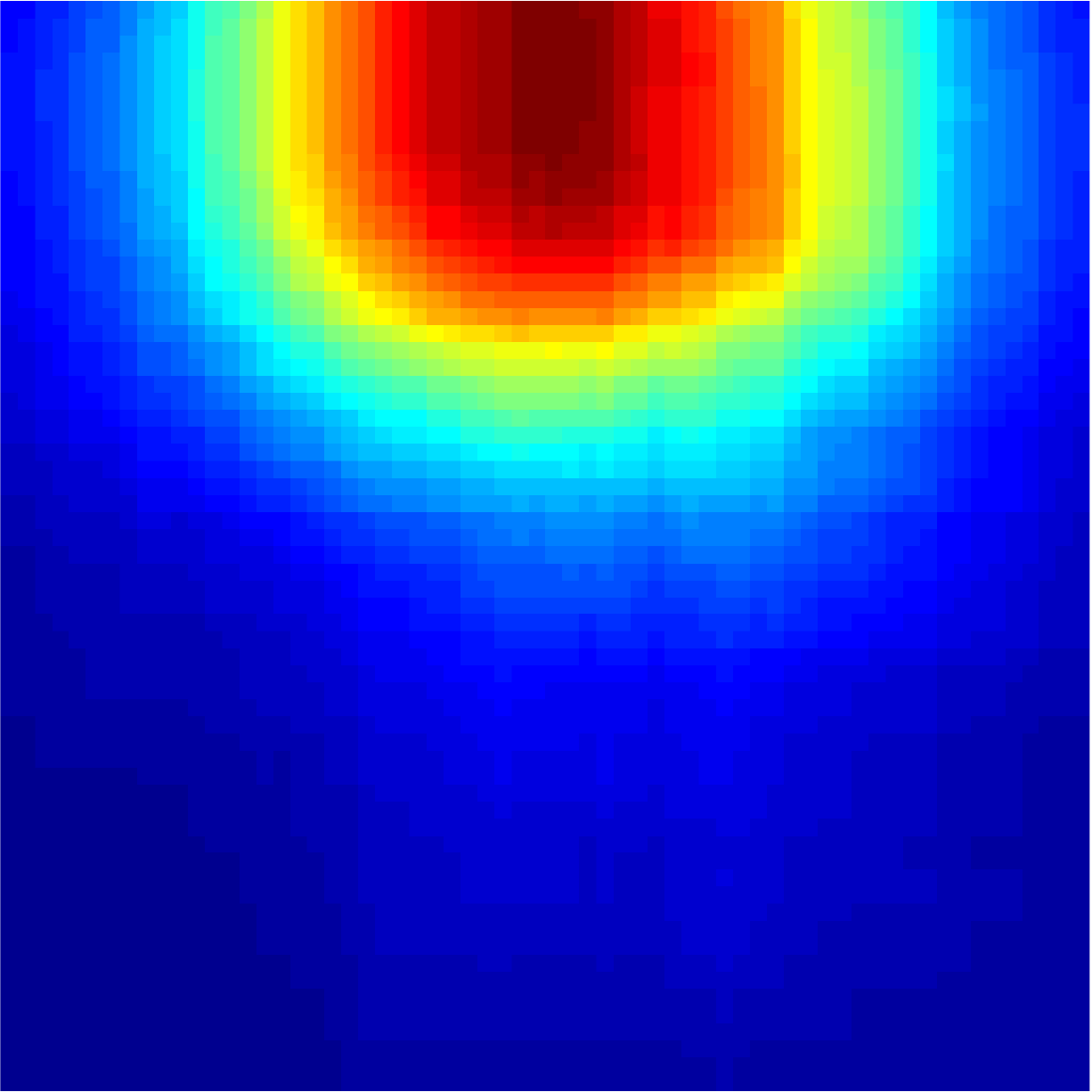}
    \includegraphics[width=0.48\textwidth]{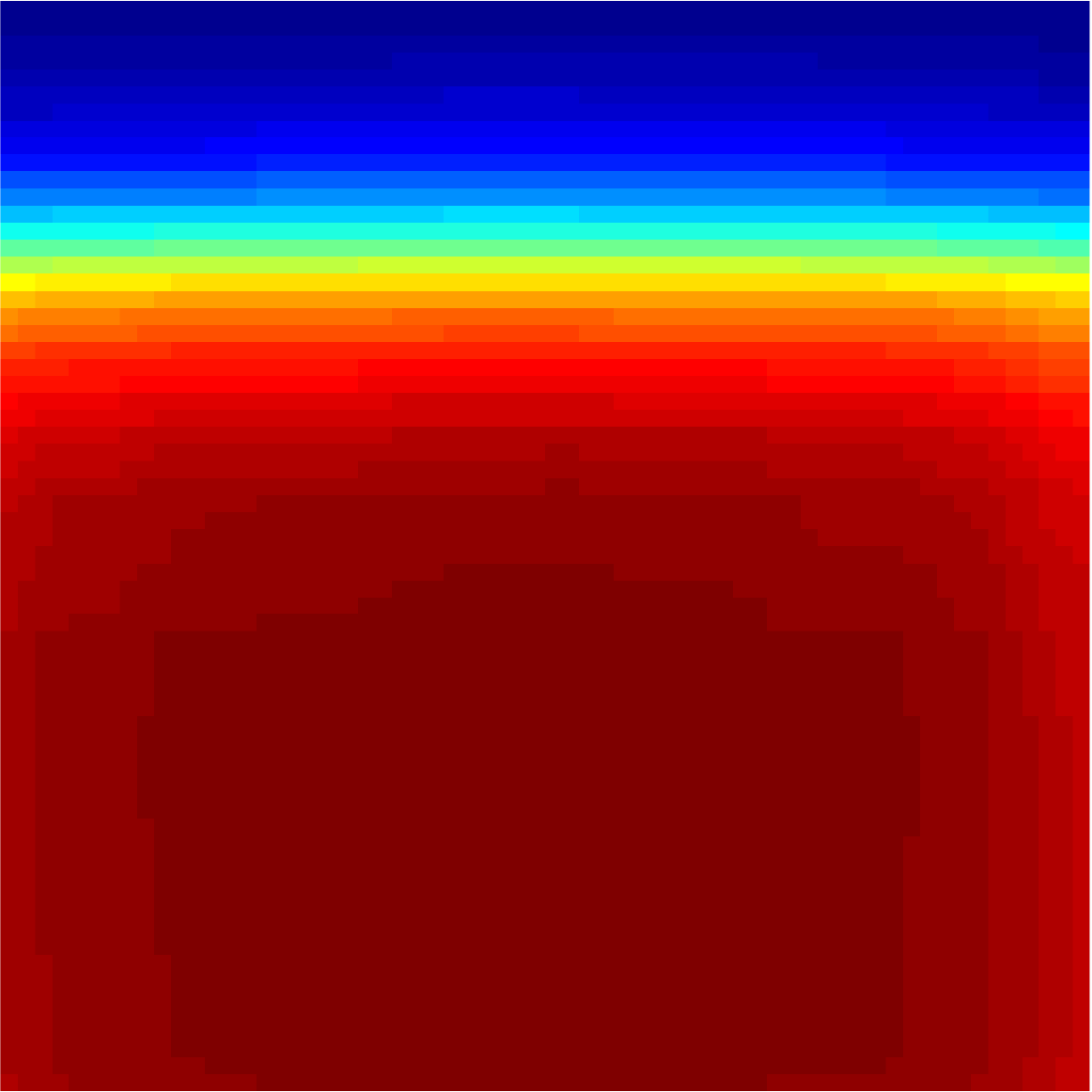}
    \caption{\footnotesize riding an elephant}
  \end{subfigure}
  \vspace{-2mm}
  \caption{\small Average Interaction Patterns for different HOI categories
obtained from ground-truth annotations. Left: average for the human channel.
Right: average for the object channel.}
  \vspace{-2mm}
  \label{fig:average_ip}
\end{figure}

We show the average Interaction Patterns obtained from the ground-truth
annotations of different HOIs in Fig.~\ref{fig:average_ip}. We see
distinguishable patterns for different interactions on the same object
category. For example, a chair involved in ``sitting on a chair'' is more
likely to be in the lower region of the Interaction Pattern, while a chair
involved in ``carrying a chair'' is more likely to be in the upper region.

We also separately evaluate the output of human, object, and pairwise stream.
Tab.~\ref{tab:onestream} shows the mAP of each stream on HO+IP1 (conv). The
object stream outperforms the other two in the Default setting. However, in the
Known Object setting, the pairwise stream achieves the highest mAP,
demonstrating the importance of human-object spatial relations for
distinguishing interactions.

\begin{table}[t]
  \centering
  \footnotesize
  \setlength{\tabcolsep}{5.45pt}
  \begin{tabular}{|l||C{0.58cm}|C{0.58cm}|C{0.58cm}||C{0.64cm}|C{0.58cm}|C{0.64cm}|}
    \hline \TBstrut
                  & \multicolumn{3}{c||}{Default}                 & \multicolumn{3}{c|}{Known Object}             \\
    \cline{2-7} \TBstrut
                  & Full          & Rare          & Non-Rare      & Full          & Rare          & Non-Rare      \\
    \hline \Tstrut
    Human         & 0.70          & 0.08          & 0.88          & 2.44          & 2.14          & 2.53          \\
    Object        & \textbf{2.11} & \textbf{1.19} & \textbf{2.39} & 3.09          & \textbf{2.98} & 3.13          \\ \Bstrut
    Pairwise      & 0.30          & 0.06          & 0.37          & \textbf{3.21} & 2.80          & \textbf{3.33} \\
    \hline
  \end{tabular}
  \vspace{-2mm}
  \caption{\small mAP (\%) of each stream on HO+IP1 (conv).}
  \vspace{-1mm}
  \label{tab:onestream}
\end{table}

\vspace{-3mm}

\paragraph{Leveraging Object Detection Scores} So far we assume the
human-object proposals always consist of true object detections, so the HO-RCNN
is only required to distinguish the interactions. In practice, the proposals
may contain false detections, and the HO-RCNN should learn to generate low
scores for all HOI categories in such case. We thus add an extra path with a
single neuron that takes the raw object detection score associated with each
proposal and produces an offset to the final HOI detection scores. This
provides a means by which the final detection scores can be lowered if the raw
detection score is low. We show the effect of adding this extra component (HO+S
and HO+IP1 (conv)+S) in Tab.~\ref{tab:score} (top) and the signifiance of the
improvements in Tab.~\ref{tab:score} (bottom). The improvement is significant
in the Default setting, since the extra background images increase the number
of false object detections.

\begin{table}[t]
  \centering
  \begin{subtable}{\linewidth}
    \centering
    \footnotesize
    \setlength{\tabcolsep}{4.81pt}
    \begin{tabular}{|l||C{0.58cm}|C{0.58cm}|C{0.58cm}||C{0.64cm}|C{0.58cm}|C{0.64cm}|}
      \hline \TBstrut
                       & \multicolumn{3}{c||}{Default}                 & \multicolumn{3}{c|}{Known Object}               \\
      \cline{2-7} \TBstrut
                       & Full          & Rare          & Non-Rare      & Full           & Rare          & Non-Rare       \\
      \hline \Tstrut
      HO               & 5.73          & 3.21          & 6.48          & 8.46           & 7.53          & 8.74           \\ \Bstrut
      HO+S             & 6.07          & 3.79          & 6.76          & 8.09           & 6.79          & 8.47           \\
      \hline \Tstrut
      HO+IP1 (conv)    & 7.30          & 4.68          & 8.08          & 10.37          & \textbf{9.06} & 10.76          \\ \Bstrut
      HO+IP1 (conv)+S  & \textbf{7.81} & \textbf{5.37} & \textbf{8.54} & \textbf{10.41} & 8.94          & \textbf{10.85} \\
      \hline
    \end{tabular}
  \end{subtable}
  \\
  \vspace{2mm}
  \begin{subtable}{\linewidth}
    \centering
    \setlength{\tabcolsep}{4.05pt}
    \fontsize{7.5}{10}\selectfont
    \begin{tabular}{|l||C{1.00cm}|C{1.00cm}|C{1.00cm}|}
      \hline \TBstrut
                          & \multicolumn{3}{c|}{Default}                  \\
      \cline{2-4} \TBstrut
                                          & Full     & Rare    & Non-Rare \\
      \hline \Tstrut
      HO+S vs. HO                         & 0.002    & 0.024   & 0.016    \\ \Bstrut
      HO+IP1 (conv)+S vs. HO+IP1 (conv)   & $<0.001$ & 0.028   & $<0.001$ \\
      \hline
    \end{tabular}
  \end{subtable}
  \vspace{-2mm}
  \caption{\small Performance comparison of combining object detection scores.
Top: mAP (\%). Bottom: p-value for the paired t-test.}
  \vspace{-1mm}
  \label{tab:score}
\end{table}

\begin{figure*}[t!]
  \centering
  \footnotesize
  \begin{tabular}{L{0.54\linewidth}@{\hspace{1.0mm}}|L{0.43\linewidth}@{\hspace{-1.0mm}}}
    \hspace{-2.5mm}
    \includegraphics[height=0.112\textwidth]{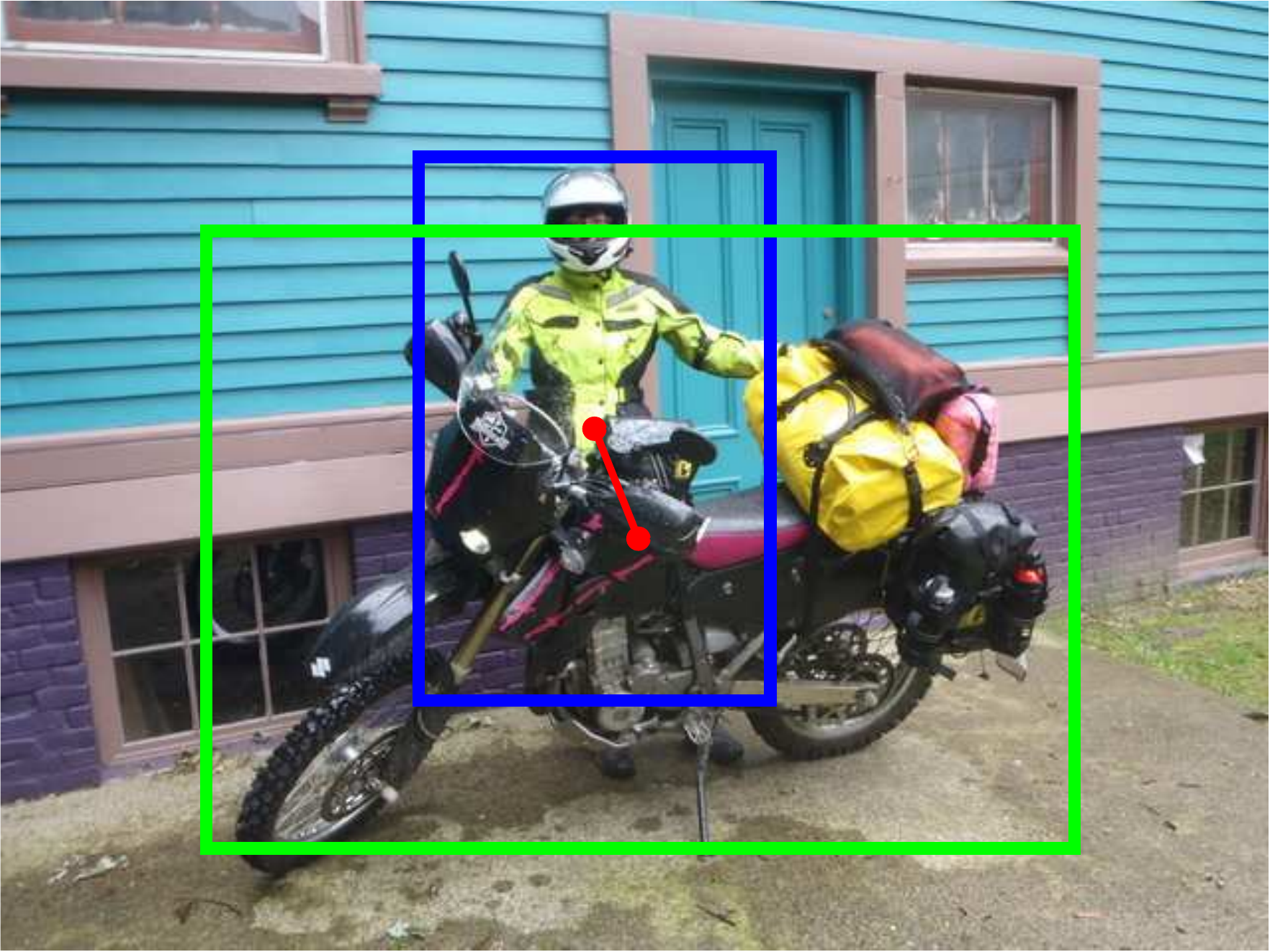}~
    \includegraphics[height=0.112\textwidth]{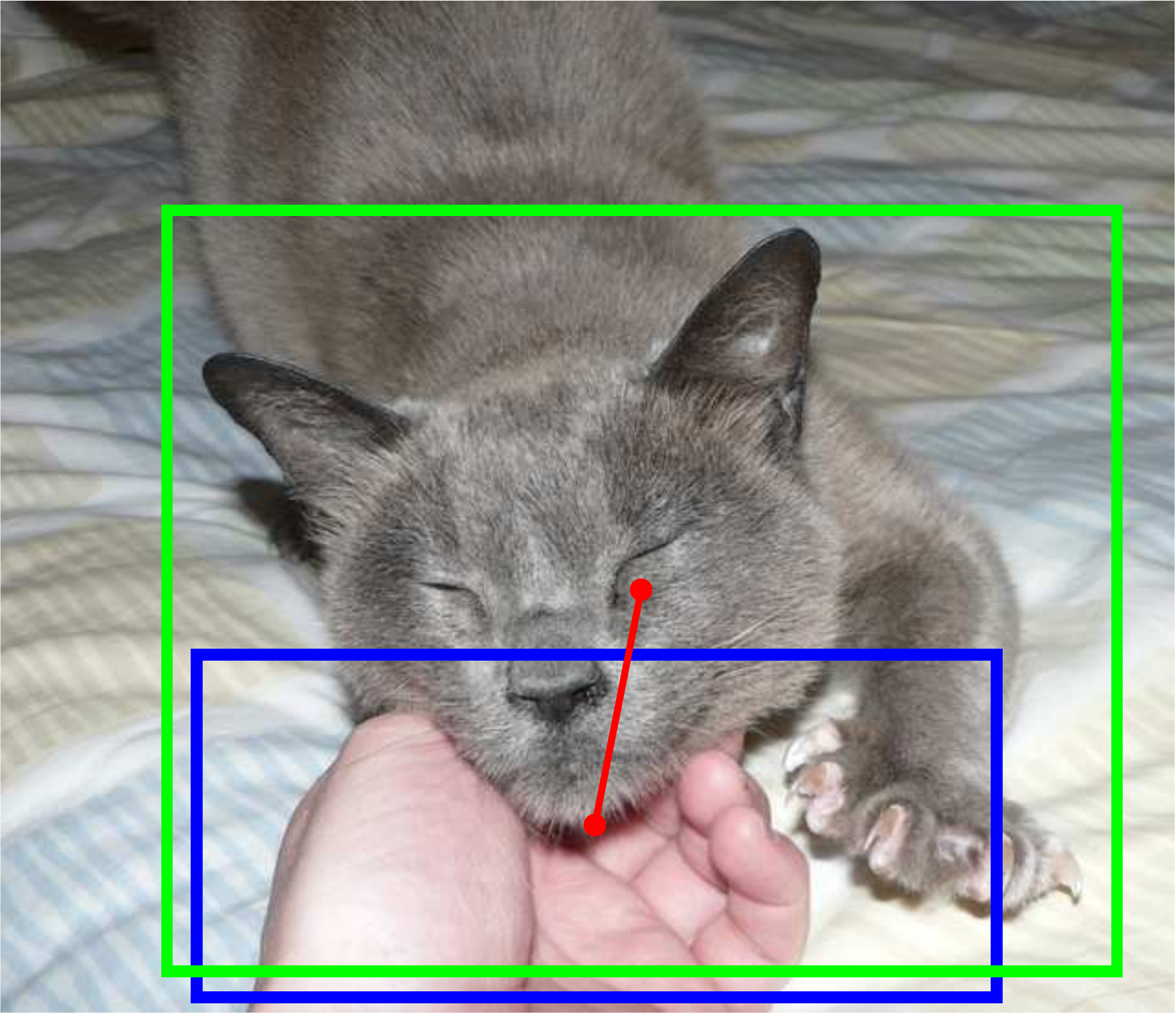}~
    \includegraphics[height=0.112\textwidth]{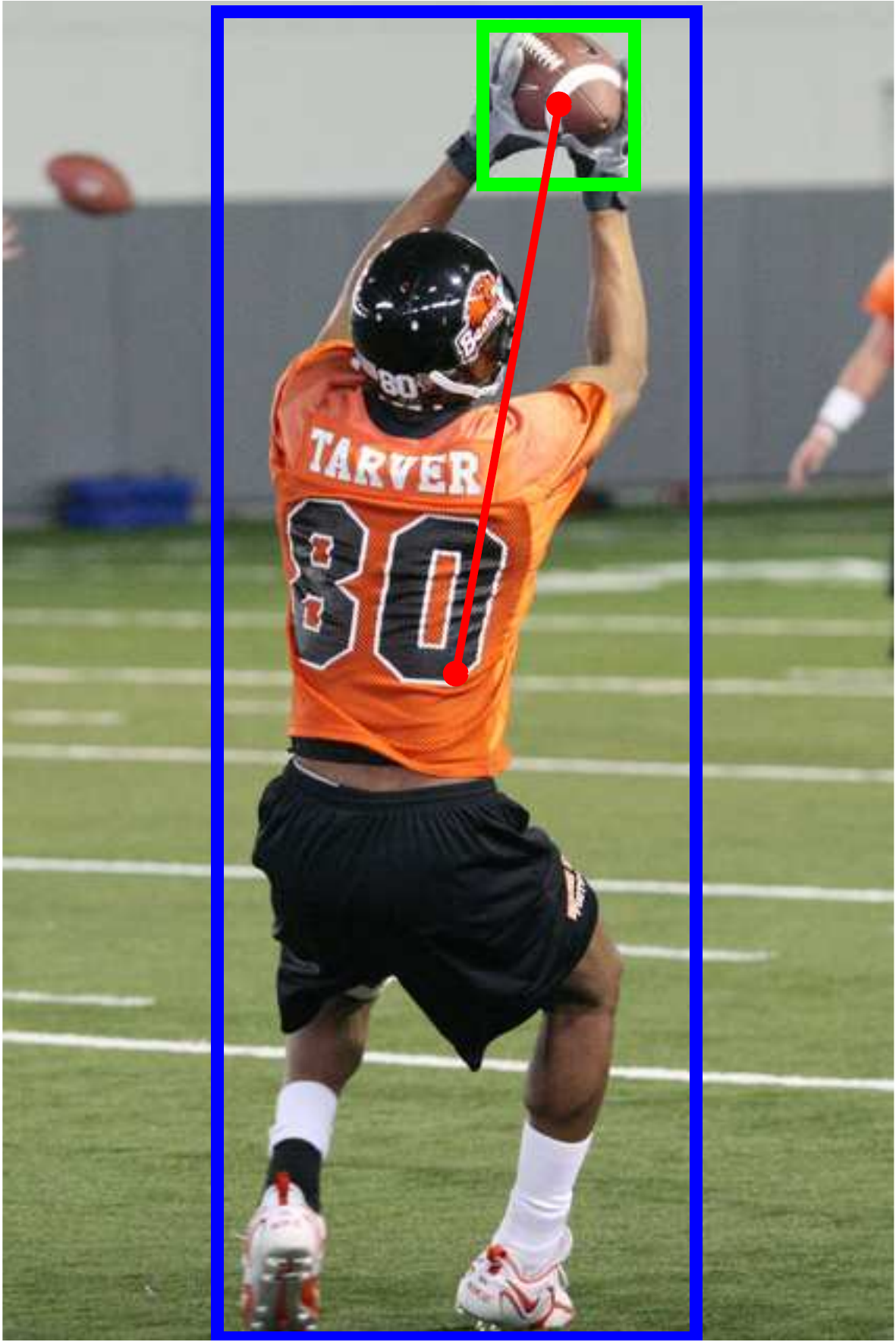}~
    \includegraphics[height=0.112\textwidth]{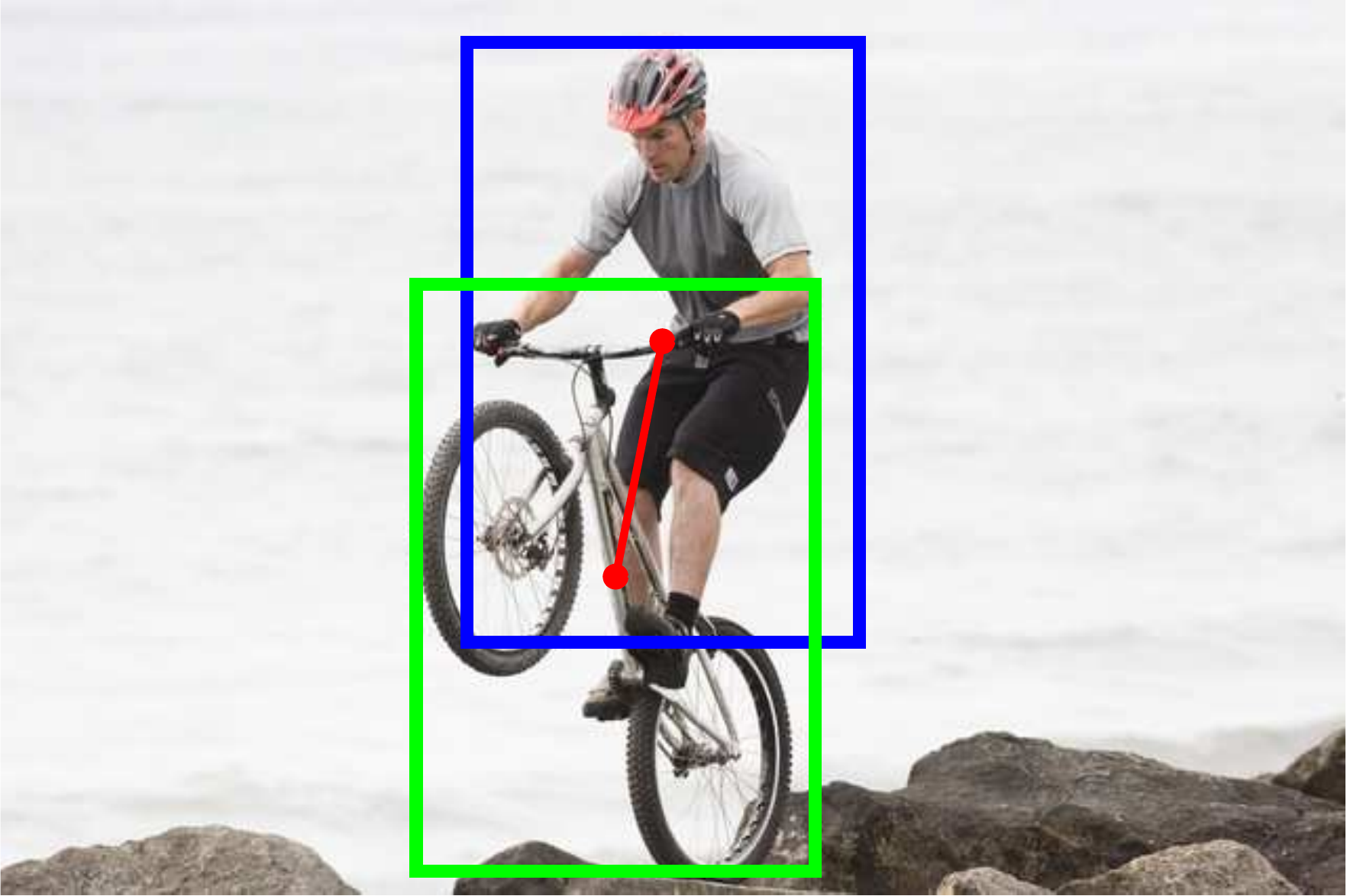}
    &
    \includegraphics[height=0.112\textwidth]{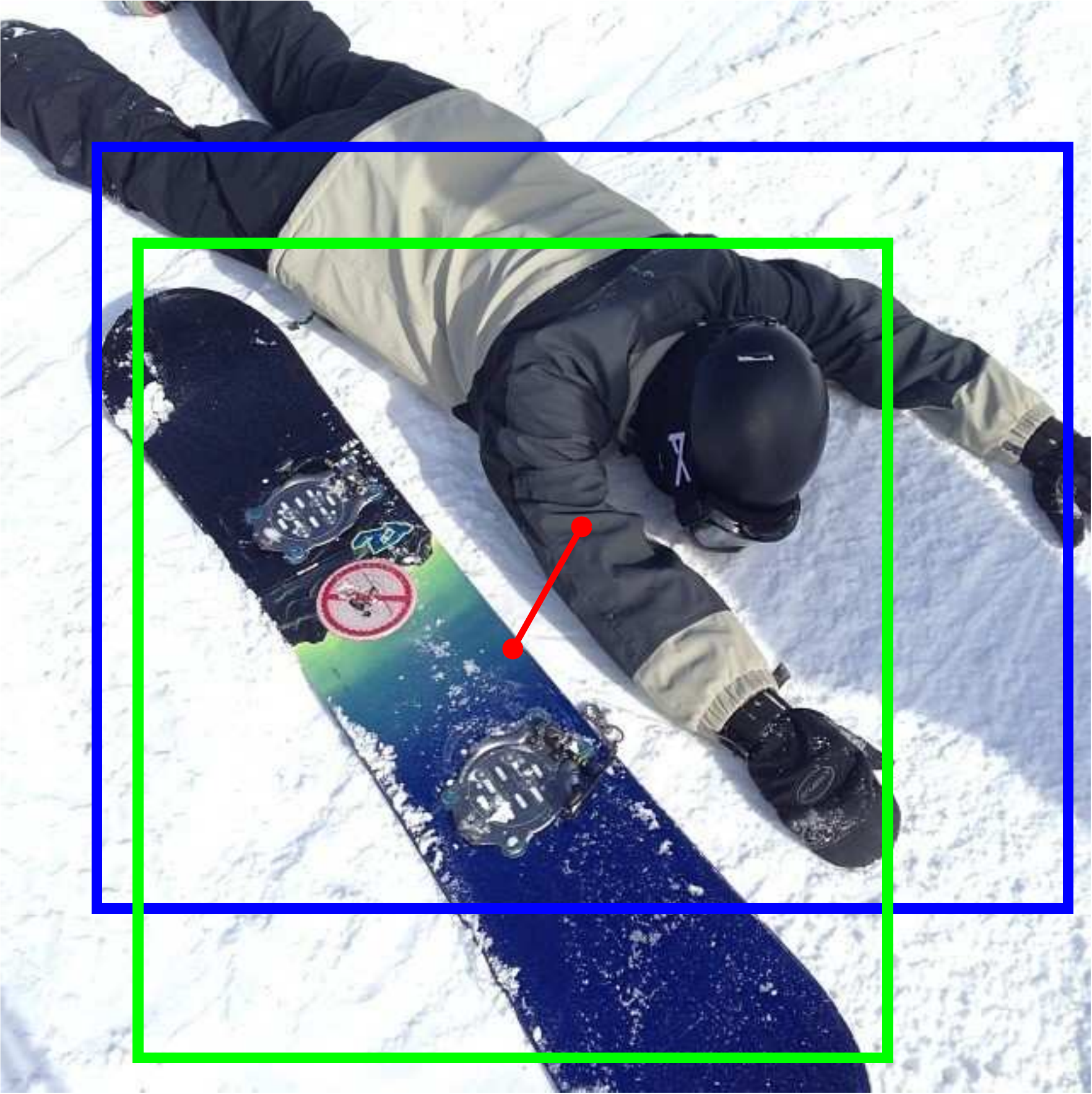}~
    \includegraphics[height=0.112\textwidth]{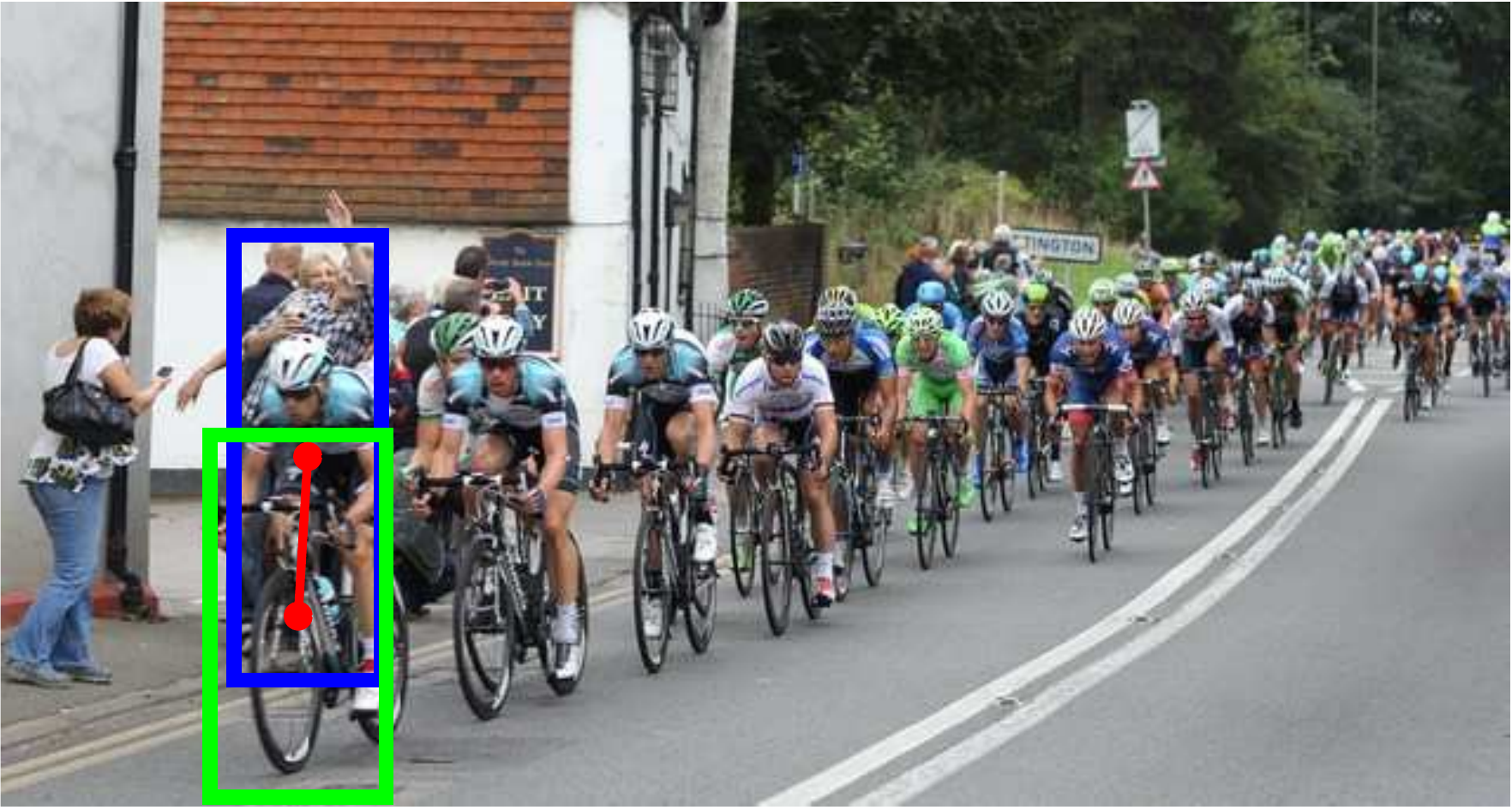}~
    \includegraphics[height=0.112\textwidth]{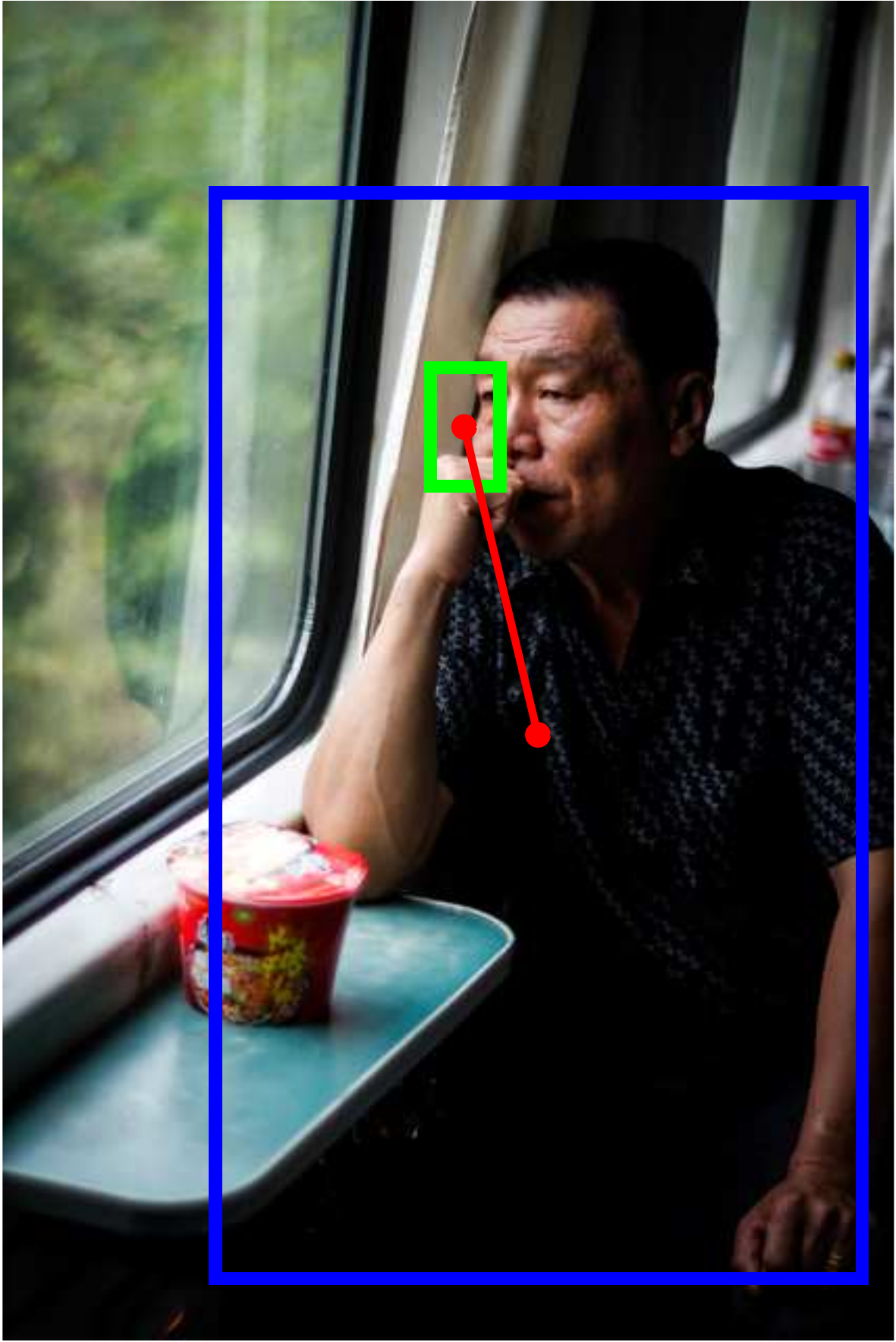}
    \\
    \hspace{-2.5mm}
    ~holding a motorcycle~~~~~~~~scratching a cat~~~~catching a ball~~~~~jumping a bicycle
    &
    standing on a snowboard~~~riding a bicycle~~~talking on a cell phone
    \\
    ~~~~~~~~~~~~0.94~~~~~~~~~~~~~~~~~~~~~~~~~~~~~~0.95~~~~~~~~~~~~~~~~~~~~~0.94~~~~~~~~~~~~~~~~~~~~~~~~~0.99
    &
    ~~~~~~~~~~~~~~~~~0.99~~~~~~~~~~~~~~~~~~~~~~~~~~~~~0.99~~~~~~~~~~~~~~~~~~~~~~~~0.81
    \\
    \hspace{-2.5mm}
    \includegraphics[height=0.115\textwidth]{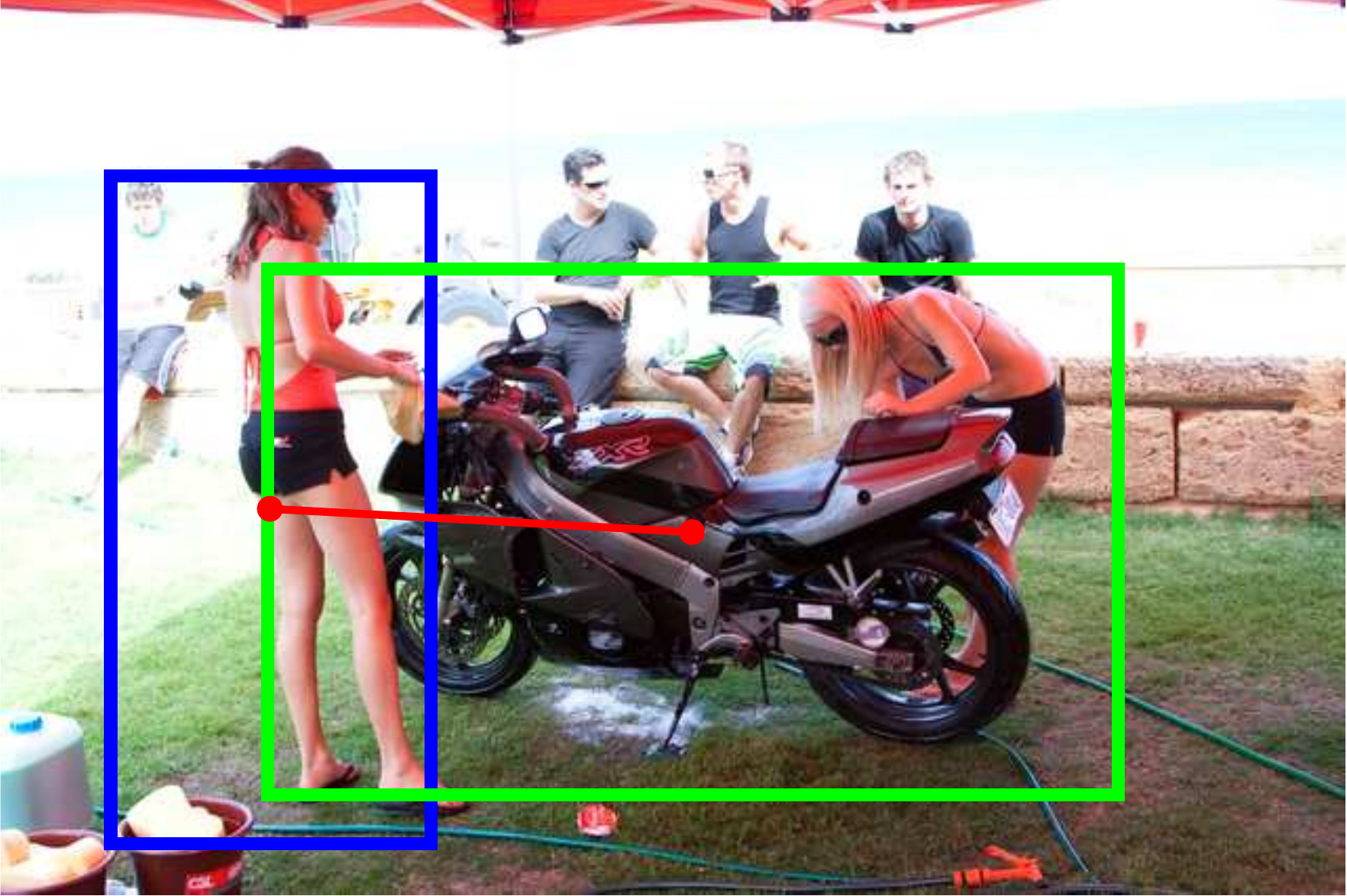}~
    \includegraphics[height=0.115\textwidth]{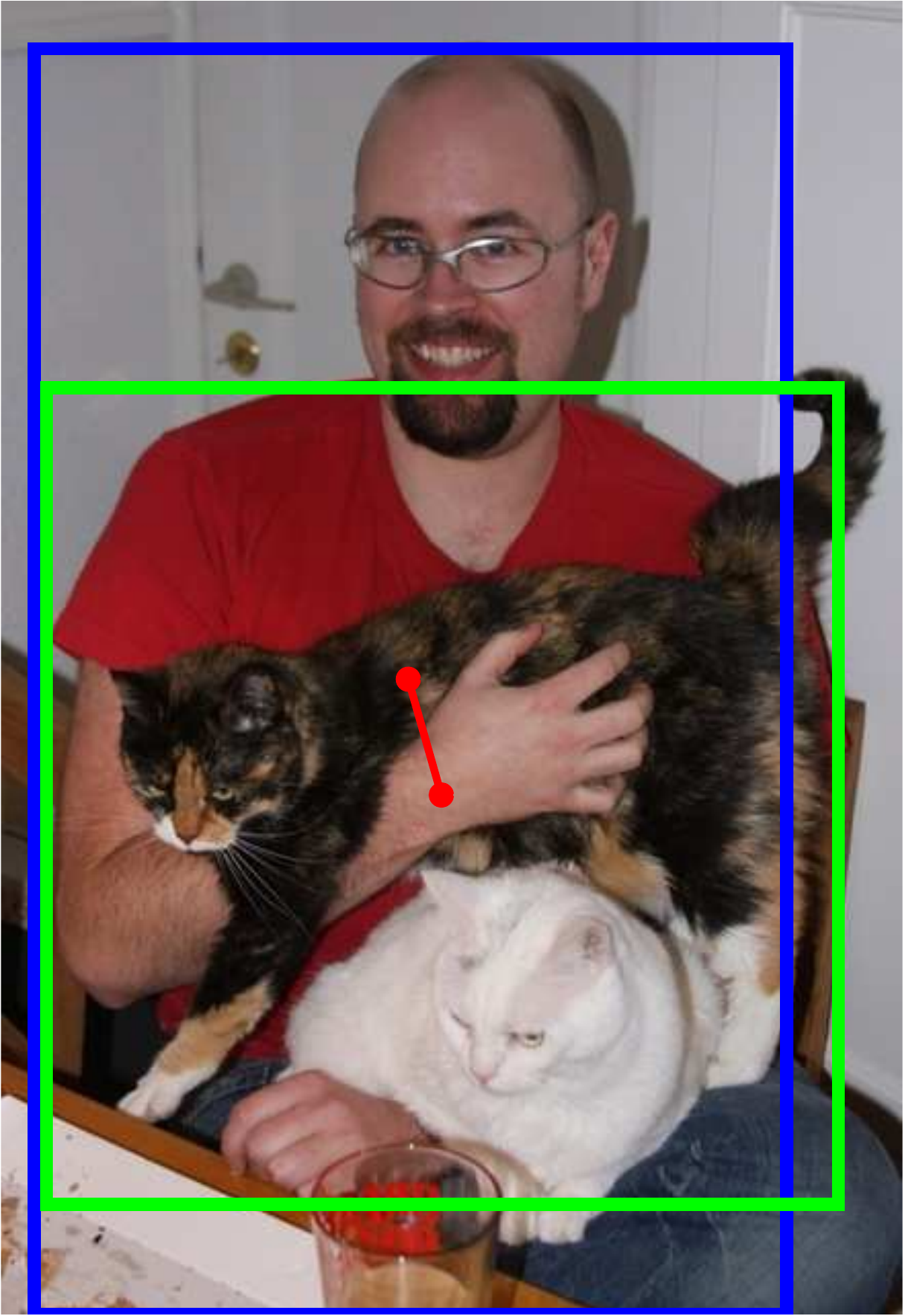}~
    \includegraphics[height=0.115\textwidth]{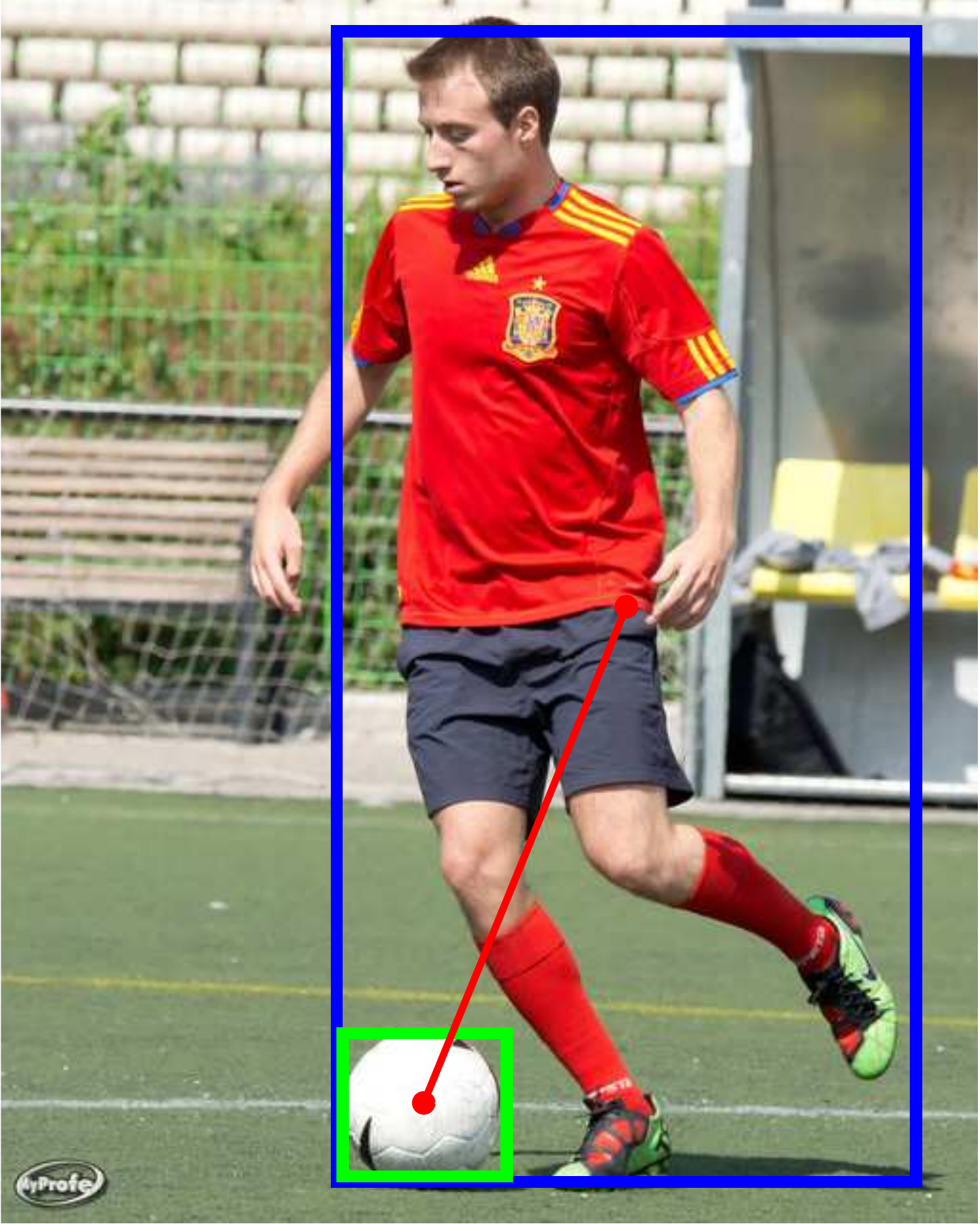}~
    \includegraphics[height=0.115\textwidth]{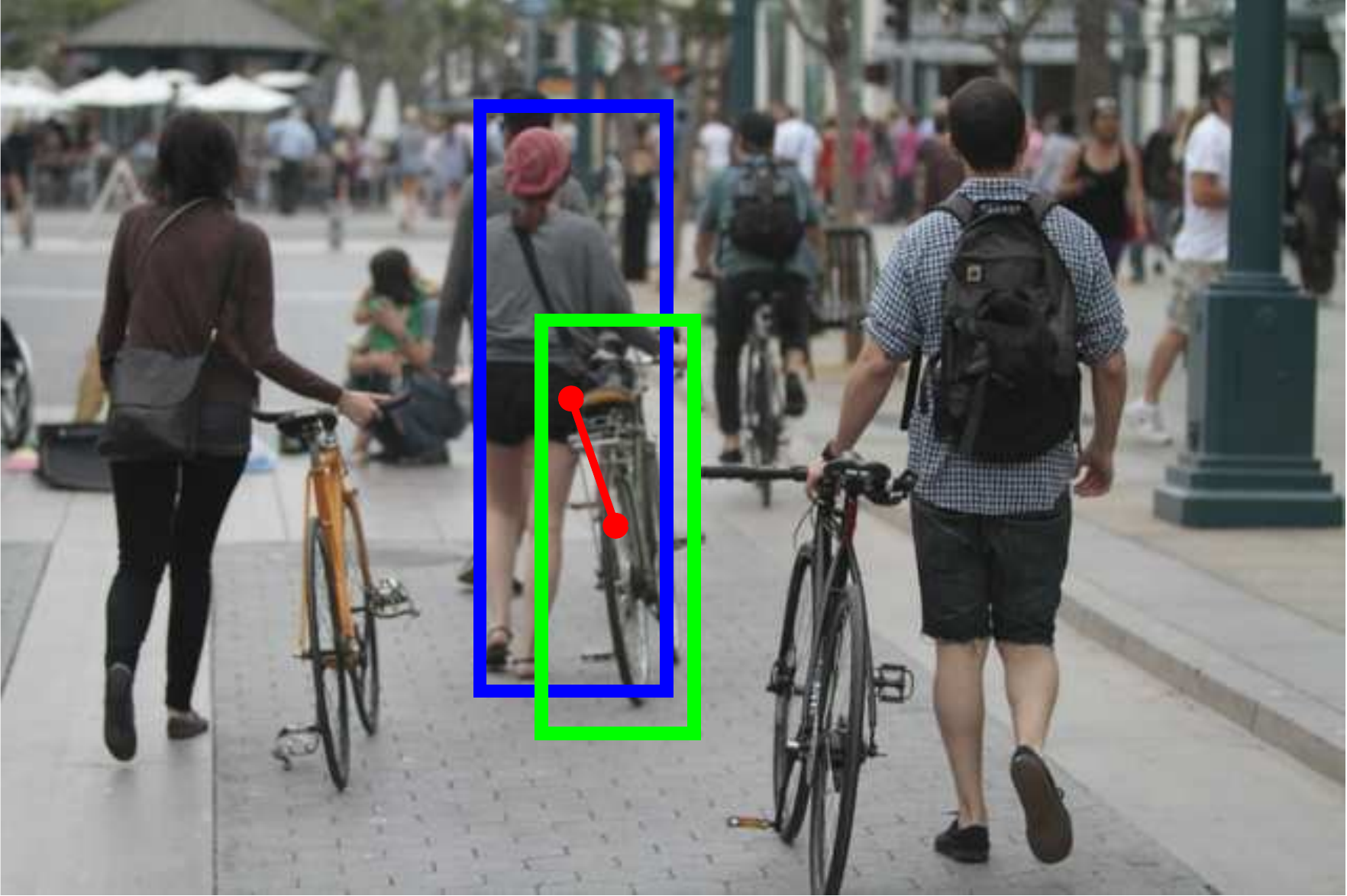}
    &
    \includegraphics[height=0.115\textwidth]{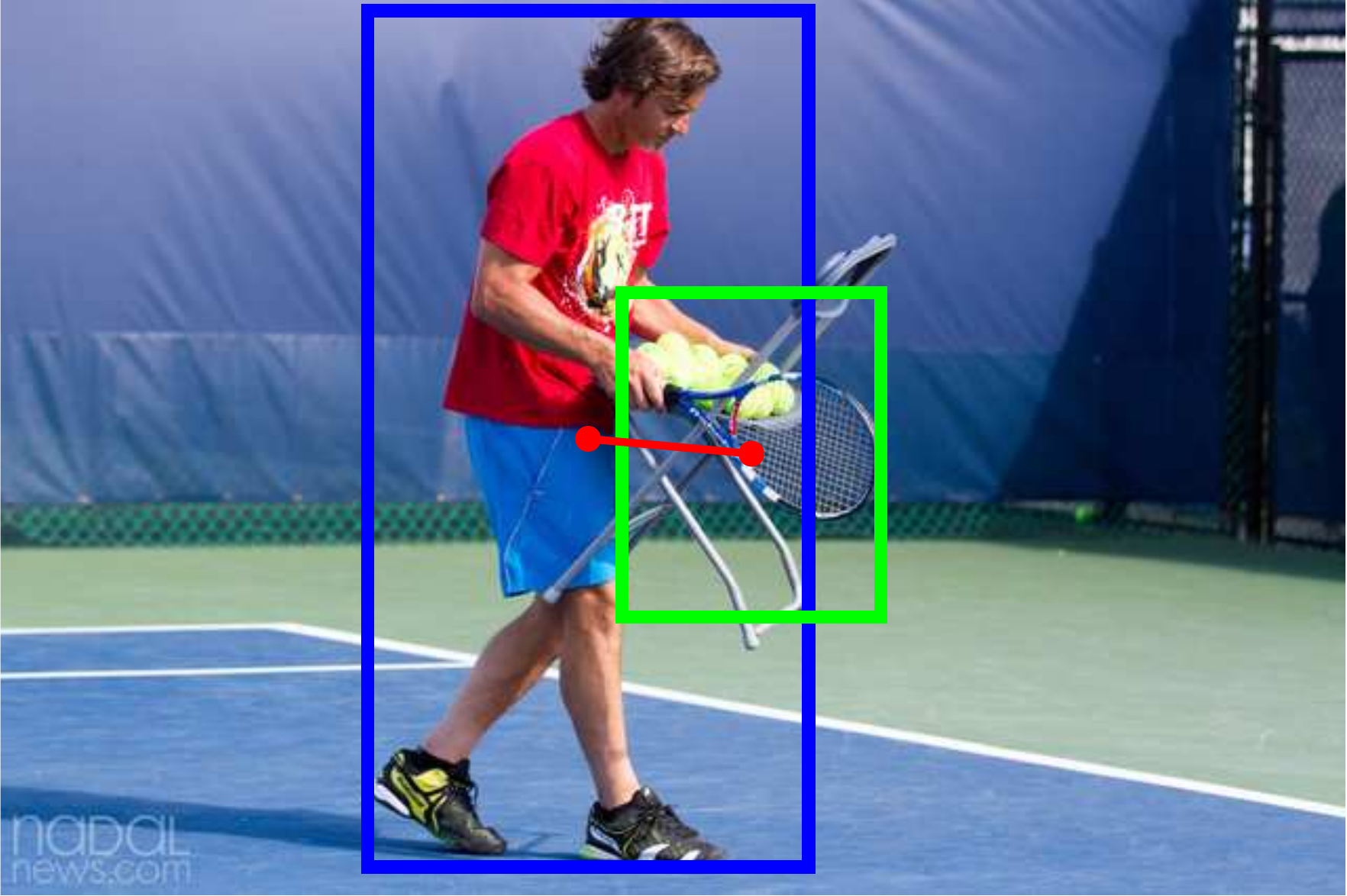}~
    \includegraphics[height=0.115\textwidth]{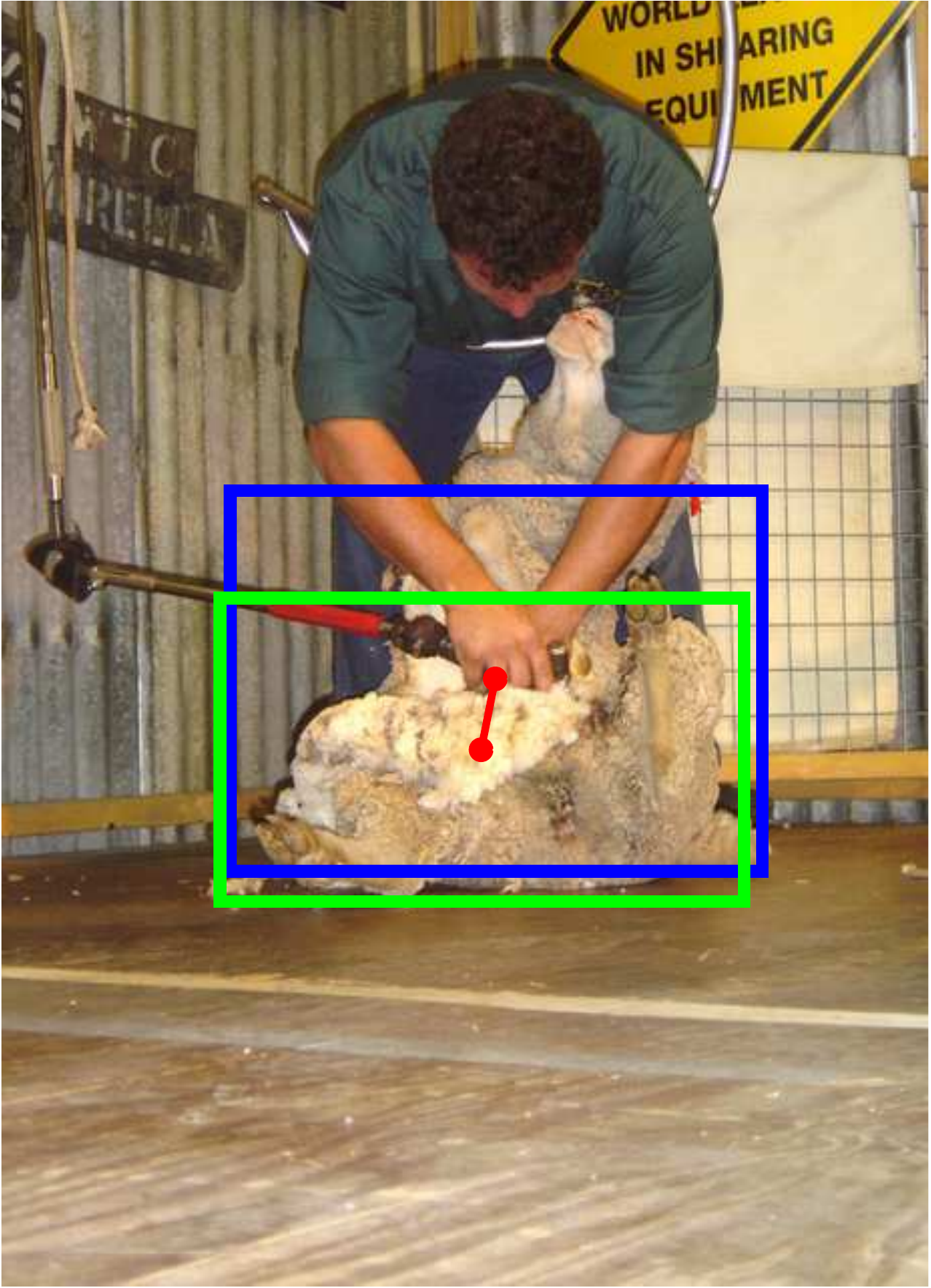}~
    \includegraphics[height=0.115\textwidth]{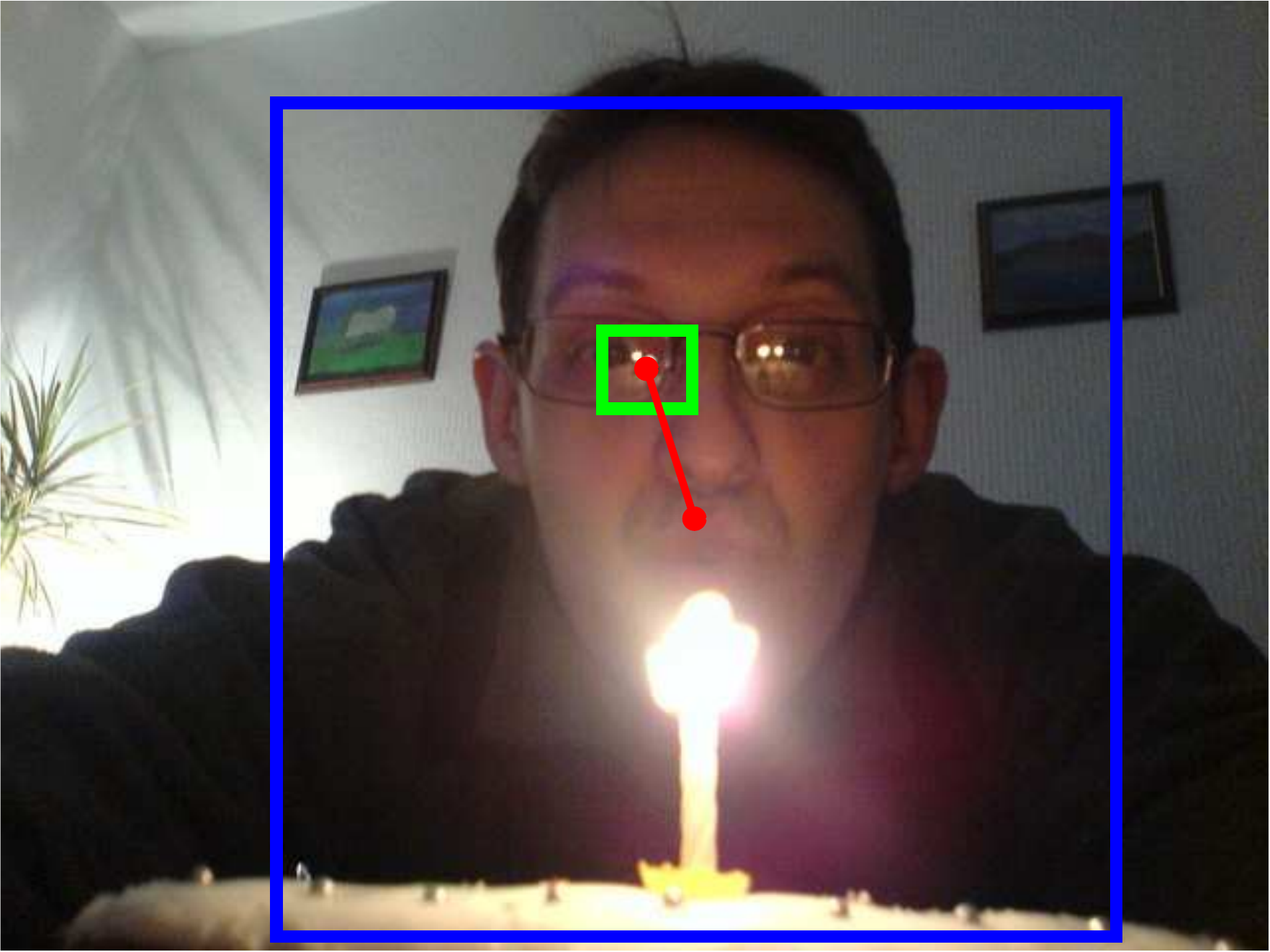}
    \\
    \hspace{-2.5mm}
    ~~~~washing a motorcycle~~~~~hugging a cat~~kicking a ball~~~~~~~~~walking a bicycle
    &
    swinging a tennis racket~~~shearing a sheep~~~sipping a wine glass
    \\
    ~~~~~~~~~~~~~~~~0.10~~~~~~~~~~~~~~~~~~~~~~~~~~0.82~~~~~~~~~~~~~~~~0.96~~~~~~~~~~~~~~~~~~~~~~~~~~~0.64
    &
    ~~~~~~~~~~~~~~~~0.85~~~~~~~~~~~~~~~~~~~~~~~~~~~~0.97~~~~~~~~~~~~~~~~~~~~~~~~~0.33
    \\
  \end{tabular}
  \vspace{-1mm}
  \caption{\small Qualitative examples of detections from our HO-RCNN. We show
the HOI class and the output probability below each detection. Left: true
positives. Right: false positives (left/middle/right: incorrect interaction
class/inaccurate bounding box/false object detection).}
  \vspace{-2mm}
  \label{fig:qual}
\end{figure*}

\vspace{-3mm}

\paragraph{Error Analysis} We hypothesize that the low AP classes suffer from
excessive false negatives. To verify this hypothesis, we compute the recall of
the human-object proposals for each HOI category. Tab.~\ref{tab:recall} shows
the mean recall on the training set by varying the numbers of used human
(object) detections. When adopting 10 human (object) detections, we see a low
mean recall (46.75\%), which expalins the low mAPs in our results. Although the
mAPs can be potentially improved by adopting more human (object) detections,
the number of human-object proposals will increase quadratically, making the
evaluation of all proposals infeasible. This thus calls for better approaches
to construct high-recall human-object proposals in future studies.

\begin{table}[t]
  \centering
  \footnotesize
  \begin{tabular}{|l||C{0.80cm}|C{0.80cm}|C{0.80cm}|C{0.80cm}|}
    \hline \TBstrut
                    & \multicolumn{4}{c|}{Number of human (object) detections} \\
    \cline{2-5} \TBstrut
                    & 10    & 20    & 50    & 100   \\
    \hline \Tstrut
    Full            & 46.75 & 51.56 & 57.17 & 60.37 \\
    Rare            & 54.15 & 58.62 & 64.98 & 68.40 \\ \Bstrut
    Non-Rare        & 44.54 & 49.45 & 54.84 & 57.97 \\
    \hline
  \end{tabular}
  \vspace{-2mm}
  \caption{\small Mean recall (\%) of human-object proposals on the training set.}
  \vspace{-1mm}
  \label{tab:recall}
\end{table}

\vspace{-3mm}

\paragraph{Comparison with Prior Approaches} To compare with prior approaches,
we consider two extensions to Fast-RCNN \cite{girshick:iccv2015}. (1) Fast-RCNN
(union): For each human-object proposal, we take their attention window as the
region proposal for Fast-RCNN. This can be seen as a ``single-stream'' version
of HO-RCNN where the feature is extracted from the tightest window enclosing
the human and object bounding box. (2) Fast-RCNN (score): Given the
human-object proposals obtained from the object detectors, we train a
classifier to classify each HOI category by linearly combining the human and
object detection scores. Note that this method does not use any features from
the human and object regions nor their spatial relations. We also report a
baseline that randomly assigns scores to our human-object proposals (Random).
Tab.~\ref{tab:comparison} shows the mAP of the compared methods and different
vairants of our HO-RCNN. In both settings, Fast-RCNN (union) performs worse
than all other methods except the random baseline. This suggests that the
feature extracted from the attention window is not suitable for distinguishing
HOI, possibly due to the irrelevant contexts between the human and object when
the two bounding boxes are far apart. Fast-RCNN (score) performs better than
Fast-RCNN (union), but still worse than all our HO-RCNN variants. This is
because object detection scores alone do not contain sufficient information for
distinguishing interactions. Finally, our HO+IP1 (conv)+S and HO+IP1 (conv)
outperform all other methods in both the Default and the Known Object setting.
Fig.~\ref{fig:qual} shows qualitative examples of the detected HOIs from our
HO-RCNN. We show both the true positives (left) and false positives (right).

\begin{table}[t]
  \centering
  \footnotesize
  \setlength{\tabcolsep}{4.6pt}
  \begin{tabular}{|l||C{1.42cm}|C{1.42cm}|C{1.42cm}|}
    \hline \TBstrut
                                               & \multicolumn{3}{c|}{Default}                                        \\
    \cline{2-4} \TBstrut
                                               & Full                & Rare                & Non-Rare                \\
    \hline \Tstrut
    Random                                     & 1.35$\times10^{-3}$ & 5.72$\times10^{-4}$ & 1.62$\times10^{-3}$     \\
    Fast-RCNN \cite{girshick:iccv2015} (union) & 1.75                & 0.58                & 2.10                    \\ \Bstrut
    Fast-RCNN \cite{girshick:iccv2015} (score) & 2.85                & 1.55                & 3.23                    \\
    \hline \Tstrut
    HO                                         & 5.73                & 3.21                & 6.48                    \\
    HO+IP1 (conv)                              & 7.30                & 4.68                & 8.08                    \\ \Bstrut
    HO+IP1 (conv)+S                            & \textbf{7.81}       & \textbf{5.37}       & \textbf{8.54}           \\
    \hline
  \end{tabular}
  \begin{tabular}{|l||C{1.42cm}|C{1.42cm}|C{1.42cm}|}
    \hline \TBstrut
                                               & \multicolumn{3}{c|}{Known Object}               \\
    \cline{2-4} \TBstrut
                                               & Full           & Rare          & Non-Rare       \\
    \hline \Tstrut
    Random                                     & 0.19           & 0.17          & 0.19           \\
    Fast-RCNN \cite{girshick:iccv2015} (union) & 2.51           & 1.75          & 2.73           \\ \Bstrut
    Fast-RCNN \cite{girshick:iccv2015} (score) & 4.08           & 2.37          & 4.59           \\
    \hline \Tstrut
    HO                                         & 8.46           & 7.53          & 8.74           \\
    HO+IP1 (conv)                              & 10.37          & \textbf{9.06} & 10.76          \\ \Bstrut
    HO+IP1 (conv)+S                            & \textbf{10.41} & 8.94          & \textbf{10.85} \\
    \hline
  \end{tabular}
  \vspace{-2mm}
  \caption{\small Comparison of mAP(\%) with prior approaches.}
  \vspace{-1mm}
  \label{tab:comparison}
\end{table}

\section{Conclusion}

We study the detection of human-object interactions in static images. We
introduce HICO-DET, a new large benchmark, by augmenting the HICO
classification benchmark with instance annotations. We propose HO-RCNN, a novel
DNN-based framework. At the core of HO-RCNN is the Interaction Pattern, a
novel DNN input that characterizes the spatial relations between two bounding
boxes. Experiments show that HO-RCNN significantly improves the performance of
HOI detection over baseline approaches.

\vspace{-3mm}

\paragraph{Acknowledgement} This publication is based upon work supported by
the King Abdullah University of Science and Technology (KAUST) Office of
Sponsored Research (OSR) under Award No. OSR-2015-CRG4-2639.

{\small
\bibliographystyle{ieee}
\bibliography{egbib}
}

\appendix

\begin{figure}[t]
  \centering
  \begin{subfigure}[c]{0.48\textwidth}
    \centering
    \includegraphics[width=1\textwidth]{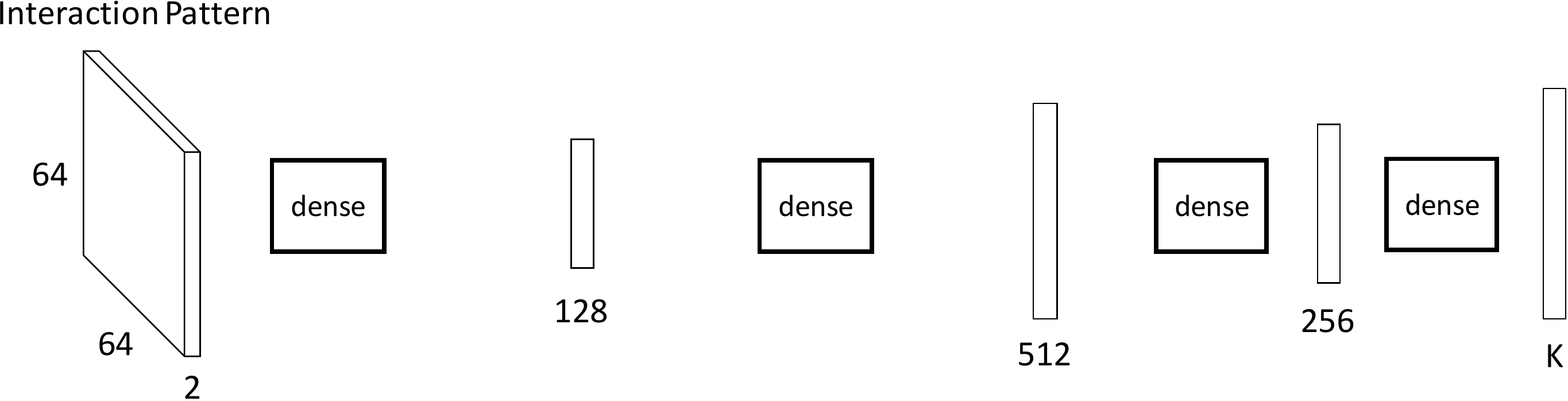}
    \caption{\small Fully-connected network (fc).}
  \end{subfigure}
  \\~\vspace{2.5mm}
  \\
  \begin{subfigure}[c]{0.48\textwidth}
    \centering
    \includegraphics[width=1\textwidth]{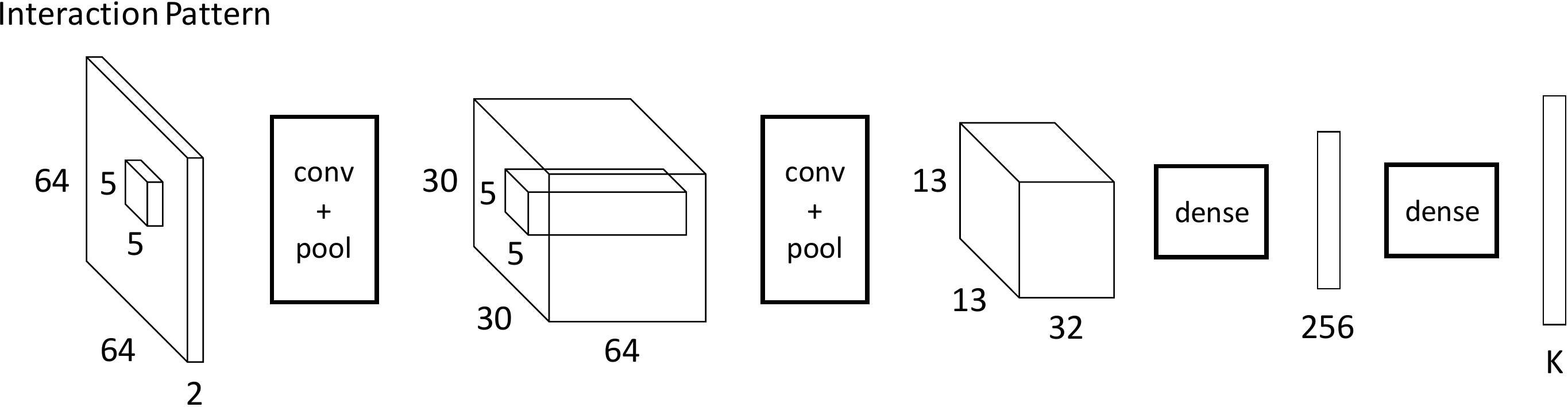}
    \caption{\small Convolutional network (conv).}
  \end{subfigure}
  \caption{\small Two different architectures for the pairwise stream.}
  \label{fig:pairwise}
\end{figure}

\section{Supplementary Material}

\subsection{Pairwise Stream}
In our HO-RCNN, we consider two different network architectures for the
pairwise stream: a fully-connected network (fc) and a convolutional network
(conv). The two architectures are illustrated in Fig. \ref{fig:pairwise}. Both
architectures take as input an Interaction Pattern and produce as output a
vector of classification scores on $K$ HOI classes of interest.  For fair
comparison, both networks have approximately the same number of parameters, and
are trained with identical schemes.

\end{document}